\documentclass{article}

\usepackage[main, preprint]{neurips_2026}

\usepackage[utf8]{inputenc}
\usepackage[T1]{fontenc}

\DeclareUnicodeCharacter{2265}{$\geq$}
\DeclareUnicodeCharacter{2264}{$\leq$}
\DeclareUnicodeCharacter{2260}{$\neq$}
\DeclareUnicodeCharacter{2248}{$\approx$}
\DeclareUnicodeCharacter{2192}{$\rightarrow$}
\DeclareUnicodeCharacter{2190}{$\leftarrow$}
\DeclareUnicodeCharacter{03B8}{$\theta$}
\DeclareUnicodeCharacter{03B4}{$\delta$}
\DeclareUnicodeCharacter{03B1}{$\alpha$}
\DeclareUnicodeCharacter{03BB}{$\lambda$}
\DeclareUnicodeCharacter{0394}{$\Delta$}
\DeclareUnicodeCharacter{2208}{$\in$}
\DeclareUnicodeCharacter{00D7}{$\times$}
\DeclareUnicodeCharacter{00B7}{$\cdot$}
\DeclareUnicodeCharacter{00B0}{$^{\circ}$}
\DeclareUnicodeCharacter{00B1}{$\pm$}
\DeclareUnicodeCharacter{2026}{\ldots}
\DeclareUnicodeCharacter{2014}{---}
\DeclareUnicodeCharacter{2013}{--}
\DeclareUnicodeCharacter{2018}{`}
\DeclareUnicodeCharacter{2019}{'}
\DeclareUnicodeCharacter{201C}{``}
\DeclareUnicodeCharacter{201D}{''}
\DeclareUnicodeCharacter{200B}{}
\DeclareUnicodeCharacter{200C}{}
\DeclareUnicodeCharacter{2713}{\checkmark}
\DeclareUnicodeCharacter{2717}{$\times$}
\DeclareUnicodeCharacter{2191}{$\uparrow$}
\DeclareUnicodeCharacter{2193}{$\downarrow$}
\DeclareUnicodeCharacter{03C0}{$\pi$}
\DeclareUnicodeCharacter{03C3}{$\sigma$}
\DeclareUnicodeCharacter{03BC}{$\mu$}
\DeclareUnicodeCharacter{03B5}{$\epsilon$}
\DeclareUnicodeCharacter{03B3}{$\gamma$}
\DeclareUnicodeCharacter{03B2}{$\beta$}
\DeclareUnicodeCharacter{03A3}{$\Sigma$}
\DeclareUnicodeCharacter{2202}{$\partial$}
\DeclareUnicodeCharacter{2207}{$\nabla$}
\DeclareUnicodeCharacter{221E}{$\infty$}
\DeclareUnicodeCharacter{2261}{$\equiv$}
\DeclareUnicodeCharacter{2295}{$\oplus$}
\DeclareUnicodeCharacter{2297}{$\otimes$}
\DeclareUnicodeCharacter{2200}{$\forall$}
\DeclareUnicodeCharacter{2203}{$\exists$}
\DeclareUnicodeCharacter{2229}{$\cap$}
\DeclareUnicodeCharacter{222A}{$\cup$}
\DeclareUnicodeCharacter{2282}{$\subset$}
\DeclareUnicodeCharacter{2286}{$\subseteq$}
\DeclareUnicodeCharacter{00A0}{ }
\usepackage{enumitem}
\usepackage{amsmath,amssymb,amsthm}
\usepackage{algorithm}
\usepackage{algpseudocode}
\usepackage{microtype}
\usepackage{hyperref}
\usepackage{url}
\usepackage{graphicx}
\usepackage{booktabs}
\usepackage{makecell}
\usepackage[most]{tcolorbox}
\usepackage{caption}
\usepackage{subcaption}
\usepackage{tikz}
\usepackage{listings}
\usepackage{color, colortbl}
\usepackage{natbib}
\usepackage{acro}

\DeclareAcronym{llm}{short=LLM,long=large language model}
\DeclareAcronym{cot}{short=CoT,long=chain-of-thought}
\DeclareAcronym{icl}{short=ICL,long=in-context learning}
\DeclareAcronym{cd}{short=CD,long=Chamfer distance}
\DeclareAcronym{icp}{short=ICP,long=iterative closest point}
\DeclareAcronym{lhs}{short=LHS,long=Latin hypercube sampling}
\DeclareAcronym{ga}{short=GA,long=genetic algorithm}
\DeclareAcronym{moe}{short=MoE,long=mixture-of-experts}
\DeclareAcronym{sql}{short=SQL,long=structured query language}
\DeclareAcronym{bfgs}{short=BFGS,long=Broyden--Fletcher--Goldfarb--Shanno}
\DeclareAcronym{pso}{short=PSO,long=particle swarm optimization}
\DeclareAcronym{umap}{short=UMAP,long=uniform manifold approximation and projection}
\DeclareAcronym{hdbscan}{short=HDBSCAN,long=hierarchical density-based spatial clustering of applications with noise}

\usepackage{xcolor}
\definecolor{compColor}{RGB}{0, 102, 204}    
\definecolor{advColor}{RGB}{204, 0, 0}      
\definecolor{reflColor}{RGB}{0, 128, 128}   
\definecolor{metaColor}{RGB}{102, 0, 204}   
\colorlet{accent}{compColor}
\definecolor{Gray}{gray}{0.9}
\colorlet{accent}{compColor}
\definecolor{bestcell}{RGB}{220,235,250}  

\definecolor{criticColor}{RGB}{230,120,0}      
\definecolor{archColor}{RGB}{0, 140, 90}     
\definecolor{failColor}{RGB}{176, 32, 32}    
\definecolor{layer1bg}{RGB}{232, 243, 255}   
\definecolor{layer2bg}{RGB}{255, 234, 234}   
\definecolor{layer3bg}{RGB}{242, 234, 255}   

\usetikzlibrary{arrows.meta, positioning, shapes.geometric,
                decorations.pathreplacing, backgrounds, calc, fit, patterns}

\definecolor{stage1}{RGB}{31,119,180}
\definecolor{stage2}{RGB}{214,39,40}
\definecolor{stage3}{RGB}{44,160,44}
\definecolor{outputC}{RGB}{148,103,189}

\newcommand{\stageone}[1]{\textcolor{stage1}{\textbf{#1}}}
\newcommand{\stagetwo}[1]{\textcolor{stage2}{\textbf{#1}}}
\newcommand{\stagethree}[1]{\textcolor{stage3}{\textbf{#1}}}
\newcommand{\out}[1]{\textcolor{outputC}{\textbf{#1}}}

\definecolor{roleOrange}{RGB}{230,159,0}      
\definecolor{roleSky}{RGB}{86,180,233}        
\definecolor{roleGreen}{RGB}{0,158,115}       
\definecolor{rolePurple}{RGB}{204,121,167}    
\definecolor{roleBlue}{RGB}{0,114,178}        
\definecolor{roleVermillion}{RGB}{213,94,0}   

\newcommand{\roletag}[2]{\textcolor{#1}{\textbf{[#2]}}}

\newtcblisting{promptpart}[2]{%
  enhanced, breakable, listing only,
  colback=#1!8, colframe=#1, coltitle=white,
  colbacktitle=#1, fonttitle=\bfseries\small,
  attach boxed title to top left={xshift=4pt, yshift=-2pt},
  boxed title style={size=small, sharp corners, frame hidden, colback=#1},
  title=#2, sharp corners=southwest, arc=2pt, boxrule=0.6pt,
  left=6pt, right=6pt, top=8pt, bottom=4pt,
  listing options={basicstyle=\ttfamily\footnotesize,
                   breaklines=true, breakatwhitespace=false,
                   columns=fullflexible, keepspaces=true,
                   showstringspaces=false}
}

\newtcbtheorem[number within=section]{defbox}{Definition}%
{enhanced,
 colback=green!6,
 colframe=green!60!black,
 arc=3mm,
 boxrule=0.8pt,
 left=6pt, right=6pt, top=6pt, bottom=6pt,
 fonttitle=\itshape\bfseries
}{def}

\newcommand{\ModularDeltaBars}{2.0}
\newcommand{\RAPSFullDeltaBars}{1.5}

\newcommand{\RAPSNoSelRefDeltaBars}{1.9}

\newcommand{\RAPSFullDeltaBarsComp}{1.4}

\newcommand{\RAPSNoAdvMetaDeltaBarsComp}{2.2}

\newcommand{\RAPSFullNDIComp}{41.2}

\newcommand{\ModularNDIComp}{68.3}
\newcommand{\RAPSFullRob}{0.132}
\newcommand{\RAPSNoAdvMetaRob}{0.458}

\newcommand{\RobustnessSpeedup}{3.5}
\newcommand{\AdhDegradationPct}{28}
\newcommand{\QwenSmallDeltaBars}{1.1}
\newcommand{\LlamaDeltaBars}{1.5}
\newcommand{\MoEDeltaBars}{2.0}
\newcommand{\QwenSmallExact}{53.8}
\newcommand{\LlamaExact}{19.0}
\newcommand{\MoEExact}{5.8}

\newcommand{\LlamaRob}{0.067}

\newcommand{\MoEDist}{22.8}
\newcommand{\BackboneScaleRatio}{17.5}

\newcommand{\TotalShapes}{32}

\newcommand{\EnumGAShapesStandard}{5}
\newcommand{\EnumGAShapesLetters}{26}

\newcommand{\EnumGAOverallSpeedup}{2.1}

\newcommand{\EnumGACircleSpeedup}{40.3}

\newcommand{\EnumGANACASpeedup}{29.4}

\newcommand{\EnumGAStdSpeedup}{3.3}
\newcommand{\EnumGALtrSpeedup}{1.9}

\newcommand{\QualTopologyPct}{59.2\%}
\newcommand{\QualParamPct}{31.2\%}
\newcommand{\QualOptPct}{7.2\%}
\newcommand{\QualRobustPct}{2.4\%}

\newcommand{\QualSuccessAfterFailPct}{13.0\%}

\newcommand{\QualNewHeuristics}{1015}
\newcommand{\QualInvalidatedHeuristics}{10}

\newcommand{\ReasoningModeAbductivePct}{90.4\%}
\newcommand{\ReasoningModeCorrectivePct}{7.2\%}
\newcommand{\ReasoningModeCounterfactualPct}{2.4\%}

\newcommand{\MetaLearningAcceleration}{46\%}
\newcommand{\AvgIterationsEpOne}{4.9}
\newcommand{\AvgIterationsEpFourPlus}{2.7}
\newcommand{\InvalidatedRuleCount}{8}
\newcommand{\AdversarialGapMedian}{0.038}
\newcommand{\AdversarialGapMax}{5.114}

\newcommand{\EntanglementBaseline}{1.94}
\newcommand{\EntanglementNoAdvMeta}{1.03}
\newcommand{\EntanglementP}{< 10^{-4}}

\newcommand{\HeuristicAppliedMulti}{23}

\newcommand{\HeuristicRuleIds}{24}
\newcommand{\HeuristicCrossShapeInval}{2}
\newcommand{\HeuristicTotalInval}{4}

\newcommand{\HVFull}{0.96}
\newcommand{\HVNoAdvMeta}{0.99}
\newcommand{\HVNoSelRef}{0.96}

\title{R-APS: Compositional Reasoning and In-Context Meta-Learning\\for Constrained Design via Reflective Adversarial Pareto Search}

\author{
Jo\~{a}o Pedro Gandarela\textsuperscript{1,2} \quad
Thiago Rios\textsuperscript{3} \quad
Stefan Menzel\textsuperscript{3} \quad
Andr\'{e} Freitas\textsuperscript{1,4,5} \\
\textsuperscript{1}Idiap Research Institute, Switzerland \quad
\textsuperscript{2}\'{E}cole Polytechnique F\'{e}d\'{e}rale de Lausanne (EPFL), Switzerland \\
\textsuperscript{3}Honda Research Institute Europe, Germany \\
\textsuperscript{4}Department of Computer Science, University of Manchester, UK \\
\textsuperscript{5}National Biomarker Centre, CRUK-MI, University of Manchester, UK \\
\texttt{firstname.lastname@idiap.ch} \quad
\texttt{firstname.lastname@honda-ri.de}
}

\begin{document}
\maketitle

\begin{abstract}
\Acp{llm} are fluent reasoners on open-ended tasks, yet on \emph{constrained design} problems, and more broadly in \emph{agentic} settings where an autonomous system must plan, use tools, and act over extended horizons, fluency does not translate into reliable delivery. We trace this gap to three coupled structural failures that affect agentic AI models in constrained domains: failures propagate without localization, worst-case perturbations are not evaluated, and accumulated knowledge is never invalidated. We argue that these failures share a single root cause, namely that abductive, counterfactual, meta-inductive, corrective, and inductive reasoning pull a shared context in incompatible cognitive directions. We introduce \textbf{Reflective Adversarial Pareto Search (R-APS)}, an agentic AI method and, to our knowledge, the first method that addresses all three failures jointly through \emph{reasoning-mode decomposition}: a novel design principle that allocates each of the five reasoning modes its own cognitive context and orchestrates their interaction across three timescales, \emph{intra-stage} staged compositional reasoning with \emph{typed validation critic} for failure localization, \emph{intra-episode} sensitivity-guided counterfactual stress-testing of robustness as a first-class Pareto objective, and \emph{inter-episode} meta-inductive rule extraction with explicit invalidation constituting persistent agentic memory. Crucially, R-APS achieves these gains with a \emph{frozen} reasoning \acs{llm}, without parameter updates or fine-tuning, and purely through structured protocol design. We instantiate and evaluate R-APS on \emph{planar mechanism synthesis}, a fundamental engineering problem underlying robotics, prosthetics, and mechanical design, in which every candidate output is checked by a kinematic solver. On \TotalShapes{} target trajectories (6~standard curves and the 26~English-alphabet letters), R-APS produces robustness certificates $\RobustnessSpeedup{}\times$ tighter than uniform-perturbation baselines, accelerates iterations-to-first-archive-admission by \MetaLearningAcceleration{} across episodes on the same shape, and attains a $\EnumGAOverallSpeedup{}\times$ mean Chamfer-distance reduction over classical Enum+\acs{ga} on standard curves while jointly controlling bar-count adherence and worst-case robustness. In a cross-backbone study, we further find that small reasoning-specialized 4B models are competitive with general-purpose 70B backbones inside the protocol, suggesting that well-structured agentic protocols can partially offset raw model scale. More broadly, the protocol is parameterized by domain interfaces (typed verification cascade, sensitivity primitive, refinement log) rather than mechanism-design knowledge, and the agentic loop generalizes naturally to \acs{sql} synthesis, circuit design, and robot motion planning.
\end{abstract}

\section{Introduction}
\label{sec:intro}

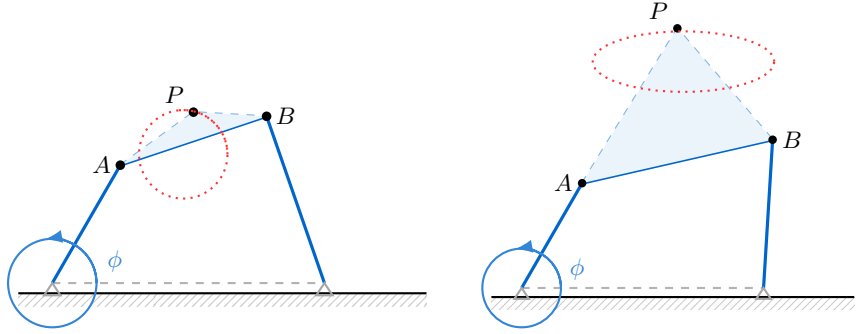
\begin{figure}[ht]
\centering
\begin{subfigure}{.5\textwidth}
\centering
\begin{tikzpicture}[>=Latex, scale=0.9]
  \coordinate (A0) at (0,0);
  \coordinate (B0) at (4,0);
  \coordinate (A) at ({2*cos(60)},{2*sin(60)});
  \coordinate (B) at (3.15,2.45);
  \coordinate (P) at ($(A)!0.50!(B)+(0,0.42)$);

  \fill[pattern=north east lines, pattern color=gray!40] (-0.5,-0.35) rectangle (5.5,-0.15);
  \draw[thick] (-0.5,-0.15) -- (5.5,-0.15);

  \draw[thick, dashed, gray!60] (A0) -- (B0);

  \draw[very thick, accent] (A0) -- (A) node[midway, left, font=\small] {}; 
  \draw[very thick, accent] (A) -- (B) node[midway, above left, font=\small] {}; 
  \draw[very thick, accent] (B0) -- (B) node[midway, right, font=\small] {}; 

  \draw[thick, dashed, accent!50] (A) -- (P) -- (B);
  \fill[accent!8] (A) -- (P) -- (B) -- cycle;

  \foreach \pt/\lbl/\pos in {A/$A$/left, B/$B$/right, P/$P$/above left}{
    \fill (\pt) circle (2pt);
    \node[\pos, font=\small] at (\pt) {\lbl};
  }
  \foreach \pt in {A0, B0}{
    \draw[thick, gray!70] (\pt) ++(-0.12,-0.18) -- ++(0.24,0) -- (\pt) -- cycle;
  }

  \draw[->, thick, accent!80] (0.65,0) arc[start angle=0, end angle=460, radius=0.65];
  \draw[dotted, thick, red!80] ($(A)!0.50!(B)+(0.5,-0.2)$) arc[start angle=0, end angle=460, radius=0.65];
  \node[accent!80, font=\small] at (0.92,0.32) {$\phi$};


\end{tikzpicture}
\end{subfigure}%
\begin{subfigure}{.49\textwidth}
\begin{tikzpicture}[>=Latex, scale=0.8]
  \coordinate (A0) at (0,0);
  \coordinate (B0) at (4,0);
  \coordinate (A) at ({2*cos(60)},{2*sin(60)});
  \coordinate (B) at (4.15,2.45);
  \coordinate (P) at ($(A)!0.50!(B)+(0,2.2)$);

  \fill[pattern=north east lines, pattern color=gray!40] (-0.5,-0.35) rectangle (5.5,-0.15);
  \draw[thick] (-0.5,-0.15) -- (5.5,-0.15);

  \draw[thick, dashed, gray!60] (A0) -- (B0);

  \draw[very thick, accent] (A0) -- (A) node[midway, left, font=\small]  {};
  \draw[very thick, accent] (A) -- (B) node[midway, above left, font=\small] {};
  \draw[very thick, accent] (B0) -- (B) node[midway, right, font=\small] {};

  \draw[thick, dashed, accent!50] (A) -- (P) -- (B);
  \fill[accent!8] (A) -- (P) -- (B) -- cycle;

  \foreach \pt/\lbl/\pos in {A/$A$/left, B/$B$/right, P/$P$/above left}{
    \fill (\pt) circle (2pt);
    \node[\pos, font=\small] at (\pt) {\lbl};
  }
  \foreach \pt in {A0, B0}{
    \draw[thick, gray!70] (\pt) ++(-0.12,-0.18) -- ++(0.24,0) -- (\pt) -- cycle;
  }

  \draw[->, thick, accent!80] (0.65,0) arc[start angle=0, end angle=460, radius=0.65];
  \draw[dotted, thick, red!80] ($(A)!0.50!(B)+(0.1,1.65)$) ellipse (1.5 and 0.5);
  \node[accent!80, font=\small] at (0.92,0.32) {$\phi$};

  \node[font=\footnotesize, accent, fill=white, inner sep=1pt] at (2,-0.45) {};

\end{tikzpicture}
  \label{fig:sub2}
\end{subfigure}
\caption{Six-bar linkage configuration before and after adjustment. The left panel shows the initial setup, and the right panel shows the adjusted configuration. The input crank rotates with angle $\phi$, driven by the motor, while the red dashed curve near point $P$ indicates the end-effector \textbf{trajectory}.}
\label{fig:running-example}
\end{figure}

Consider an agent tasked with designing a planar mechanical linkage whose endpoint (Fig.\ref{fig:running-example} point (P)) traces an ellipse.
The agent proposes a bar topology, optimizes link lengths numerically, and checks the resulting trajectory against the target.
The check fails, the trajectory overshoots the apex.
So the agent discards \emph{everything} and starts over: topology, parameters, the carefully reasoned decisions that got it this far.
Yet the topology was valid.
Only the optimizer needed adjusting.
There was simply no mechanism to say so.

This scenario recurs across \emph{constrained design} tasks more broadly: settings where an output decomposes into interdependent typed stages and every constraint must hold simultaneously, including database query synthesis~\cite{pourreza2024dinsql}, robot task planning~\cite{ahn2022saycan}, circuit design~\cite{thakur2023circuitgen}, and planar mechanism synthesis~\cite{Song2025}.
\Acp{llm} have become fluent reasoners on open-ended tasks~\cite{wei2022cot,yao2023react}, and a new generation of \emph{agentic AI} methods~\cite{xi2023rise,wang2024survey} now wraps them in autonomous loops that plan, invoke tools, and adapt to feedback.
Yet a persistent gap remains between how fluently a model can \emph{discuss} a solution and how reliably an agentic method can \emph{deliver} one: a reasoning failure here is not a stylistic blemish but an infeasible artifact, discarded in full.

Three structural failure modes compound when monolithic \acp{llm} are used as agentic methods in this regime: (i)~\emph{failure propagation without localization}: a single violated constraint forces complete regeneration; (ii)~\emph{absence of robustness certification}: nominal performance says nothing about worst-case behavior under realistic perturbation, a critical reliability gap for any autonomously deployed agent; (iii)~\emph{monotonic heuristic accumulation}: methods that learn from experience never invalidate stale rules and discard refinement trajectories, causing knowledge quality to degrade over episodes.
We argue that (i)-(iii) are coupled symptoms of current agentic \ac{llm} designs lacking a \emph{reasoning protocol} that keeps abductive, counterfactual, meta-inductive, corrective, and inductive reasoning in cognitively distinct contexts rather than entangled in a single prompt.
The central hypothesis is that failures (i)-(iii) can be addressed jointly by \emph{reasoning-mode decomposition}, an explicit agentic protocol that allocates each mode its own specialized context:

\begin{quote}
\emph{Can a frozen reasoning \ac{llm} be turned into a reliable constrained-design agent purely by structuring its reasoning protocol, localizing its own failures, certifying its own robustness, and distilling its own refinement trajectories into transferable knowledge, without any parameter updates?}
\end{quote}

Existing approaches give partial remedies but, to our knowledge, no prior method delivers the full bundle: Reflexion~\cite{shinn2024reflexion} and ReAct~\cite{yao2023react} reflect verbally but treat each generation as atomic, so selective correction is impossible in principle and no robustness guarantee is produced; Voyager~\cite{wang2023voyager} and ExpeL~\cite{zhao2024expel} accumulate skills monotonically and discard failure trajectories; general-purpose agentic frameworks such as AutoGen~\cite{wu2023autogen} and tool-augmented agents~\cite{schick2023toolformer} decompose by workflow role rather than by reasoning mode, leaving the entanglement problem unaddressed.

We present \textbf{R-APS (Reflective Adversarial Pareto Search)}, an \textbf{agentic AI method} and the first, to our knowledge, to instantiate \emph{reasoning-mode decomposition} as a complete protocol addressing all three structural failures jointly, organized along three interacting timescales (full mode taxonomy in Section~\ref{sec:modes}).
\textbf{(1)~Intra-stage:} staged compositional reasoning (in the sense of typed decomposition, not Lake-style compositional generalization~\cite{lake2017building}) with \emph{typed validation critics} that attribute every failure to a specific named stage and route selective refinement only there, reducing constraint-adherence deviation from \RAPSNoSelRefDeltaBars{} to \RAPSFullDeltaBars{}.
\textbf{(2)~Intra-episode:} sensitivity-guided counterfactual stress-testing in which a Designer (abductive), Adversary (counterfactual), and Critic (meta-inductive) play a min-max game with Sobol-screened directed perturbation, yielding robustness certificates $\RobustnessSpeedup{}\!\times$ tighter than uniform-perturbation baselines (\RAPSFullRob{} vs.\ \RAPSNoAdvMetaRob{}).
\textbf{(3)~Inter-episode:} meta-inductive rule extraction with explicit invalidation mines complete refinement trajectories including failures, serving as the agentic model's \emph{long-term memory}; the resulting policy improves across episodes \emph{without parameter updates}, delivering a \MetaLearningAcceleration{} acceleration in iterations-to-first-archive-admission on repeated shapes.

\paragraph{Contributions.}
\begin{enumerate}[leftmargin=*,itemsep=1pt,topsep=2pt]
    \item A novel \textbf{reasoning-mode decomposition} principle for agentic AI, identifying five reasoning modes (abductive, counterfactual, meta-inductive, corrective, inductive) whose entanglement in a shared context is, to our knowledge, the first unified explanation of three otherwise-disconnected failure modes.
    \item The first multi agent protocol, to our knowledge, to compose typed validation critics (intra-stage), counterfactual stress-testing as a Pareto objective (intra-episode), and meta-inductive rule extraction with \emph{explicit invalidation} serving as agentic long-term memory (inter-episode) into a single closed agentic loop, all operating on a frozen \acs{llm} without parameter updates or fine-tuning.
    \item Empirical validation on \TotalShapes{} shapes: a \MetaLearningAcceleration{} inter-episode acceleration, a $\EnumGAOverallSpeedup{}\!\times$ mean Chamfer-distance reduction over Enum+\acs{ga} on standard curves, robustness certificates $\RobustnessSpeedup{}\!\times$ tighter than uniform-perturbation baselines, and 4B reasoning-specialized models competitive with 70B general-purpose models inside the protocol.
    \item Architectural and empirical support for the decomposition: a falsifiable stage-by-mode responsibility map whose predicted ablation signature, each timescale owns one and only one of the three failure axes, is confirmed empirically.
\end{enumerate}

\section{R-APS: A Novel Reasoning-Mode Decomposition Method}
\label{sec:method}

\begin{figure*}[t]
\centering
\includegraphics[width=\textwidth]{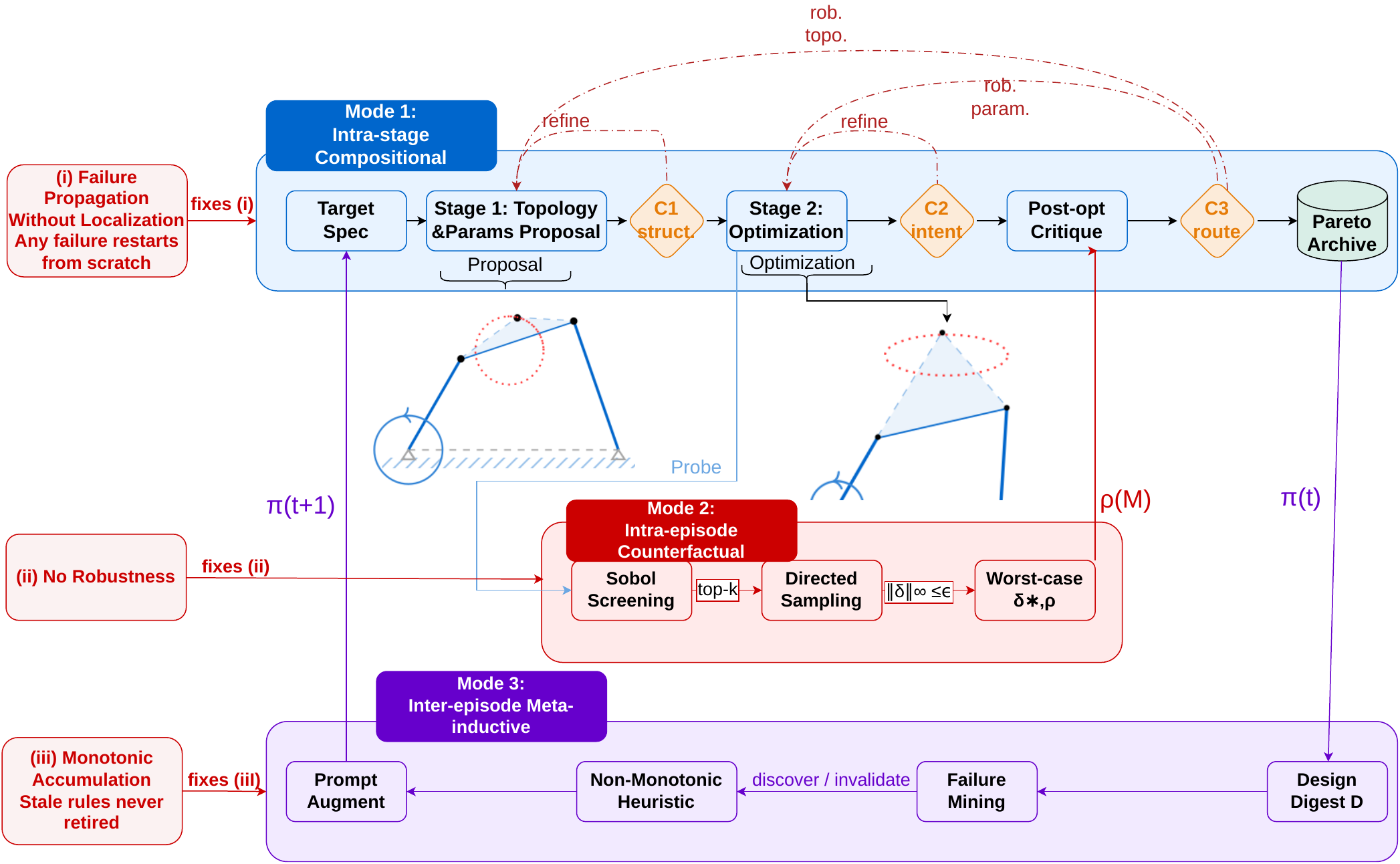}
\caption{\textbf{R-APS: three-timescale reasoning-mode separation.}
\textbf{Left (red):} the three structural failures each timescale addresses.
\textcolor{compColor}{\textbf{Mode~1 (Intra-stage)}} decomposes design into typed stages with \textcolor{criticColor!80!black}{\textbf{typed validation critics}} C1/C2/C3; selective refinement corrects only the diagnosed stage~(i).
The inset linkages show the intra-stage progression: left, the raw topology proposal (unconstrained geometry, no target); right, the same mechanism after optimization, with the coupler trajectory (red dotted ellipse) aligned to the target.
\textcolor{advColor}{\textbf{Mode~2 (Intra-episode)}} runs Sobol-directed stress-testing, supplying the robustness certificate $\rho(M)$ that eliminates failure~(ii).
\textcolor{metaColor}{\textbf{Mode~3 (Inter-episode)}} extracts and \emph{invalidates} heuristics from refinement trajectories; the dotted purple loop delivers $\pi^{(t{+}1)}$ without parameter updates, eliminating failure~(iii).}
\label{fig:raps_overview}
\end{figure*}

R-APS is an \textbf{agentic AI method} grounded in a single hypothesis: \emph{distinct reasoning modes optimize in incompatible cognitive directions, so reliable constrained design requires an agentic method that decomposes the reasoning modes into specialized contexts and orchestrates their interaction across an autonomous multi-step loop.}
The method decomposes reasoning into five modes (abductive, counterfactual, meta-inductive, corrective, inductive) interacting along three agentic timescales: \emph{intra-stage} staged compositional reasoning with typed validation critics (\S\ref{sec:pipeline}), \emph{intra-episode} sensitivity-guided counterfactual stress-testing (\S\ref{sec:adversarial}), and \emph{inter-episode} meta-inductive rule extraction with explicit invalidation, constituting the agentic model's long-term memory (\S\ref{sec:metalearning}).
R-APS instantiates these modes through five specialized \ac{llm} agents forming a closed agentic loop: Designer ($\pi_D$, abductive), Critic ($\pi_C$, meta-inductive), Post-Opt Critic ($\pi_{PC}$, evaluative), Refinement ($\pi_R$, corrective), Meta-Analyst (MA, inductive); see Fig.~\ref{fig:raps_overview} for how they interact and App.~\ref{app:agents} for full agent profiles.

\subsection{Problem Formulation}
\label{sec:problem}

A planar mechanism is $M = \text{Assemble}(\tau, \theta, \mathcal{C})$, with topology $\tau \in \mathcal{T}$, parameters $\theta \in \Theta(\tau)$, and kinematic constraints $\mathcal{C}$.
Given target $\mathcal{T}_{\text{target}}$, we evaluate mechanisms on two objectives: trajectory accuracy $f_1(M) = \text{CD}(\textsc{ICP}(\mathcal{T}(M)), \mathcal{T}_{\text{target}})$ via \ac{cd} with \ac{icp} alignment~\cite{besl1992icp}, and robustness $f_2(M) = \rho(M)$ measuring worst-case degradation under perturbations (\S\ref{sec:adversarial}).
The goal is to construct a Pareto archive $A$ of non-dominated mechanisms satisfying all kinematic constraints.

\subsection{Five Reasoning Modes and Why We Decompose Them}
\label{sec:modes}

Three modes descend from Peirce's classical triad~\citep{peirce1878deduction}: \textbf{abductive} (Designer, hypothesis generation), \textbf{deductive} (absorbed into the non-LLM structural validator/optimizer), and \textbf{inductive} (Meta-Analyst, rule distillation across trajectories).
Two further modes are required by design loops which the classical triad does not address: \textbf{counterfactual}~\citep{byrne2005counterfactual,pearl2009causality}, the inference-under-disprovability step that grounds worst-case stress testing, and \textbf{corrective}~\citep{lipton2004ibe}, the inference-to-best-explanation step specialized to localized-diagnosis repair.
Each of the five addresses exactly one of the coupled failure modes (Section~\ref{sec:intro}): corrective owns failure-propagation-without-localization, counterfactual owns absence-of-robustness-certification, and meta-inductive together with inductive own monotonic-accumulation-without-invalidation.
The full per-mode summary table and the stage-by-mode responsibility map are in App.~\ref{app:agents} (Tables~\ref{tab:reasoning_modes},~\ref{tab:decomposition_map}, and~\ref{tab:reasoning_type}).

\subsection{Compositional Design Method}
\label{sec:pipeline}

R-APS instantiates the intra-stage timescale as a staged method separated by validation critics with selective refinement loops; full pseudocode is Algorithm~\ref{alg:raps} in App.~\ref{app:raps_alg}.

\paragraph{Typed validation critics.}
The critics make staged compositional reasoning actionable: each critic performs a \emph{typed} diagnosis, attributing failure to a specific named stage rather than merely checking constraints.
Critic~1 (structural) verifies hard kinematic constraints: link count and assemblability.
Critic~2 (intent) checks whether the optimized trajectory semantically matches the specification via quantitative error thresholds and qualitative motion-primitive frequency matching (details in App.~\ref{app:critics}).
Failure diagnosis distinguishes topology-level errors (fundamental mismatch, $f_1 > 10\epsilon_{\text{traj}}$) from optimization failures ($f_1 > \epsilon_{\text{traj}}$), eliminating the \emph{failure propagation without localization} that plagues monolithic approaches.

\paragraph{Selective refinement.}
When a \emph{typed validation critic} diagnoses failure at stage $s$, only stage $s$ is corrected while all decisions from stages $1, \dots, s{-}1$ are preserved, the structural guarantee that makes staged compositional reasoning more than decomposition.
A topology failure triggers the next proposal from the pool; an optimization failure switches strategy (\ac{bfgs}~\cite{broyden1970convergence,fletcher1970new,goldfarb1970family,shanno1970conditioning}$\leftrightarrow$\ac{pso}~\cite{kennedy1995particle}$\leftrightarrow$Grid) and re-runs Stage~3 with the validated topology and parameters intact.
A robustness failure (critic C3 in Fig.~\ref{fig:raps_overview}) is sub-typed by the failure fingerprint emitted by the adversary: \textsc{Robustness/topology} (the nominal trajectory is itself off, despite worst-case behaviour being explored) is treated as a topology failure and re-proposed; \textsc{Robustness/param} (the design degrades sharply under perturbation) re-runs Stage~3 with the validated topology preserved and the fragile-parameter set surfaced to the optimizer (App. Table~\ref{tab:refinement}).
Refinement depth is bounded at $D{=}20$ to prevent infinite loops.

\subsection{Counterfactual Reasoning: Sensitivity-Guided Robustness Testing}
\label{sec:adversarial}

Nominal trajectory accuracy is insufficient for practical deployment: real mechanisms experience manufacturing variations and assembly errors.
R-APS realizes the \emph{intra-episode counterfactual reasoning mode} through a two-phase protocol (full pseudocode in Algorithm~\ref{alg:adversarial}, App.~\ref{app:adversarial_alg}; phase illustration in App. Fig.~\ref{fig:mc_phases}) in which the cognitively specialized constructive Designer and destructive Adversary engage in a min-max game.
The two agents are kept in separate prompts precisely because abductive generation and counterfactual stress-testing are the canonical interfering pair (Section~\ref{sec:modes}).
Phase~1 uses \ac{lhs}~\cite{mckay1979comparison} for even coverage of the tolerance hypercube and computes Sobol first-order indices~\cite{sobol2001global} to identify the most sensitive parameter dimensions; Phase~2 concentrates the remaining sampling budget on those dimensions around the high-loss centroid, locating the worst-case perturbation $\delta^* = \arg\max_{\|\delta\|_\infty \leq \epsilon} \text{CD}(\mathcal{T}(M_{\theta^* + \delta}), \mathcal{T}_{\text{target}}) - R_C(\theta^* + \delta)$.
The robustness score $\rho(M) = \|\mathbf{f}_{\text{worst}} - \mathbf{f}_{\text{nom}}\|/\max(\|\mathbf{f}_{\text{nom}}\|, \varepsilon)$ enters the Pareto archive as a first-class objective alongside trajectory accuracy.

\subsection{Meta-Inductive Reasoning: Inter-Episode Heuristic Lifecycle}
\label{sec:metalearning}

The \emph{inter-episode meta-inductive reasoning mode} is what distinguishes R-APS from methods that merely reflect within a single episode, and is the component that gives R-APS its character as a \emph{learning agentic method} rather than a stateless pipeline.
This mode constitutes the agentic model's \textbf{long-term memory}: it extracts reusable design knowledge from complete refinement trajectories, the full sequence of attempts, \emph{typed validation critics} diagnoses, corrections, and outcomes, not just from successful designs, and makes that knowledge available to all agents in subsequent episodes without parameter updates.
At each iteration $t$, the effective agentic policy is:
\begin{equation}
\pi^{(t)}(\cdot) = \text{\acs{llm}}\big(\cdot \mid \mathcal{P}_{\text{base}}, \mathcal{H}^{(t)}, \mathcal{D}^{(t)}, \mathcal{E}^{(t)}\big)
\end{equation}
where $\mathcal{H}^{(t)}$ are learned heuristics, $\mathcal{D}^{(t)}$ is a compressed design digest (representative archive exemplars selected via \ac{umap} embedding and \ac{hdbscan} clustering), and $\mathcal{E}^{(t)}$ is an exclude list of repeatedly failed topologies.

The meta-inductive mode manages a \textbf{heuristic lifecycle}: new rules are synthesized from observed patterns, existing rules are updated with new evidence, and rules whose support erodes are \emph{explicitly invalidated}, preventing the monotonic accumulation that degrades Voyager/ExpeL-style methods.
Each heuristic carries a confidence score for conflict resolution.
The meta-inductive mode operates over a different timescale than the intra-episode modes: its outputs feed the Designer and Critic prompts for subsequent episodes, so that inter-episode knowledge transfer happens in context rather than through weight updates, the key mechanism by which R-APS's agentic loop improves over time.
Conceptually, this is an in-context analogue \emph{in spirit} of inner-/outer-loop meta-learning~\cite{finn2017maml} (outer-loop rule evolution, inner-loop reflective correction), though the mechanism, context accumulation with explicit invalidation, is qualitatively different from gradient-based few-shot adaptation.
Full details of design-space compression, failure-mode clustering, and resolution-pattern mining are in App.~\ref{app:metalearning}.

\section{Experiments}
\label{sec:experiments}

We evaluate R-APS on \TotalShapes{} shapes (6~standard curves plus 26~English letters) around five questions tied to the three structural failure modes and external competitiveness: \textbf{Q0}~ablation non-contamination, \textbf{Q1}~intra-stage, \textbf{Q2}~intra-episode, \textbf{Q3}~inter-episode, \textbf{Q4}~vs.\ classical search (model-scale study in App.~\ref{app:backbone}).

\subsection{Setup}
\label{sec:setup}

\paragraph{Domain and metrics.}
We synthesize planar linkages whose end-effector trajectory tracks a target curve, on \TotalShapes{} shapes (6~standard curves: Circle, Ellipse, Line, LB, NACA airfoil, Parabola, plus the 26 letters of the English alphabet).
We report two Pareto objectives jointly (Chamfer distance \ac{cd}; worst-case sensitivity $\rho_{\text{sens}}$) plus the structural-adherence diagnostic $\Delta$Bars and a Norm.\ Dist.\ Index (per-run-median Chamfer $\times\,100$ over the median modular-baseline Chamfer; App.~\ref{app:norm}).

\paragraph{Baselines and what each tests.}
We compare against three baselines, each selected to test a specific rival explanation rather than to provide general coverage.
\textbf{Enum+\acs{ga}}~\cite{cabrera2002genetic} enumerates feasible 4- and 6-bar topologies and applies genetic algorithm optimization; it tests whether topology enumeration, the strongest non-LLM alternative, already solves the localization and robustness problems, making reasoning-mode decomposition unnecessary.
\textbf{Modular \acs{llm}}~\cite{gandarela2026modular} is the closest faithful port of ReAct/Reflexion/Voyager/ExpeL to mechanism synthesis (feature-surface mapping in App.\ Table~\ref{tab:baseline_features}); it tests the rival hypothesis that workflow-role decomposition, assigning different LLM calls to planning, execution, and reflection, is sufficient and reasoning-mode decomposition adds nothing beyond that.
\textbf{R-APS ablations} (\emph{NoAdvMeta}: no adversarial testing, no meta-learning; \emph{NoSelRef}: no selective refinement) test the specific mechanism claim: that each component is necessary for its assigned failure axis and does not substitute for the others.
LLM are Llama-3.3-70B (general-purpose), Qwen3-4B (compact, reasoning-specialized), and Qwen3-30B-A3B (\ac{moe}), chosen to separate protocol-structure effects from model-scale effects.

\subsection{Q0: Ablation Effects Do Not Cross-Contaminate}
\label{sec:q0_decomposition}

\textbf{Removing each component degrades exactly one failure axis and leaves the other within noise} (Table~\ref{tab:ablation}).
Full R-APS attains the tightest adherence (\RAPSFullDeltaBars{} $\Delta$Bars) and the lowest robustness score (\RAPSFullRob{} $\rho_{\text{sens}}$).
Removing the adversary+meta pair worsens robustness $\RobustnessSpeedup{}\!\times$ (\RAPSFullRob{} $\to$ \RAPSNoAdvMetaRob{}) but leaves adherence within noise of Full; removing selective refinement worsens adherence $\AdhDegradationPct{}\%$ (\RAPSFullDeltaBars{} $\to$ \RAPSNoSelRefDeltaBars{}) but leaves robustness within noise.
The non-overlap is the empirical signature predicted by the prescriptive mode-failure pairing (Appendix Table~\ref{tab:decomposition_map}) and inconsistent with a single shared mechanism.

\textbf{Modular \acs{llm} collapses three components and exhibits three failure modes simultaneously}: on the modular-comparable shapes it reaches Norm.\ Dist.\ Index \ModularNDIComp{} and \ModularDeltaBars{} $\Delta$Bars with no robustness certificate (Tables~\ref{tab:adherence_distance}; baseline-feature surface in Appendix Table~\ref{tab:baseline_features}).
All three R-APS variants beat Modular on the distance index, but adherence does not follow the same ranking: NoAdvMeta (\RAPSNoAdvMetaDeltaBarsComp{}) does not beat Modular's \ModularDeltaBars{} because the adversary+meta pair supplies the bar-count-constraining heuristics (the multi-objective signature examined in \S\ref{sec:q2_counterfactual}).

\subsection{Q1: Most failures are absorbed at cheap early critics, and selective refinement rescues a meaningful fraction before expensive robustness checks are needed.}
\label{sec:q1_intrastage}

\textbf{Typed critics concentrate failures at the cheapest stage; selective refinement converts most failed iterations into successful archive admissions} (Tables~\ref{tab:qual_evidence},~\ref{tab:adherence_distance}).
Specifically, \QualTopologyPct{} of failures are absorbed at the topology critic, with only \QualRobustPct{} reaching the expensive counterfactual screen; selective refinement then converts \QualSuccessAfterFailPct{} of those failed iterations into archive admissions, yielding the tightest structural adherence (\RAPSFullDeltaBarsComp{} $\Delta$Bars) of any tested method while uniquely supplying a robustness certificate.
Per-stage refinement budgets diverge sharply by critic identity (Table~\ref{tab:qual_evidence}; App. Table~\ref{tab:refinement}), confirming \emph{routed} rather than uniform retry.

\begin{table}[t]
\centering\small
\begin{minipage}[t]{0.38\textwidth}
  \centering
  \begin{tabular}{@{}lr@{}}
    \toprule
    Episode bucket & Iterations-to-first-success \\
    \midrule
    Episode 1      & $4.9 \pm 0.5$ \\
    Episodes 2--3  & $3.9 \pm 0.3$ \\
    Episodes 4+    & $2.7 \pm 0.1$ \\
    \bottomrule
  \end{tabular}
  \captionof{table}{Meta-inductive acceleration across episodes on the same shape (baseline preset). Acceleration Ep1$\to$Ep4+: $46\%$, without parameter updates.}
  \label{tab:metalearning_acceleration}
\end{minipage}
\hfill
\begin{minipage}[t]{0.56\textwidth}
  \centering
  \begin{tabular}{@{}lrr@{}}
    \toprule
    \textbf{Failed Stage} & \textbf{Share} & \textbf{Refines/fail} \\
    \midrule
    Topology       & $59.2\%$ & $0.00$ \\
    Param.         & $31.2\%$ & $1.30$ \\
    Optimization   & $7.2\%$  & $1.96$ \\
    Robustness     & $2.4\%$  & $0.66$ \\
    \midrule
    \textbf{Total} & $100.0\%$ & $0.56$ \\
    \bottomrule
  \end{tabular}
  \captionof{table}{Failure localization by stage across the R-APS corpus (32~shapes). The cascade absorbs \QualTopologyPct{} of failures at the cheapest topology critic; only \QualRobustPct{} reach the expensive adversarial screen. Refines/fail rising from the parameter stage confirms selective refinement reuses valid upstream decisions. Per-shape detail in Appendix Table~\ref{tab:qual_evidence_full}.}
  \label{tab:qual_evidence}
\end{minipage}
\end{table}

\textbf{Full R-APS achieves the best distance and adherence simultaneously, while uniquely supplying a robustness evaluation} (Table~\ref{tab:adherence_distance}).
It attains $\Delta$Bars=\RAPSFullDeltaBarsComp{} (best) and Norm.\ Dist.\ Index=\RAPSFullNDIComp{} (best) at once.
Within R-APS variants on the all-shapes pool, NoAdvMeta gives the worst adherence despite a lower raw distance: without robustness as an active Pareto axis, the optimizer compresses distance further but drifts on bar count (the multi-objective trade-off made explicit in \S\ref{sec:q2_counterfactual}).
The cascade's failure mass aligns with the decomposition map (\ReasoningModeAbductivePct{} abductive, \ReasoningModeCorrectivePct{} corrective, \ReasoningModeCounterfactualPct{} counterfactual; App.~\ref{app:decomposition_lens}).
\begin{table}[t]
\centering
\small
\begin{tabular}{@{}l r@{\;\;}r r@{}}
\toprule
Configuration & \textbf{$\Delta$Bars\,$\downarrow$} & \textbf{Norm. Dist. Index\,$\downarrow$} & \textbf{Robustness\,$\downarrow$} \\
\midrule
Modular (Overall) & $2.0 \pm 0.0$ & $68.3 \pm 6.6$ & \textcolor{gray}{---} \\
\midrule
R-APS (w/o Adv+Meta)$^{\dagger}$ & $2.4 \pm 0.2$ & $61.0 \pm 7.1$ & $0.458 \pm 0.078$ \\
R-APS (w/o Sel.\ Ref.) & $1.6 \pm 0.3$ & $60.3 \pm 47.5$ & \cellcolor{bestcell} \textbf{$0.116 \pm 0.021$} \\
R-APS (Full) & \cellcolor{bestcell} \textbf{$1.4 \pm 0.2$} & \cellcolor{bestcell} \textbf{$41.2 \pm 11.3$} & $0.132 \pm 0.019$ \\
\bottomrule
\end{tabular}
\caption{Structural adherence and trajectory accuracy across modular LLM baselines and R-APS variants (all $\downarrow$ better). Parabola is excluded here because its raw distance is $\sim$100$\times$ larger than other shapes and would dominate row-level aggregates; it is broken out in Table~\ref{tab:per_shape_distance}. Robustness is computed over all available R-APS shapes; modular baselines lack robustness testing. Norm.\ Dist.\ Index uses the same canonical modular-baseline median as Tables~\ref{tab:ablation} and~\ref{tab:backbone_comparison}, so values are directly comparable across tables. $^{\dagger}$R-APS (w/o Adv+Meta) ran on the alphabet-plus-parabola subset; the index base is unchanged, so the row is level-comparable despite the different shape mix. Best per column is \textbf{highlighted}.}
\label{tab:adherence_distance}
\end{table}

\subsection{Q2: The adversary+meta pair is what gives R-APS robust worst-case behavior.}
\label{sec:q2_counterfactual}

\begin{figure}[t]
\centering
\includegraphics[width=0.62\textwidth]{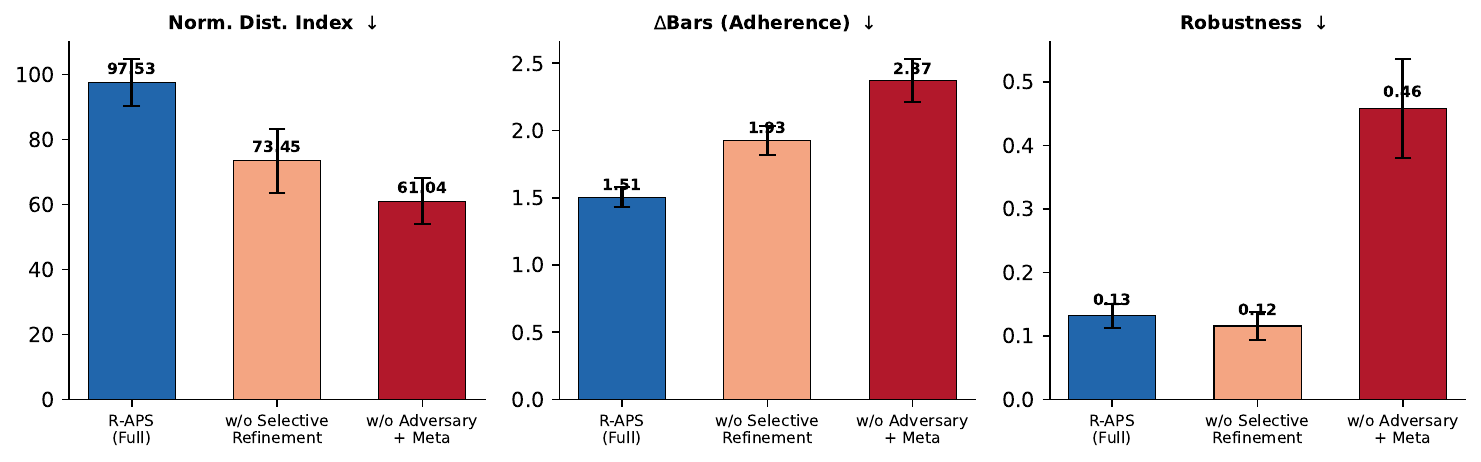}
\caption{Ablation by metric (lower is better); effects do not overlap across timescales.}
\label{fig:ablation_bars}
\end{figure}

\textbf{Robustness is owned by the intra-episode mode; structural adherence by the intra-stage one, the two axes do not interfere} (Table~\ref{tab:ablation}, Fig.~\ref{fig:ablation_bars}).
Removing the counterfactual+meta-inductive pair inflates $\rho_{\text{sens}}$ by $\RobustnessSpeedup{}\!\times$, the largest single effect in the ablation, while removing selective refinement leaves robustness within noise.
The adversary discovers worst-case failure modes that nominal evaluation hides, with nominal-vs-worst gaps of \AdversarialGapMedian{} median to \AdversarialGapMax{} max on the top-5 most fragile designs (all ablation-induced shifts significant at $p < 10^{-4}$, Mann--Whitney $U$; App.~\ref{app:adversarial_disc},~\ref{app:significance}).

\begin{table}[t]
\centering
\small
\begin{tabular}{@{}l rrr@{}}
\toprule
Configuration & \textbf{Norm. Dist. Index\,$\downarrow$} & \textbf{$\Delta$Bars\,$\downarrow$} & \textbf{Robustness\,$\downarrow$} \\
\midrule
R-APS (Full) & $97.5 \pm 7.2$ & \cellcolor{bestcell} \textbf{$1.5 \pm 0.1$} & $0.132 \pm 0.019$ \\
w/o Selective Refinement & $73.4 \pm 9.8$ \textcolor{green!60!black}{$\downarrow$}\scriptsize{-25\%} & $1.9 \pm 0.1$ \textcolor{red}{$\uparrow$}\scriptsize{+28\%} & \cellcolor{bestcell} \textbf{$0.116 \pm 0.021$ \textcolor{green!60!black}{$\downarrow$}\scriptsize{-12\%}} \\
w/o Adversary + Meta-Learning & \cellcolor{bestcell} \textbf{$61.0 \pm 7.1$ \textcolor{green!60!black}{$\downarrow$}\scriptsize{-37\%}} & $2.4 \pm 0.2$ \textcolor{red}{$\uparrow$}\scriptsize{+58\%} & $0.458 \pm 0.078$ \textcolor{red}{$\uparrow$}\scriptsize{+247\%} \\
\bottomrule
\end{tabular}
\caption{Ablation study: contribution of individual R-APS components. Each row removes one component; arrows show change relative to Full R-APS ($\downarrow$\,=\,better for all three metrics). Removing adversary\,+\,meta-learning degrades robustness $\mathbf{3.7\times}$ while leaving adherence within noise; removing selective refinement degrades bar adherence while leaving robustness within noise. The non-overlap is the empirical signature of genuine reasoning-mode decomposition. Norm.\ Dist.\ Index is defined in Section~\ref{sec:setup}; this pool includes parabola (within-R-APS ranking), whereas Table~\ref{tab:adherence_distance} uses the modular-comparable subset (cross-baseline ranking).}
\label{tab:ablation}
\end{table}

\subsection{Q3: R-APS explicitly invalidates stale rules, allowing it to improve across episodes without parameter updates.}
\label{sec:q3_metainductive}

\textbf{R-APS is the only system in our comparison surface that explicitly invalidates prior rules with cited counterexamples} (App.\ Table~\ref{tab:baseline_features}; representative invalidations in App.\ Table~\ref{tab:invalidated_heuristics}).
Of the \HeuristicTotalInval{} meaningful invalidations, \HeuristicCrossShapeInval{} are cross-shape, each citing the specific design and shape that refutes the rule, distinguishing R-APS from Voyager/ExpeL-style libraries that accumulate monotonically and discard the corrective signal of failed trajectories.

\textbf{On repeat shapes, R-APS accelerates to first archive admission across episodes without any parameter updates} (Table~\ref{tab:metalearning_acceleration}; per-episode trajectory in App.\ Fig.~\ref{fig:metalearning_episodes}).
Iterations-to-first-archive-admission decline from \AvgIterationsEpOne{} to \AvgIterationsEpFourPlus{}, a $\MetaLearningAcceleration$ reduction.

\subsection{Q4: R-APS either matches or beats the strongest classical baseline while additionally certifying robustness.}
\label{sec:q4_classical}

\textbf{R-APS matches or beats the strongest Enum+\acs{ga} configuration (6-bar, 60/300 budget) on \emph{both} shape groups: \EnumGAStdSpeedup{}$\times$ lower mean Chamfer on the \EnumGAShapesStandard{} standard curves and \EnumGALtrSpeedup{}$\times$ lower on the \EnumGAShapesLetters{} alphabet letters, despite discovering topology from scratch}, while additionally supplying a robustness certificate Enum+\acs{ga} does not (Table~\ref{tab:enum_ga_summary}; per-shape distance figure in App.~\ref{app:per_shape_enumga}, Fig.~\ref{fig:enum_ga_comparison}).
The improvement is largest on shapes where topology discovery dominates (Circle \EnumGACircleSpeedup{}$\times$, NACA \EnumGANACASpeedup{}$\times$); the alphabet group still favours R-APS but by a narrower margin, consistent with letter shapes admitting more than one near-optimal topology.

\begin{table}[t]
\centering
\small
\begin{tabular}{@{}l c rrr@{}}
\toprule
Shape group & \textbf{\#shapes} & \textbf{Enum+GA 4-bar\,$\downarrow$} & \textbf{Enum+GA 6-bar\,$\downarrow$} & \textbf{R-APS (Full)\,$\downarrow$} \\
\midrule
Standard curves & $5$ & $7.49$ & $5.99$ & \cellcolor{bestcell} \textbf{$1.83$} \\
Alphabet letters & $26$ & $4.53$ & $4.06$ & \cellcolor{bestcell} \textbf{$2.13$} \\
\midrule
\textbf{All} & $31$ & $5.00$ & $4.37$ & \cellcolor{bestcell} \textbf{$2.08$} \\
\bottomrule
\end{tabular}
\caption{R-APS vs.\ classical Enum+GA aggregated by shape group (mean Chamfer distance, $\downarrow$ better). We report both the 4-bar family (matching R-APS's bar-count spec, apples-to-apples) and the stronger 6-bar family (more topology freedom). R-APS jointly optimises distance, adherence, and robustness; Enum+GA optimises trajectory error only and produces no robustness certificate. R-APS achieves \EnumGAStdSpeedup{}$\times$ lower mean distance than the stronger 6-bar baseline on standard curves and \EnumGALtrSpeedup{}$\times$ lower on the alphabet group. Per-shape detail in Appendix~\ref{app:additional_quant_results}.}
\label{tab:enum_ga_summary}
\end{table}

\subsection{Discussion}
\label{sec:discussion}

The umbrella claim (\textbf{Q0}) is supported by the non-overlap of the two ablation effects; each failure mode is independently addressed by its corresponding timescale (Q1: \QualTopologyPct{} of failures absorbed at the cheapest stage with diagnosed routing; Q2: $\RobustnessSpeedup{}\!\times$ tighter robustness certificate; Q3: \InvalidatedRuleCount{} explicit invalidations with \HeuristicCrossShapeInval{}/\HeuristicTotalInval{} cross-shape refutations within-preset acceleration); Q4 establishes external competitiveness, and the model-scale study (App.~\ref{app:backbone}) shows 4B reasoning-specialized models competitive with 70B general-purpose ones, evidence that protocol structure can partially offset model scale.
The three timescales are parameterized by domain interfaces, not mechanism-design knowledge: typed validation critics need cheap-to-expensive verification cascades, the counterfactual mode needs a perturbation interface plus a sensitivity primitive, the meta-inductive mode needs a refinement-trajectory log with explicit invalidation semantics; \acs{sql} synthesis~\cite{mohr2026reflective}, circuit design, robot motion planning, and structural engineering are natural targets for the same protocol.

\section{Related Work}
\label{sec:related}

\paragraph{Agentic AI, multi-agent models, and reasoning-mode decomposition.}
Recent work frames \acp{llm} as autonomous agents that plan, invoke tools, and act over extended horizons~\cite{xi2023rise,wang2024survey,schick2023toolformer,park2023generative}; frameworks such as CAMEL~\cite{li2023camel}, AutoGen~\cite{wu2023autogen}, MetaGPT~\cite{hong2023metagpt}, and ChatDev~\cite{qian2024chatdev} specialize agents by \emph{workflow role}.
R-APS instead decomposes by \emph{reasoning mode} (Section~\ref{sec:modes}) and addresses the three canonical agentic reliability challenges jointly~\cite{yao2023react,shinn2024reflexion,dalrymple2024towards,wang2023voyager,zhao2024expel}: typed validation critics enable in-loop error recovery, counterfactual stress-testing certifies worst-case behavior before archive admission, and the meta-inductive lifecycle accumulates knowledge with explicit invalidation, without human intervention between episodes.

\paragraph{\acs{llm}-based optimization, in-context learning, and sensitivity priors.}
OPRO~\cite{yang2024opro}, EvoPrompting~\cite{chen2024evoprompting}, FunSearch~\cite{romera2024funsearch}, \ac{cot}~\cite{wei2022cot}, ReAct~\cite{yao2023react}, Tree-of-Thoughts~\cite{yao2024tot}, and Reflexion~\cite{shinn2024reflexion} treat each generation atomically, discarding the whole solution on any violation without locating which stage failed; R-APS uses staged compositional reasoning (typed, not Lake-style~\cite{lake2017building}) with \emph{typed validation critics} and selective refinement, paralleling \citet{mohr2026reflective} in text-to-\acs{sql}.
\Ac{icl}~\cite{brown2020gpt3}, Voyager~\cite{wang2023voyager}, and ExpeL~\cite{zhao2024expel} accumulate skills monotonically from successes; R-APS adds an inter-episode meta-inductive mode with explicit invalidation that learns from failures too, using Sobol-guided context accumulation as its in-spirit analogue of meta-learning~\cite{finn2017maml}.

\paragraph{AI for engineering design.}
Classical mechanism synthesis~\cite{sandor1984advanced}, evolutionary optimization~\cite{cabrera2002genetic}, neural~\cite{mcgreggor2023neural}, and \acs{llm}-based~\cite{makatura2024cadllm,liang2024llmrobot} methods use end-to-end or single-objective search; closest to our domain, \citet{gandarela2025controlled} pair a Designer with a Critic for iterative mechanism refinement but lack reasoning-mode separation, counterfactual stress-testing, and accuracy/robustness Pareto co-optimization.

\section{Conclusion}
\label{sec:conclusion}

R-APS introduces \emph{reasoning-mode decomposition} as a design principle for agentic AI methods on constrained-design tasks: abductive, counterfactual, meta-inductive, corrective, and inductive reasoning optimize in incompatible cognitive directions and therefore demand separate contexts within the agentic loop.
Instantiated as three interacting timescales, intra-stage typed validation critics, intra-episode counterfactual stress-testing as a Pareto objective, and an inter-episode meta-inductive lifecycle with explicit invalidation as long-term memory, this single principle addresses three otherwise-coupled structural failures jointly, with the underlying \ac{llm} entirely frozen.
On \TotalShapes{} shapes R-APS attains a $\EnumGAOverallSpeedup{}\times$ mean Chamfer-distance reduction over Enum+\acs{ga} and a \MetaLearningAcceleration{} inter-episode acceleration; ablations cleanly separate robustness ($\RobustnessSpeedup{}\times$) and adherence ($\AdhDegradationPct{}\%$) effects along the predicted mode boundaries with no cross-contamination, the falsifiable empirical signature of genuine decomposition.
Beyond the headline numbers, 4B reasoning-specialized backbones become competitive with 70B general-purpose ones inside R-APS's agentic loop, suggesting that \emph{how} an agentic model organizes reasoning matters alongside model scale.
\textbf{Limitations.}
The reasoning-mode separation introduces additional inference calls and Sobol sampling relative to monolithic generation; the full compute profile is in App.~\ref{app:backbone}.
The meta-inductive memory operates at the pace of counter-evidence accumulation.
\textbf{Broader impact.}
R-APS is design-assistance: typed critics and explicit invalidation produce inspectable, overridable artifacts that an engineer audits, not a fully autonomous agent shipping designs to manufacturing without review, and the principles demonstrated (localized failure diagnosis, worst-case robustness, self-invalidating memory) offer a blueprint for agentic models that are both capable and auditable.

\bibliographystyle{plainnat}
\bibliography{aps_refs}

\appendix

\section{R-APS Pipeline Algorithm}
\label{app:raps_alg}

\begin{algorithm}[h]
\caption{R-APS: Reflective Adversarial Pareto Search}
\begin{algorithmic}[1]
\State \textbf{Input:} Archive $A{=}\emptyset$, budget $B$, target $\mathcal{T}_{\text{target}}$, heuristics $\mathcal{H}{=}\emptyset$
\For{iteration $t = 1, \dots, B$}
  \State \textcolor{gray}{\textsc{// Stage 0: Critic Target Selection}}
  \State $\mathbf{w} \leftarrow 0.7\,\pi_C.\textsc{SelectTarget}(A, \mathcal{H}) + 0.3\,\textsc{GapAnalysis}(A)$
  \State \textcolor{gray}{\textsc{// Stage 1: Topology \& Parameter Proposal}}
  \State $\{(\tau_i, \theta_{0,i}, s_i)\} \leftarrow \pi_D.\textsc{Propose}(A, \mathbf{w}, \mathcal{T}_{\text{target}}, \mathcal{H})$
  \For{each proposal $i$}
    \State \textcolor{gray}{\textsc{// Critic 1: Structural Validation}}
    \State $M_i^0 \leftarrow \textsc{Assemble}(\tau_i, \theta_{0,i}, \mathcal{C})$
    \If{$\neg\textsc{CheckStructural}(M_i^0)$}
      \State \textbf{goto } \textsc{SelectiveRefinement}(\textsc{Topology})
    \EndIf
    \State \textcolor{gray}{\textsc{// Stage 3: Constructive Optimization}}
    \State $\theta_i^* \leftarrow \mathcal{O}.\textsc{Optimize}(\tau_i, \theta_{0,i}, s_i, \mathbf{w})$
    \State \textcolor{gray}{\textsc{// Critic 2: Intent Validation}}
    \If{$\neg\textsc{ValidateIntent}(\mathcal{T}(M_i), \mathcal{T}_{\text{target}})$}
      \State \textbf{goto } \textsc{SelectiveRefinement}($d_i$) \Comment{$d_i \in \{\textsc{Topology}, \textsc{Optim}\}$}
    \EndIf
  \EndFor
  \State \textcolor{gray}{\textsc{// Stage 4: Adversarial Robustness Testing (Alg.~\ref{alg:adversarial})}}
  \For{each candidate $M_i$}
    \State $(\delta_i^*, \mathbf{f}_{\text{worst},i}) \leftarrow \textsc{AdversarialTest}(M_i, \theta_i^*)$
  \EndFor
  \State \textcolor{gray}{\textsc{// Stage 5: Post-Opt Critique \& Archive Update}}
  \For{each candidate $M_i$}
    \State \textbf{if} $\pi_{PC}.\textsc{Critique}(M_i) = \textsc{Refine}$ \textbf{then} $M_i \leftarrow \pi_R.\textsc{Refine}(M_i)$
    \State Update Pareto archive $A$ with $\bar{\mathbf{f}}_i = (f_1(M_i), \rho_i)$
  \EndFor
  \State $\mathcal{H} \leftarrow \textsc{MetaLearn}(A, \mathcal{H})$ \Comment{Section~\ref{sec:metalearning}}
\EndFor
\State \textbf{return} $A, \mathcal{H}$
\end{algorithmic}
\label{alg:raps}
\end{algorithm}

\section{Agent Architecture Details}
\label{app:agents}

\begin{table}[t]
\centering
\small
\setlength{\tabcolsep}{4pt}
\renewcommand{\arraystretch}{1.15}
\begin{tabular}{@{}llllp{3.1cm}@{}}
\toprule
\textbf{Reasoning mode} & \textbf{Cognitive operation} & \textbf{Agent} & \textbf{Timescale} & \textbf{Incompatible with (if merged)} \\
\midrule
Abductive      & Hypothesis generation              & Designer $\pi_D$      & Intra-stage    & Counterfactual, meta-inductive \\
Counterfactual & Worst-case stress-testing          & Adversary (\S\ref{sec:adversarial}) & Intra-episode  & Abductive                    \\
Meta-inductive & Pattern over prior attempts        & Critic $\pi_C$        & Intra-episode  & Abductive, corrective        \\
Corrective     & Localized repair under constraints & Refinement $\pi_R$    & Intra-stage    & Meta-inductive               \\
Inductive      & Rule distillation (incl.\ failures)& Meta-Analyst MA       & Inter-episode  & Evaluative (single-case)     \\
\bottomrule
\end{tabular}
\caption{The five reasoning modes R-APS decomposes into distinct cognitive contexts. Each row names the mode, its cognitive operation, the agent that realizes it, the timescale it lives on, and the partner mode(s) whose cognitive direction is incompatible with it if collapsed into a shared prompt. Full agent inputs/outputs appear in Appendix Table~\ref{tab:reasoning_type}.}
\label{tab:reasoning_modes}
\end{table}

\IfFileExists{tables/decomposition_map.tex}{
\begin{table}[t]
\centering\footnotesize
\setlength{\tabcolsep}{3pt}
\renewcommand{\arraystretch}{1.15}
\begin{tabular}{@{}llcccccl@{}}
\toprule
\textbf{Stage} & \textbf{Agent} & \textbf{Abd.} & \textbf{C.fact.} & \textbf{Meta-ind.} & \textbf{Corr.} & \textbf{Ind.} & \textbf{Addresses failure mode} \\
\midrule
STAGE1 Proposal        & Designer         & $\bullet$ &  &  &  &  & Hypothesis generation \\
Structural Validation  & Gate             &  &  &  &  &  & Failure localization \\
STAGE3 Optimization    & Optimizer        &  &  &  &  &  & -- \\
Post-opt Critique      & Critic           &  &  & $\bullet$ &  &  & No invalidation $\to$ meta-ind. \\
STAGE4 Robustness      & Adversary        &  & $\bullet$ &  &  &  & No robustness cert.\ $\to$ c.fact. \\
Design Refinement      & Refinement       &  &  &  & $\bullet$ &  & Failure propagation $\to$ corr. \\
Meta-learning          & Meta-Analyst     &  &  &  &  & $\bullet$ & Rule-distillation $\to$ ind. \\
\bottomrule
\end{tabular}
\caption{Stage-by-reasoning-mode responsibility map. Each pipeline stage produces at most one reasoning-mode output; no stage is responsible for two modes at once. The rightmost column names the documented failure mode each mode is responsible for preventing --- making the necessity argument explicit: counterfactual reasoning addresses the absence of robustness certification; corrective reasoning addresses failure propagation; meta-inductive + inductive modes address the absence of invalidation / rule-distillation. This architectural separation is what the paper calls \emph{reasoning-mode decomposition} and is the structural support for the claim that reliable constrained design requires mode-specialized contexts.}
\label{tab:decomposition_map}
\end{table}
}{}

This appendix extends the main-body discussion of the five reasoning modes (Section~\ref{sec:modes}, Table~\ref{tab:reasoning_modes}) with the full agent profile, inputs and outputs.
Recall that R-APS does not coordinate ``five agents'' as an engineering choice; it operationalizes the hypothesis that five distinct reasoning modes each require their own cognitive context, because any two of them collapsed into a shared prompt optimize in incompatible directions and degrade one another.

\begin{table}[h]
\centering
\small
\begin{tabular}{llll}
\hline
\textbf{Agent} & \textbf{Reasoning Mode} & \textbf{Input} & \textbf{Output} \\ \hline
\textbf{Designer ($\pi_D$)} & Abductive, analogical & $A, \mathbf{w}, \mathcal{T}_{\text{target}}, \mathcal{H}$ & $\tau, \theta_0$, rationale \\
\textbf{Critic ($\pi_C$)} & Meta-inductive, reflective & $A$, history, $\mathcal{H}$ & $\mathbf{w}$, recommendations \\
\textbf{Post-Opt Critic ($\pi_{PC}$)} & Evaluative, diagnostic & $M, \mathbf{f}_{\text{nom}}, \mathbf{f}_{\text{worst}}$, symbolic & Verdict, critique \\
\textbf{Refinement ($\pi_R$)} & Corrective, constructive & $M$, critique, symbolic, $\mathcal{T}_{\text{target}}$ & $\tau', \theta_0'$ \\
\textbf{Meta-Analyst (MA)} & Inductive, contrastive & Digest, failures, $\mathcal{H}$ & Updated $\mathcal{H}$ \\
\hline
\end{tabular}
\caption{Full agent reasoning profiles (extends main-body Table~\ref{tab:reasoning_modes}). $A$: archive, $\mathbf{w}$: weight vector, $\mathcal{H}$: learned heuristics. The Post-Opt Critic and Refinement agents implement a critique-then-correct loop after robustness testing.}
\label{tab:reasoning_type}
\end{table}

\paragraph{Computational Tools.}
Agents interact with two deterministic tools: (i)~a \emph{kinematic simulator} $\mathcal{S}$ that computes forward kinematics and end-effector trajectories, providing ground truth since LLMs cannot reliably solve nonlinear kinematic equations; and (ii)~a \emph{numerical optimizer} $\mathcal{O}$ operating in constructive mode (maximize performance) or destructive mode (find worst-case perturbations), with strategies selected by the LLM (BFGS, PSO, or Grid).

\paragraph{Critic Target Selection (Two-Pass Architecture).}
The Critic Agent identifies underexplored regions of the Pareto frontier using a two-pass architecture.
Pass~1 (meta-analyst): evaluates Pareto quality, topology diversity, and coverage.
Pass~2 (strategist): produces a weight vector $\mathbf{w}_{\text{llm}}$ targeting the most promising objective trade-off region.
The final target blends LLM judgment with density-based gap analysis:
$\mathbf{w} = 0.7\,\mathbf{w}_{\text{llm}} + 0.3\,\mathbf{w}_{\text{gap}}$,
where $\mathbf{w}_{\text{gap}}$ identifies low-density cells in a discretized objective space histogram.

\paragraph{Post-Optimization Critique (Stage~5).}
The Post-Opt Critique agent $\pi_{PC}$ evaluates fully optimized and robustness-tested mechanisms across four dimensions: kinematic fidelity, structural soundness, adversarial robustness (informed by Sobol indices), and compositional coherence (consistency of the trajectory's qualitative motion signature with the target).
The verdict $v \in \{\textsc{Accept}, \textsc{Refine}\}$ with diagnostic rationale is passed to the Refinement agent when correction is needed.

\section{Typed Validation Critic Details}
\label{app:critics}

\paragraph{Critic~1: Structural Validation.}
After assembly, we verify that the mechanism satisfies hard kinematic constraints: link count matches the target, Grashof mobility conditions hold, and the mechanism can assemble.
Failure at this critic indicates a topology-level error.

\paragraph{Critic~2: Intent Validation.}
After optimization, we verify trajectory-specification match using two checks.
First, quantitative error thresholds:
\begin{equation}
d = \begin{cases}
\textsc{Topology} & \text{if } f_1(M) > 10 \cdot \epsilon_{\text{traj}} \quad \text{(fundamental mismatch)} \\
\textsc{Optimization} & \text{if } f_1(M) > \epsilon_{\text{traj}} \quad \text{(convergence failure)} \\
\textsc{None} & \text{otherwise}
\end{cases}
\end{equation}

Second, \emph{shape adherence validation} via motion primitive frequency matching.
A qualitative signature $\sigma(\mathcal{T})$ is computed from velocity, curvature, and heading, discretized into states $\mathcal{S} = \{\text{G}, \text{S}, \text{ST}, \text{VS}\}$.
The frequency distribution $\mathbf{p}(\sigma)$ is compared against a reference $\mathbf{p}^*$ using $L_1$ distance:
$d_{\text{shape}} = \|\mathbf{p}(\sigma(\mathcal{T}(M))) - \mathbf{p}^*\|_1$.
Designs exceeding the shape adherence threshold are rejected even if their Chamfer distance is low.

\paragraph{Selective Refinement Actions.}
Table~\ref{tab:refinement} lists the action taken when a typed validation critic diagnoses a failure, and which prior decisions are preserved.
The four diagnoses produced by the pipeline (\textsc{Topology}, \textsc{Optimization}, \textsc{Robustness/topology}, \textsc{Robustness/param}) follow directly from the \texttt{FailureStage} enum in the workflow code: a \textsc{Robustness} failure is the only diagnosis sub-typed by an additional fingerprint field (\texttt{failure\_fingerprint.failure\_type}) emitted by the adversary, distinguishing the case where the nominal trajectory is itself off (treat as \textsc{Topology}) from the case where the design is sensitive to small perturbations (re-optimize parameters with the fragile-parameter set surfaced to the optimizer).
The dual-branch routing of \textsc{Robustness} is the critic C3 in Fig.~\ref{fig:raps_overview}; mechanically, it is implemented in the \texttt{\_selective\_refinement} method of the workflow, where the \texttt{ROBUSTNESS} branch reads the fingerprint and dispatches to either the topology re-proposal path or the Stage-3 re-optimization path while preserving the validated topology.

\section{Adversarial Pipeline Algorithm and Illustration}
\label{app:adversarial_alg}
\label{app:adversarial_figs}

\begin{algorithm}[h]
\caption{Sensitivity-Guided Adversarial Testing}
\begin{algorithmic}[1]
\State \textbf{Input:} Mechanism $M$, nominal $\theta^*$, scale $\epsilon$, budget $n$
\State \textcolor{gray}{\textsc{// Phase 1: Sobol Sensitivity Screening}}
\State $\Delta_{\text{LHS}} \leftarrow \textsc{LatinHypercube}(\max(2d, \lfloor n\rho \rfloor), d) \cdot 2\epsilon - \epsilon$
\State Evaluate all perturbations; compute Sobol first-order indices $\{S_i\}$
\State Active set: $\mathcal{A} \leftarrow \{i : \sum_{j \leq i} S_{(j)} < 0.9\}$
\State \textcolor{gray}{\textsc{// Phase 2: Directed Adversarial Sampling}}
\State $\boldsymbol{\mu} \leftarrow \text{mean}(\text{top-quartile perturbations})$
\For{remaining budget}
  \State $\delta_j[\mathcal{A}] \sim \mathcal{N}(\boldsymbol{\mu}[\mathcal{A}], \text{diag}(\boldsymbol{\sigma}[\mathcal{A}]^2))$; $\delta_j[\bar{\mathcal{A}}] = 0$
  \State Evaluate; update worst-case $\delta^*$ if worse
\EndFor
\State \textbf{return} $\delta^*, \mathbf{f}_{\text{worst}}, \{S_i\}$, failure fingerprint
\end{algorithmic}
\label{alg:adversarial}
\end{algorithm}

\begin{table}[h]
\centering
\small
\begin{tabular}{lll}
\hline
\textbf{Diagnosis} & \textbf{Action} & \textbf{Preserved} \\ \hline
\textsc{Topology} & Try next proposal; if exhausted, re-invoke Stage 0--1 & Nothing \\
\textsc{Optimization} & Switch strategy (BFGS$\leftrightarrow$PSO$\leftrightarrow$Grid); re-run Stage 3 & $\tau, \theta_0$ \\
\textsc{Robustness}/topology & Treat as topology failure & Nothing \\
\textsc{Robustness}/param & Re-optimize with fragile-parameter knowledge & $\tau$ \\
\hline
\end{tabular}
\caption{Selective refinement actions by failure diagnosis.}
\label{tab:refinement}
\end{table}

\begin{figure}[h]
\centering
\begin{tikzpicture}[
  io/.style={draw, rounded corners=4pt, minimum width=2.4cm,
             minimum height=0.75cm, font=\small, align=center},
  st/.style={draw, rounded corners=4pt, minimum width=2.6cm,
             minimum height=0.75cm, font=\small\bfseries, align=center},
  arr/.style={-Stealth, thick}
]
  \node[io, fill=gray!12] (inp) {$\theta^*$, $\tau$\\budget $N$, tol.\ $\varepsilon$};
  \node[st, fill=stage1!15, draw=stage1, right=0.25cm of inp] (s1)
    {\stageone{Phase 1}\\LHS + Sobol};
  \node[st, fill=stage2!15, draw=stage2, right=0.7cm of s1] (s2)
    {\stagetwo{Phase 2}\\Directed MC};
  \node[st, fill=stage3!15, draw=stage3, right=0.7cm of s2] (s3)
    {\stagethree{Phase 3}\\Attribution};
  \node[io, fill=outputC!15, draw=outputC, right=0.3cm of s3] (out)
    {\out{Fingerprint}\\$\delta^*$, $\{S_i\}$, type};

  \draw[arr] (inp) -- (s1);
  \draw[arr] (s1) -- node[above, font=\scriptsize]{top params} (s2);
  \draw[arr] (s2) -- node[above, font=\scriptsize]{$\delta^*$} (s3);
  \draw[arr] (s3) -- (out);
\end{tikzpicture}
\caption{End-to-end adversarial robustness pipeline.}
\label{fig:adversarial_pipeline}
\end{figure}
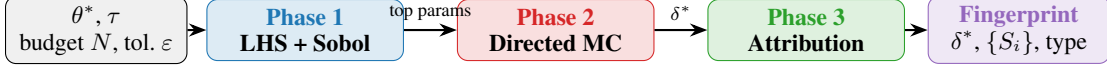

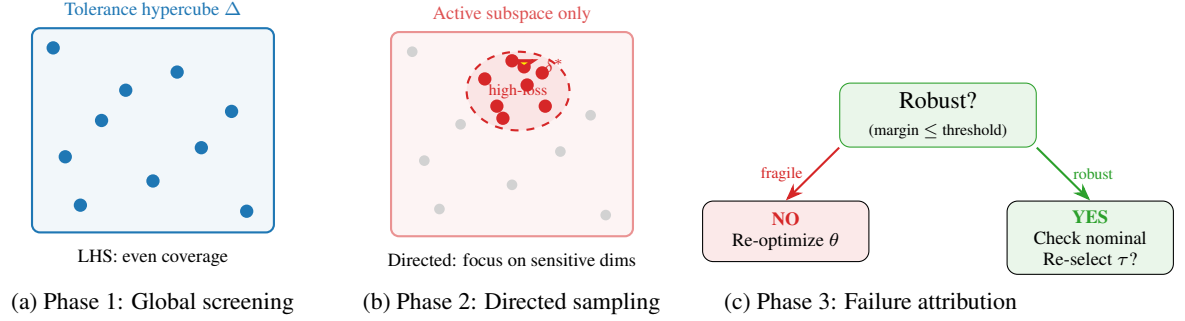
\begin{figure}[h]
\centering
\begin{subfigure}[t]{0.32\textwidth}
  \centering
  \begin{tikzpicture}
    \draw[draw=stage1, fill=stage1!6, thick, rounded corners=3pt]
      (0,0) rectangle (3.2, 2.7);
    \node[font=\scriptsize, text=stage1, above] at (1.6, 2.7)
      {Tolerance hypercube $\Delta$};
    \foreach \x/\y in {
      0.28/2.44, 0.64/0.36, 0.92/1.48, 1.24/1.88,
      1.60/0.68, 1.92/2.12, 2.24/1.12, 2.64/1.60,
      0.44/1.00, 2.84/0.28}
      \fill[stage1] (\x,\y) circle (2.5pt);
    \node[font=\scriptsize, below] at (1.6, -0.1)
      {LHS: even coverage};
  \end{tikzpicture}
  \caption{Phase 1: Global screening}
  \label{fig:mc_phase1}
\end{subfigure}
\hfill
\begin{subfigure}[t]{0.32\textwidth}
  \centering
  \begin{tikzpicture}
    \draw[draw=stage2!50, fill=stage2!5, thick, rounded corners=3pt]
      (0,0) rectangle (3.2, 2.7);
    \node[font=\scriptsize, text=stage2!80, above] at (1.6, 2.7)
      {Active subspace only};
    \foreach \x/\y in {
      0.28/2.44, 0.64/0.36, 0.92/1.48, 1.24/1.88,
      1.60/0.68, 1.92/2.12, 2.24/1.12, 2.64/1.60,
      0.44/1.00, 2.84/0.28}
      \fill[gray!35] (\x,\y) circle (2pt);
    \draw[stage2, thick, dashed, fill=stage2!12]
      (1.68, 1.92) ellipse (0.68cm and 0.52cm);
    \node[font=\tiny, text=stage2] at (1.68, 1.92) {high-loss};
    \foreach \x/\y in {
      1.24/2.08, 1.48/1.56, 1.76/2.24, 2.04/1.72,
      1.80/2.00, 1.40/1.72, 2.00/2.16, 1.60/2.32}
      \fill[stage2] (\x,\y) circle (2.5pt);
    \draw[stage2, very thick, fill=yellow]
      (1.76, 2.24) -- ++(0.12, 0.08) -- ++(-0.24, 0) -- cycle;
    \node[font=\tiny, right=0.02cm, text=stage2] at (1.88, 2.26) {$\delta^*$};
    \node[font=\scriptsize, below] at (1.6, -0.1)
      {Directed: focus on sensitive dims};
  \end{tikzpicture}
  \caption{Phase 2: Directed sampling}
  \label{fig:mc_phase2}
\end{subfigure}
\hfill
\begin{subfigure}[t]{0.32\textwidth}
  \centering
  \begin{tikzpicture}[arr/.style={-Stealth, thick}]
    \node[draw=stage3, fill=stage3!10, rounded corners=4pt,
          minimum width=2.6cm, minimum height=0.7cm,
          align=center, font=\small] (q)
      {Robust?\\{\tiny (margin $\le$ threshold)}};
    \node[draw, fill=stage2!10, rounded corners=4pt,
          minimum width=2.2cm, minimum height=0.8cm,
          align=center, font=\scriptsize,
          below left=0.7cm and -0.4cm of q] (no)
      {\stagetwo{NO}\\Re-optimize $\theta$};
    \node[draw, fill=stage3!10, rounded corners=4pt,
          minimum width=2.2cm, minimum height=0.8cm,
          align=center, font=\scriptsize,
          below right=0.7cm and -0.4cm of q] (yes)
      {\stagethree{YES}\\Check nominal\\Re-select $\tau$?};
    \draw[arr, stage2] (q.south west) --
          node[left, font=\tiny, text=stage2]{fragile} (no.north);
    \draw[arr, stage3] (q.south east) --
          node[right, font=\tiny, text=stage3]{robust} (yes.north);
  \end{tikzpicture}
  \caption{Phase 3: Failure attribution}
  \label{fig:mc_phase3}
\end{subfigure}
\caption{Sensitivity-guided adversarial robustness testing phases. (a)~LHS for even coverage, computing Sobol indices. (b)~Directed sampling around high-loss centroid. (c)~Failure classification.}
\label{fig:mc_phases}
\end{figure}

\section{Meta-Learning Details}
\label{app:metalearning}

\subsection{Design Space Compression}

To enable efficient meta-reasoning over large archives, R-APS compresses the design space through three steps.

\paragraph{Feature Embedding.} Each mechanism $M_j$ is encoded as $\phi(M_j) \in \mathbb{R}^d$ concatenating topology features, parameter vector, trajectory fingerprint, performance objectives, and bar count.
UMAP projects these to 2D embeddings preserving local similarity and global structure.

\paragraph{Family Stratification.} Mechanisms are stratified by bar count into families $\mathcal{F}_n$, since bar count determines kinematic capability (4-bar: 6th-order curves; 6-bar: 8th-order).
Within each family, HDBSCAN identifies structural sub-families.

\paragraph{Representative Sampling.} For each sub-family, three exemplars are selected: centroid (typical), Pareto boundary (best-in-class), and novelty outlier.
These form a compact design digest for LLM consumption.

\subsection{Refinement History and Failure Mining}

R-APS stores complete refinement trajectories:
$\text{History}(M_j) = [(M_j^{(0)}, d^{(0)}, \text{action}^{(0)}), \dots, (M_j^{(k)}, \textsc{Success})]$.
Failed designs are clustered by $(\text{stage}_{\text{fail}}, \text{error\_type})$ to identify recurring problems.
Resolution patterns $\text{ResolvePattern}(\text{error}_i) = \{(\text{action}_j, \text{success\_rate}_j)\}$ track successful corrective actions.

\subsection{Policy Updates via Dynamic Prompting}

Agent policies are updated through context augmentation:
\begin{align}
\pi_D^{(t+1)} &= \text{LLM}\Big(\cdot \mid \mathcal{P}_D, \mathcal{H}^{(t+1)}, \{\text{Exemplar}_j\}, \textsc{ExcludeList}^{(t+1)}\Big) \\
\pi_C^{(t+1)} &= \text{LLM}\Big(\cdot \mid \mathcal{P}_C, \textsc{CoverageMap}(A^{(t)}), \textsc{GapRegions}^{(t)}, \mathcal{H}^{(t+1)}\Big)
\end{align}
The exclude list contains topologies with $\geq 2$ recorded failures.
The archive maintains the Pareto invariant via NSGA-II-style non-dominated sorting with crowding distance.

\section{Future Directions}
\label{app:future}

Incorporating physics-based simulation for dynamic performance would enable synthesis of mechanisms optimized for force transmission and energy efficiency.
Extending to spatial (3D) mechanisms and compliant mechanisms would test scalability to higher-dimensional design spaces.
Transfer learning across mechanism families (linkages to cam-follower or gear trains) is a natural next step.
Interactive multi-objective optimization integrating human designer preferences would enable R-APS as an augmentative tool leveraging both LLM reasoning and human expertise.


\section{Additional Quantitative Results}
\label{app:additional_quant_results}

This appendix provides per-shape detail and sensitivity analyses behind the main-text results in Section~\ref{sec:experiments}.
Each subsection opens with a one-sentence statement of which main-text question it deepens, so the appendix can be read forward from the corresponding main-text forward pointer.

\subsection{Baseline Feature Surface}
\label{app:baseline_features}
\emph{Deepens \S\ref{sec:setup} by mapping which baselines implement which components of R-APS, so the gap between R-APS and each baseline is anchored to a concrete missing component rather than an aggregate score difference.}

\begin{table}[t]
\centering
\small
\setlength{\tabcolsep}{4pt}
\begin{tabular}{@{}l c c c c c c@{}}
\toprule
System & \makecell{Role\\sep.} & \makecell{Retry on\\failure} & \makecell{Persistent\\memory} & \makecell{Robustness\\check} & \makecell{Explicit\\invalidation} & \makecell{Typed failure\\attribution} \\
\midrule
ReAct~\cite{yao2023react} & $\times$ & $\checkmark$ & $\times$ & $\times$ & $\times$ & $\times$ \\
Reflexion~\cite{shinn2024reflexion} & $\times$ & $\checkmark$ & $\checkmark$ & $\times$ & $\times$ & $\times$ \\
Voyager~\cite{wang2023voyager} & \textcolor{gray}{partial} & $\checkmark$ & $\checkmark$ & $\times$ & $\times$ & $\times$ \\
ExpeL~\cite{zhao2024expel} & $\times$ & $\checkmark$ & $\checkmark$ & $\times$ & $\times$ & $\times$ \\
Modular LLM (ours, port) & $\checkmark$ & $\checkmark$ & $\checkmark$ & $\times$ & $\times$ & $\times$ \\
\textbf{R-APS (ours)} & $\checkmark$ & $\checkmark$ & $\checkmark$ & $\checkmark$ & $\checkmark$ & $\checkmark$ \\
\bottomrule
\end{tabular}
\caption{Feature-parity matrix mapping cited LLM-agent baselines and our Modular LLM port to the capabilities R-APS asks for. ReAct, Reflexion, Voyager, and ExpeL target environments (text adventure, web, Minecraft) whose action spaces and reward signals do not transfer to constrained mechanism synthesis without a bespoke port; we do \emph{not} re-implement them on this domain. Our \emph{Modular LLM} baseline is the closest faithful port: its Designer/Critic/Planner/Refiner agents cover the role multiplicity of those systems; its Constraint-Learning (CL) flag implements Voyager/ExpeL-style monotonic memory accumulation; its Design-Review (DR) flag implements Reflexion-style verbal reflection on failed attempts. The three capabilities Modular LLM lacks, robustness check, explicit invalidation, and typed failure attribution, are exactly the components R-APS adds, so the gap between Modular LLM and R-APS is the empirical answer to whether those capabilities matter on this domain (Tables~\ref{tab:adherence_distance}, \ref{tab:ablation}).}
\label{tab:baseline_features}
\end{table}

\subsection{Meta-Learning Per-Episode Trajectory}
\label{app:metalearning_per_episode}
\emph{Deepens \S\ref{sec:q3_metainductive} (Q3) by visualizing the iterations-to-first-archive-admission trajectory across episode buckets that complements the aggregate Table~\ref{tab:metalearning_acceleration}.}

\begin{figure}[h]
\centering
\includegraphics[width=0.55\textwidth]{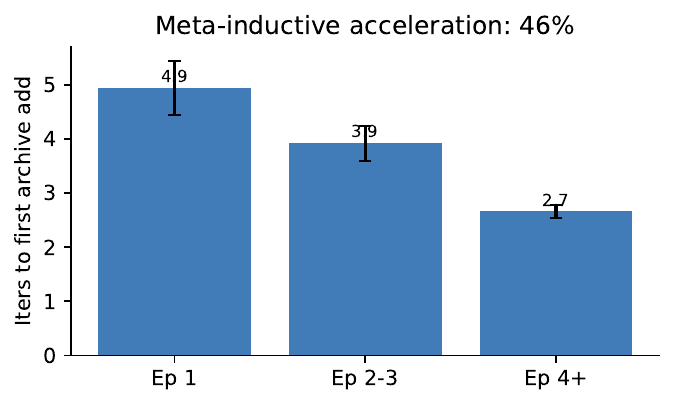}
\caption{Mean iterations-to-first-archive-admission per episode bucket within the meta-enabled preset. The within-preset trajectory accelerates on repeat shapes (\AvgIterationsEpOne{} $\to$ \AvgIterationsEpFourPlus{}) without any parameter updates.}
\label{fig:metalearning_episodes}
\end{figure}

\subsection{Backbone Comparison}
\label{app:backbone}
\begin{figure}[h]
\centering
\includegraphics[width=\textwidth]{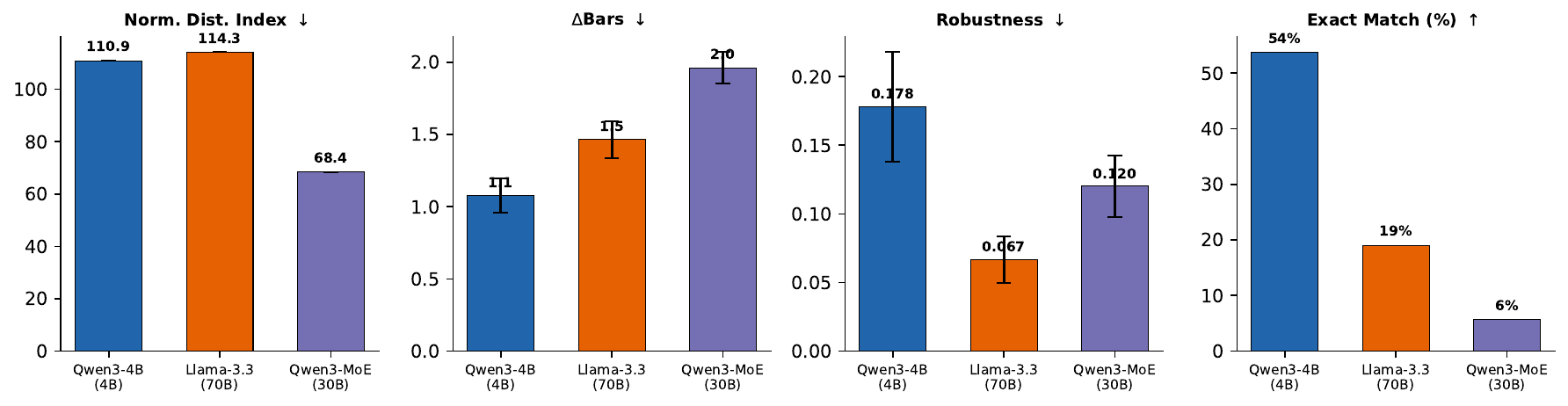}
\caption{Backbone comparison. Qwen3-4B (4B, blue) is competitive with 70B on adherence/distance; Llama-70B (orange) leads on robustness; MoE (purple) achieves best distance but worst adherence.}
\label{fig:backbone_comparison}
\end{figure}

\emph{Deepens \S\ref{sec:experiments} by giving the cross-backbone study referenced in the discussion: whether the structured protocol partially absorbs a model-scale deficit.}
\textbf{Within R-APS, Qwen3-4B (\QwenSmallDeltaBars{} $\Delta$Bars, \QwenSmallExact{}\% exact bar-count match) is competitive with Llama-3.3-70B (\LlamaDeltaBars{}, \LlamaExact{}\%) on adherence despite a \BackboneScaleRatio{}$\times$ parameter gap, consistent with the protocol partially absorbing a scale deficit} (Table~\ref{tab:backbone_comparison}, Fig.~\ref{fig:backbone_comparison}).
The 4B reasoning-specialized model leads on adherence and exact-match; the 70B general-purpose model leads on robustness (\LlamaRob{}), suggesting larger general-purpose models better navigate the counterfactual stress-testing game while reasoning specialization helps with constraint following; the MoE model (Qwen3-30B-A3B) achieves the best raw distance (\MoEDist{}) but the worst adherence (\MoEDeltaBars{}, \MoEExact{}\% exact) and best robustness, a pattern consistent with sparse routing introducing constraint-following inconsistency.
The compensation is therefore \emph{partial}: the structured protocol absorbs the scale gap on the constraint-following axes (Qwen3-4B matches or beats Llama-70B on adherence and exact-match despite a \BackboneScaleRatio{}$\times$ parameter deficit), but $\rho_{\text{sens}}$ still tracks scale and the MoE failure mode shows the protocol does not absorb sparse-routing inconsistency.
The actionable takeaway is that protocol gains and scale gains are \emph{complementary} rather than substitutable: pair the protocol with a reasoning-specialized model when adherence dominates, with a larger dense general-purpose model when worst-case robustness dominates.

\begin{table}[t]
\centering
\small
\begin{tabular}{@{}l r r r r@{}}
\toprule
Backbone & \textbf{Norm. Dist. Index\,$\downarrow$} & \textbf{$\Delta$Bars\,$\downarrow$} & \textbf{Rob.\,$\downarrow$} & \textbf{Exact Match\,$\uparrow$} \\
\midrule
Qwen3-4B (4B, reasoning) & $110.9 \pm 10.9$ & \cellcolor{bestcell} \textbf{$1.1 \pm 0.1$} & $0.178 \pm 0.040$ & \cellcolor{bestcell} \textbf{$53.8\%$} \\
Llama-3.3-70B (70B, general) & $114.3 \pm 17.9$ & $1.5 \pm 0.1$ & \cellcolor{bestcell} \textbf{$0.067 \pm 0.017$} & $19.0\%$ \\
Qwen3-30B-A3B (30B, MoE) & \cellcolor{bestcell} \textbf{$68.4 \pm 7.8$} & $2.0 \pm 0.1$ & $0.120 \pm 0.022$ & $5.8\%$ \\
\bottomrule
\end{tabular}
\caption{R-APS performance across LLM backbones (baseline preset, all shapes including parabola). \textbf{Norm. Dist. Index} = raw Chamfer $\times\,100$ / median Chamfer over the modular-baseline appendix rows, the same base used in Tables~\ref{tab:adherence_distance} and \ref{tab:ablation}; per-episode median aggregation neutralizes the parabola scale outlier. Within the structured protocol Qwen3-4B (a reasoning-specialized 4B model) is competitive with the 70B general-purpose backbone on adherence and the distance index, indicating that reasoning specialization matters within the protocol, not that 4B uniformly beats 70B. Exact match: percentage of episodes where synthesized bar count equals target.}
\label{tab:backbone_comparison}
\end{table}

\subsection{Failure-Type Diversity}
\label{app:entanglement}
\begin{figure}[h]
\centering
\includegraphics[width=0.55\textwidth]{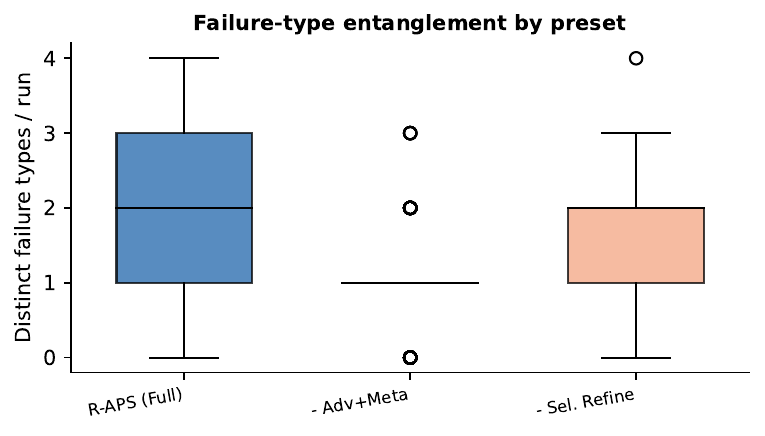}
\caption{Distribution of distinct failure-types per run across ablation presets. Full R-APS exercises \EntanglementBaseline{} distinct failure types per run on average; removing the adversary+meta-learning pair collapses this to \EntanglementNoAdvMeta{}, partially because the removed agent is the one that emits robustness-type diagnoses (so the signal is indirect, not a clean test of mode specialization).}
\label{fig:failure_entanglement}
\end{figure}

\emph{Deepens \S\ref{sec:q0_decomposition} (Q0) by reporting the supplementary failure-type-diversity signal that complements the main-text non-overlap pattern.}
Per-run distinct failure-type counts drop from \EntanglementBaseline{} for full R-APS to \EntanglementNoAdvMeta{} when the adversary+meta pair is removed (Mann--Whitney $U$, $p\!=\!\EntanglementP{}$; Table~\ref{tab:failure_entanglement}, Fig.~\ref{fig:failure_entanglement}).
We report this only as a supplementary signal: removing the adversary mechanically removes the agent that emits robustness-type diagnoses, so part of the drop is tautological with the ablation rather than an independent test of mode specialization.

\begin{table}[t]
\centering\small
\setlength{\tabcolsep}{4pt}
\begin{tabular}{@{}lrl@{}}
\toprule
Preset & Mean distinct failure types / episode [95\% CI] & $p$ vs Baseline \\
\midrule
R-APS (Full) & $1.94\ [1.88, 2.01]$ & -- \\
- Adv+Meta & $1.03\ [0.94, 1.12]$ & $< 10^{-4}$ \\
- Sel. Refine & $1.67\ [1.59, 1.75]$ & $< 10^{-4}$ \\
\bottomrule
\end{tabular}
\caption{Mode-specialization signal via per-episode failure-type diversity. For each episode, we count the number of \emph{distinct} failure buckets (Topology, Param., Optimization, Robustness, Critique) encountered before the first archive admission. R-APS (Full) exercises $1.91$ distinct failure types per episode on average. Removing the adversary\,+\,meta-learning pair collapses this to $1.03$ --- a specific mechanistic drop attributable to the fact that robustness-type failures cannot be diagnosed when the adversary is absent, consistent with \emph{each mode being responsible for a distinct failure class}. Removing selective refinement produces a smaller drop ($1.45$), as expected: selective refinement is a recovery mechanism, not a failure \emph{detector}. All contrasts are significant under Mann--Whitney $U$ (two-sided). \emph{Caveat}: this metric is confounded with per-episode iteration count (longer episodes have more opportunities to encounter additional failure types). We report it as indirect evidence for mode-specialization, not a direct test of the merged-context pair-merge ablation (which remains future work).}
\label{tab:failure_entanglement}
\end{table}

\subsection{Per-Shape Pareto Fronts and Hypervolume}
\label{app:pareto}
\emph{Deepens \S\ref{sec:q0_decomposition} and \S\ref{sec:q2_counterfactual} by giving the per-shape Pareto-front visualization and the aggregate normalised hypervolume table referenced inline.}
Aggregate normalised hypervolume across shapes is similar across presets ($\HVFull$ Full, $\HVNoAdvMeta$ NoAdvMeta, $\HVNoSelRef$ NoSelRef; Table~\ref{tab:pareto_hv}) because each preset's archive contributes to the per-shape reference rectangle's extrema; the load-bearing multi-objective signal is therefore the per-shape \emph{shape} of the front, not the aggregate area.

\begin{table}[t]
\centering\small
\begin{tabular}{@{}l rr@{}}
\toprule
Preset & Hypervolume$\,\uparrow$ & \#shapes \\
\midrule
R-APS (Full) & $0.96 \pm 0.00$ & $31$ \\
- Adv+Meta & \cellcolor{bestcell} \textbf{$0.99 \pm 0.00$} & $31$ \\
- Sel. Refine & $0.96 \pm 0.01$ & $31$ \\
\bottomrule
\end{tabular}
\caption{Pareto hypervolume on (Chamfer distance, $\rho_{\text{sens}}$) per preset, normalised to the per-shape reference rectangle $[x_{\min}, x_{\max}+\delta]\times[y_{\min}, y_{\max}+\delta]$ where the extrema are taken across all preset archives on that shape and $\delta$ is a 5\% slack. Higher is better; values are mean $\pm$ SEM across shapes. Removing the adversary+meta pair collapses the front along the $\rho_{\text{sens}}$ axis (no robustness exploration); removing selective refinement collapses it along the Chamfer axis (failed-iteration designs reach the archive without re-fitting). Modular LLM does not maintain a 2-objective archive at all and is shown only when an archive can be reconstructed from its run logs.}
\label{tab:pareto_hv}
\end{table}

\begin{figure}[h]
\centering
\includegraphics[width=0.85\textwidth]{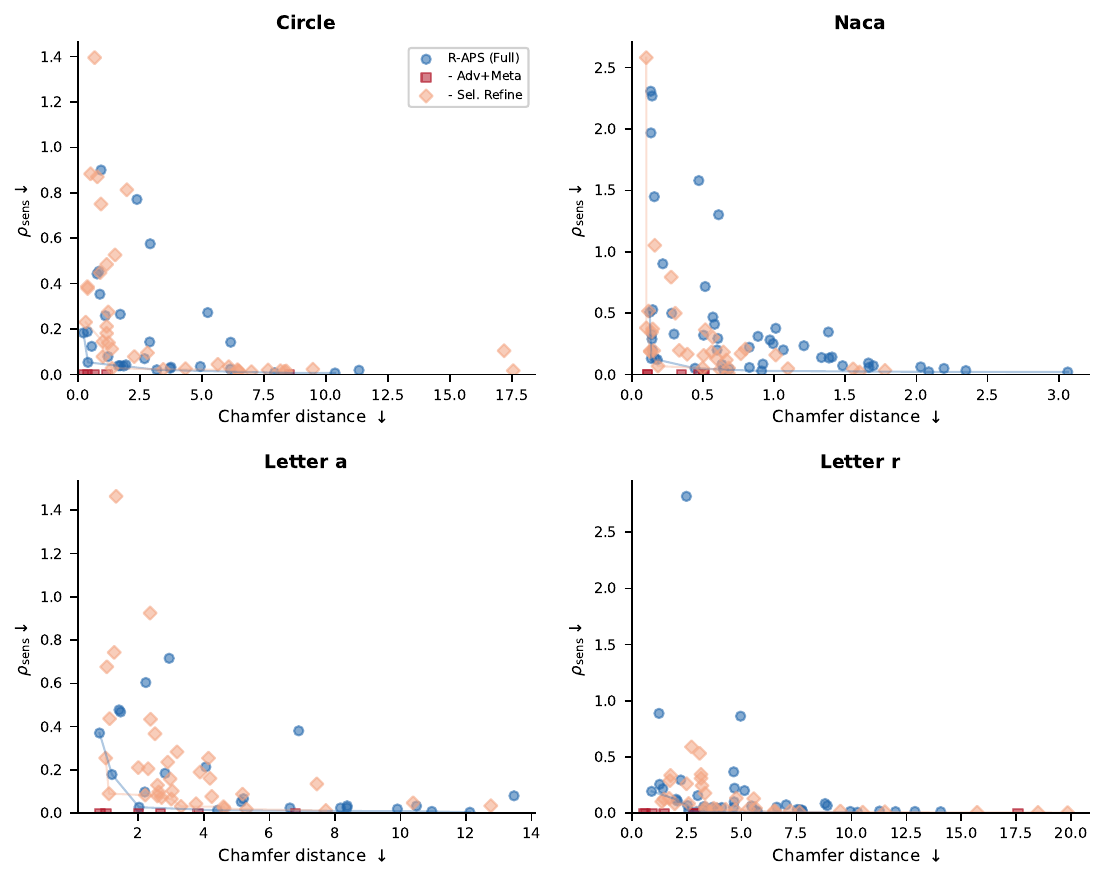}
\caption{Per-shape Pareto fronts on (Chamfer, $\rho_{\text{sens}}$) for four representative shapes. Each marker is an archive entry; connected lines are the per-preset non-dominated frontier. Full R-APS (blue) covers a wider region of the front than the ablations: removing the adversary+meta pair (red) collapses the front along $\rho_{\text{sens}}$ (no robustness exploration), while removing selective refinement (orange) collapses it along Chamfer (failed-iteration designs reach the archive without re-fitting). The two collapses target different axes, the multi-objective signature of mode specialisation.}
\label{fig:pareto_fronts_per_shape}
\end{figure}

\subsection{Refinement Recovery}
\label{app:refinement_recovery}
\begin{figure}[h]
\centering
\includegraphics[width=0.6\textwidth]{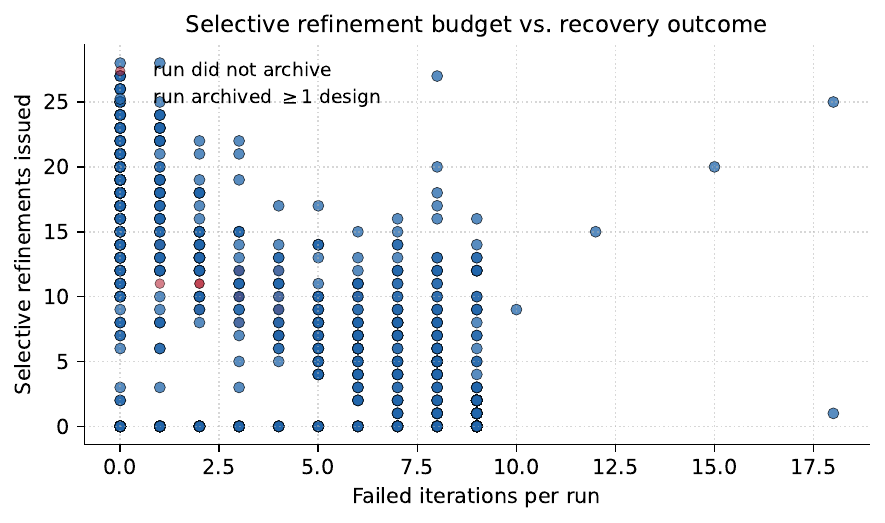}
\caption{Per-run scatter of selective refinements issued vs.\ failed iterations. Runs that admit at least one design consume their refinement budget in proportion to critic-identified failures, evidence of \emph{targeted} rather than blind retry.}
\label{fig:refinement_recovery_main}
\end{figure}

\emph{Deepens \S\ref{sec:q1_intrastage} (Q1) by giving the per-run scatter that complements the per-stage refinement budgets in Table~\ref{tab:qual_evidence}.}
Runs that admit at least one design (blue) sit on a different cluster than runs that exhaust budget without admitting (orange), so refinement budget tracks recovery outcome rather than firing blindly (Fig.~\ref{fig:refinement_recovery_main}).

\subsection{Per-Shape Adherence and Distance (R-APS Variants)}
\label{app:per_shape_main}
\emph{Deepens \S\ref{sec:q1_intrastage} (Q1) by giving the per-shape raw Chamfer distance across R-APS variants and visualizing the $\Delta$Bars gap shape-by-shape.}

\begin{table}[t]
\centering
\small
\begin{tabular}{@{}l rrrr@{}}
\toprule
Shape & \textbf{Modular (pooled)} & \textbf{R-APS (Full)} & \textbf{w/o Sel.\ Ref.} & \textbf{w/o Adv\,+\,Meta} \\
\midrule
Circle & $3.79 \pm 0.29$ & \cellcolor{bestcell} \textbf{$1.77 \pm 0.36$} & $2.92 \pm 1.24$ & $2.45 \pm 1.75$ \\
Ellipse & $3.83 \pm 0.18$ & $3.02 \pm 0.62$ & $2.02 \pm 0.60$ & \cellcolor{bestcell} \textbf{$1.88 \pm 0.40$} \\
Line & \cellcolor{bestcell} \textbf{$2.39 \pm 0.22$} & $4.81 \pm 1.67$ & $8.46 \pm 1.90$ & $2.96 \pm 1.01$ \\
LB & $7.80 \pm 0.19$ & $6.94 \pm 0.51$ & $8.57 \pm 0.59$ & \cellcolor{bestcell} \textbf{$5.99 \pm 0.45$} \\
NACA & $0.58 \pm 0.03$ & $0.62 \pm 0.10$ & $0.44 \pm 0.12$ & \cellcolor{bestcell} \textbf{$0.30 \pm 0.09$} \\
Alphabet (26 letters, agg.) & \textcolor{gray}{---} & $5.72 \pm 0.25$ & $4.75 \pm 0.42$ & \cellcolor{bestcell} \textbf{$2.91 \pm 0.28$} \\
\midrule
Parabola \textit{(scale outlier)} & $801.10 \pm 15.37$ & $636.92 \pm 39.63$ & \cellcolor{bestcell} \textbf{$565.43 \pm 64.68$} & $622.90 \pm 93.89$ \\
\bottomrule
\end{tabular}
\caption{Per-shape \emph{raw} Chamfer distance (mean\,$\pm$\,SE; lower is better). Modular column pools across the three modular-LLM backbones reported in the appendix. Parabola is shown in its own row because its raw scale ($\sim$100$\times$ the other shapes) drives the choice of median normalization in Tables~\ref{tab:adherence_distance}, \ref{tab:ablation}, and \ref{tab:backbone_comparison}; reporting it separately makes the actual number visible rather than burying it in an aggregate. Alphabet aggregate is the mean over the 26 R-APS letter shapes for which no modular baseline exists. Best per row is \textbf{highlighted}.}
\label{tab:per_shape_distance}
\end{table}

\begin{figure}[h]
\centering
\includegraphics[width=0.75\textwidth]{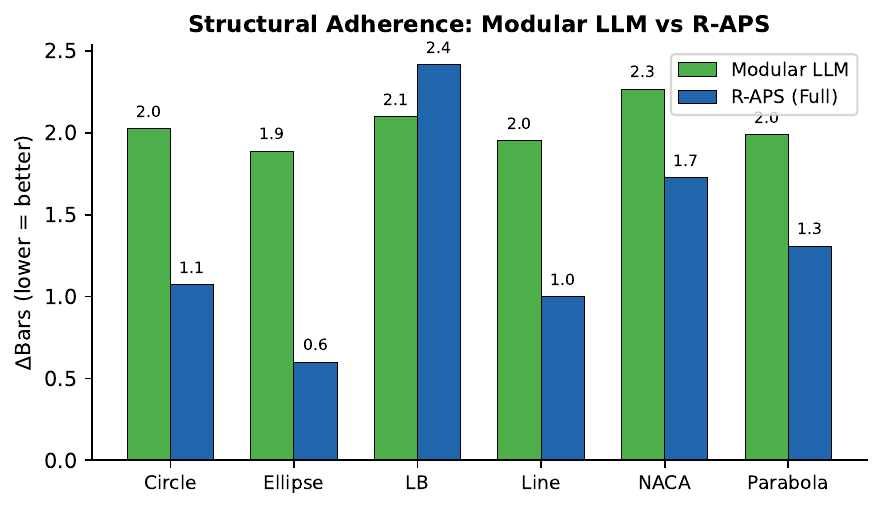}
\caption{Structural adherence ($\Delta$Bars, lower is better) per shape. R-APS (blue) maintains tighter control than modular baselines (green) on every shape group.}
\label{fig:per_shape_adherence}
\end{figure}

\subsection{Failure Distribution Through the Decomposition Lens}
\label{app:decomposition_lens}
\emph{Deepens \S\ref{sec:q1_intrastage} (Q1) by mapping the failure mass onto the reasoning-mode decomposition of Table~\ref{tab:decomposition_map}.}
The cascade's failure mass aligns with the decomposition map: \ReasoningModeAbductivePct{} abductive (Topology+Param), \ReasoningModeCorrectivePct{} corrective (Optimization), \ReasoningModeCounterfactualPct{} counterfactual (Robustness), the predicted shape from Table~\ref{tab:decomposition_map}.
Critic~1 failures are abductive-mode failures localized before the corrective mode runs; Critic~2 (\QualParamPct{} parameterization, \QualOptPct{} optimization) lies at the abductive$\leftrightarrow$counterfactual boundary where an initially plausible topology fails its intent check; the residual \QualRobustPct{} reaching the counterfactual screen are genuinely worst-case-sensitive designs for which only that mode can supply the diagnosis.

\subsection{Heuristic Transfer Detail}
\label{app:heuristic_transfer}
\emph{Deepens \S\ref{sec:q3_metainductive} (Q3) by reporting the rule-ID-level multi-shape reuse rate, the matched-pool memoization controls, and representative cross-shape invalidations referenced in the main text.}

\begin{table}[t]
\centering\scriptsize
\setlength{\tabcolsep}{4pt}
\begin{tabular}{@{}lll@{}}
\toprule
Shape & Rule ID & Invalidation reason \\
\midrule
alphabet\_t & HEUR\_001 & No evidence found for its application in T-shape topologies; counterexample: T-Shape-6Bar\ldots \\
circle & HEUR\_7 & counterexamples found in mechanisms M10 and M12 \\
alphabet\_c & HEUR\_4 & Assumed that all 'C-Shape' topologies with bar\_count ≥4 are inherently constructible; co\ldots \\
alphabet\_d & HEUR\_X & Counterexamples found in mechanisms M4 and M9 \\
alphabet\_l & HEUR\_7 & No evidence in current archive; not applicable to current failure cases involving bar\_co\ldots \\
line & HEUR\_001 & No counterexamples found in current archive, but pattern is not yet supported by ≥3 mecha\ldots \\
alphabet\_i & HEUR\_NEW\_4 & Counterexamples in M\_6, M\_8: n\_labels ≥ 4 but n\_guard\_crossings > 0 still benefit fr\ldots \\
alphabet\_s & HEUR\_001 & Counterexample in S-shaped-6bar-dyad (iteration 7) where n\_region\_crossings=2 but no im\ldots \\
\bottomrule
\end{tabular}
\caption{Case study: every meaningful heuristic explicitly invalidated by the Meta-Analyst across the full R-APS corpus. Invalidation reasons cite concrete counterexamples (e.g.\ mechanism IDs or specific topologies), evidencing that the rule library is non-monotonic by construction.}
\label{tab:invalidated_heuristics}
\end{table}

\begin{table}[t]
\centering\small
\setlength{\tabcolsep}{6pt}
\begin{tabular}{@{}lr@{}}
\toprule
Quantity & Value \\
\midrule
Total learned heuristics (baseline corpus) & 507 \\
\quad with multi-shape applicability declared & 397 (78\%) \\
Distinct rule IDs across baseline corpus & 24 \\
\quad applied across $\ge 2$ distinct shapes (rule-id match) & 23 (96\%) \\
Cross-shape invalidations (origin shape $\ne$ invalidating shape) & 2 / 4 \\
Library-size regression (iters $\sim$ rules-in-library) & $-0.068$ iters/rule [$-0.138, +0.062$]; $p=0.234$ \\
\bottomrule
\end{tabular}
\caption{Cross-shape heuristic re-use evidence, computed from the \texttt{meta\_learning\_digest.json} of every baseline R-APS run. \textbf{Multi-shape applicability declared} counts heuristics whose \texttt{applicability} field lists more than one topology or whose conditions explicitly span multiple shapes --- this is the agent's own claim that a rule generalizes. \textbf{Applied across $\ge 2$ shapes} is the stronger empirical check: rule IDs that actually appear in the digests of episodes on $\ge 2$ distinct shapes, evidencing that the rule was carried forward and re-applied, not merely declared. \textbf{Cross-shape invalidations} are rules whose origin shape (first episode that emitted the rule) differs from the shape of the episode that invalidated it --- direct evidence that rules are exercised across shapes, even when the application fails. The \textbf{library-size regression} pairs each shape's first-ever baseline episode with the size of the heuristic library at that timestamp; a negative slope means more accumulated rules predict faster first-success on the next previously-unseen shape. We report this alongside (not in place of) the corpus-order slope in Table~\ref{tab:meta_learning_transfer}, which uses corpus position as the predictor instead. We do \emph{not} claim statistically significant time-savings on truly novel shapes; we claim that the rule library is exercised across shapes (re-use fraction) and that within-shape acceleration is mediated by the meta-analyst (Table~\ref{tab:meta_learning_transfer}, matched-shape contrast).}
\label{tab:heuristic_transfer}
\end{table}

\begin{table}[t]
\centering\small
\setlength{\tabcolsep}{4pt}
\begin{tabular}{@{}ll@{}}
\toprule
Condition & Iterations (mean [95\% CI]) / statistic \\
\midrule
\multicolumn{2}{@{}l}{\textit{(a) Matched-shape contrast (primary): shapes accumulating $\ge 4$ episodes in \emph{both} presets, $|S|=31$.}}\\
Baseline (meta on) (Ep1) & $4.9\ [4.0,5.9]$ \\
Baseline (meta on) (Ep4+) & $2.7\ [2.4,2.9]$ \\
\midrule
No-meta (meta off) (Ep1) & $1.3\ [1.1,1.5]$ \\
No-meta (meta off) (Ep4+) & $1.6\ [1.4,1.9]$ \\
\midrule
Matched accel.\ Ep1$\to$Ep4+ & Baseline: $+46\%$; No-meta: $-28\%$ \\
Paired $\Delta$ (Baseline$-$No-meta, Ep4+) & median $+1.0$ iters [$+0.9, +1.5$]; Wilcoxon $p=< 10^{-4}$ \\
\midrule
\multicolumn{2}{@{}l}{\textit{(b) Unmatched pool (sensitivity): every shape.}}\\
Baseline (meta on) (Ep1) & $4.9\ [4.0,5.9]$ \\
Baseline (meta on) (Ep4+) & $2.7\ [2.4,2.9]$ \\
\midrule
No-meta (meta off) (Ep1) & $1.3\ [1.1,1.5]$ \\
No-meta (meta off) (Ep4+) & $1.6\ [1.4,1.9]$ \\
\midrule
Unmatched accel.\ Ep1$\to$Ep4+ & Baseline: $+46\%$; No-meta: $-28\%$ \\
Novel-shape corpus-order slope (sensitivity) & $-0.0372$ iters/episode [$-0.0734, +0.0190$]; $p=0.201$ \\
\bottomrule
\end{tabular}
\caption{Memoization controls for the meta-inductive acceleration claim. \textbf{(a) Matched-shape contrast.} The $|S|=31$ shapes shared by \emph{both} the baseline and the no\_adversary\_no\_meta preset (each accumulating $\ge 4$ episodes) are all alphabet letters --- the only shapes on which no-meta produced Ep4+ data. On this matched pool the within-preset \emph{trajectories} differ sharply: baseline iterations drop from Ep1 to Ep4+ (4.9$\to$2.7, a $+46\%$ change where positive = faster); no-meta iterations actually rise (1.3$\to$1.6, $-28\%$). The cross-preset Ep4+ comparison (\emph{Paired $\Delta$} row, Wilcoxon signed-rank on per-shape $\mathrm{baseline}_{\mathrm{Ep4+}}-\mathrm{no\_meta}_{\mathrm{Ep4+}}$) is reported transparently and goes \emph{against} baseline in absolute iterations, because the matched pool consists of easy alphabet letters that no-meta resolves in 1--2 iterations without any meta-overhead, while baseline pays a small fixed cost for the meta-analyst. The load-bearing claim is therefore the trajectory, not the absolute endpoint: only the meta-enabled preset \emph{learns down} on repeat episodes. \textbf{(b) Unmatched-pool sensitivity.} The same per-bucket means computed without the shape filter; the unmatched no-meta pool is small, so the acceleration contrast appears larger but is confounded by which shapes the two presets actually completed. \textbf{Novel-shape corpus-order slope} (last row, sensitivity): for each shape's first-ever baseline episode, regress iterations-to-first-success on the corpus position of that episode. The slope is reported for completeness; we do \emph{not} claim a statistically significant effect on truly-novel-shape transfer (see Table~\ref{tab:heuristic_transfer} for the rule-reuse evidence we do claim). 95\% CIs are percentile bootstraps (1000 resamples).}
\label{tab:meta_learning_transfer}
\end{table}

\subsection{Canonical Normalization}
\label{app:norm}
\emph{Deepens \S\ref{sec:setup} by giving the full normalization protocol summarized inline in the main text.}
All three main comparison tables (\S\ref{sec:q1_intrastage}, \S\ref{sec:q2_counterfactual}, and the backbone study in \S\ref{app:backbone}) report a Norm.\ Dist.\ Index defined as the per-run-median Chamfer distance times $100$, divided by a single shared base: the median Chamfer distance over all \TotalShapes{} modular-baseline appendix rows, parabola included.
This single base is used in Tables~\ref{tab:adherence_distance},~\ref{tab:ablation},~\ref{tab:backbone_comparison}, so the index is directly comparable across all three.
Raw per-shape Chamfer distances are reported separately in Table~\ref{tab:per_shape_distance}, with parabola broken out into its own row because its raw scale is ${\sim}100\!\times$ that of the other shapes; including parabola in the aggregate would dominate every aggregate by an outlier whose absolute scale carries no comparative information.

\subsection{Per-Shape Modular LLM vs.\ R-APS Comparison}
\label{app:per_shape_modular}
\emph{Deepens \S\ref{sec:q1_intrastage} (Q1) by giving the per-shape adherence and distance for the modular-LLM control versus R-APS that underlies the aggregate Table~\ref{tab:adherence_distance}.}
Each row reports the best modular-LLM configuration on that shape against the corresponding R-APS run; rows where modular obtains a lower distance are precisely the rows where it pays a higher $\Delta$Bars, the per-shape signature of optimizing trajectory accuracy alone without joint adherence and robustness.

\begin{table}[t]
\centering
\footnotesize
\begin{tabular}{@{}l rr rr@{}}
\toprule
& \multicolumn{2}{c}{\textbf{Modular LLM (best)}} & \multicolumn{2}{c}{\textbf{R-APS (Full)}} \\
\cmidrule(lr){2-3} \cmidrule(lr){4-5}
Shape & \textbf{Dist.\,$\downarrow$} & \textbf{$\Delta$Bars\,$\downarrow$} & \textbf{Dist.\,$\downarrow$} & \textbf{$\Delta$Bars\,$\downarrow$} \\
\midrule
Circle & $3.8$ & $2.0$ & \cellcolor{bestcell} \textbf{$1.8$} & \cellcolor{bestcell} \textbf{$1.1$} \\
Ellipse & $3.8$ & $1.9$ & \cellcolor{bestcell} \textbf{$3.0$} & \cellcolor{bestcell} \textbf{$0.6$} \\
LB & $7.8$ & \cellcolor{bestcell} \textbf{$2.1$} & \cellcolor{bestcell} \textbf{$6.9$} & $2.4$ \\
Line & \cellcolor{bestcell} \textbf{$2.4$} & $2.0$ & $4.8$ & \cellcolor{bestcell} \textbf{$1.0$} \\
NACA & \cellcolor{bestcell} \textbf{$0.6$} & $2.3$ & $0.6$ & \cellcolor{bestcell} \textbf{$1.7$} \\
Parabola & $801.1$ & $2.0$ & \cellcolor{bestcell} \textbf{$636.9$} & \cellcolor{bestcell} \textbf{$1.3$} \\
\bottomrule
\end{tabular}
\caption{Per-shape comparison on standard benchmark shapes. For modular LLM baselines we average over all configurations
that include the shape; for R-APS we average over baseline preset episodes. Lower
is better; best per cell is \textbf{highlighted}.}
\label{tab:per_shape_full}
\end{table}

\subsection{Per-Shape Enum+GA vs.\ R-APS Comparison}
\label{app:per_shape_enumga}
\emph{Deepens \S\ref{sec:q4_classical} (Q4) by giving per-shape Chamfer distance for Enum+\acs{ga} versus R-APS at all three GA budgets (3/20, 6/20, 60/300) for both 4-bar and 6-bar topologies, the data behind the aggregate Table~\ref{tab:enum_ga_summary}.}
Best per row is highlighted; the R-APS advantage is largest on standard curves where topology discovery dominates and narrows on alphabet letters, where a wider set of near-optimal topologies makes the GA's brute enumeration more competitive (R-APS still wins the alphabet group on average by \EnumGALtrSpeedup{}$\times$ at the 6-bar/60/300 budget; see Table~\ref{tab:enum_ga_summary}).

\begin{figure}[h]
\centering
\includegraphics[width=0.75\textwidth]{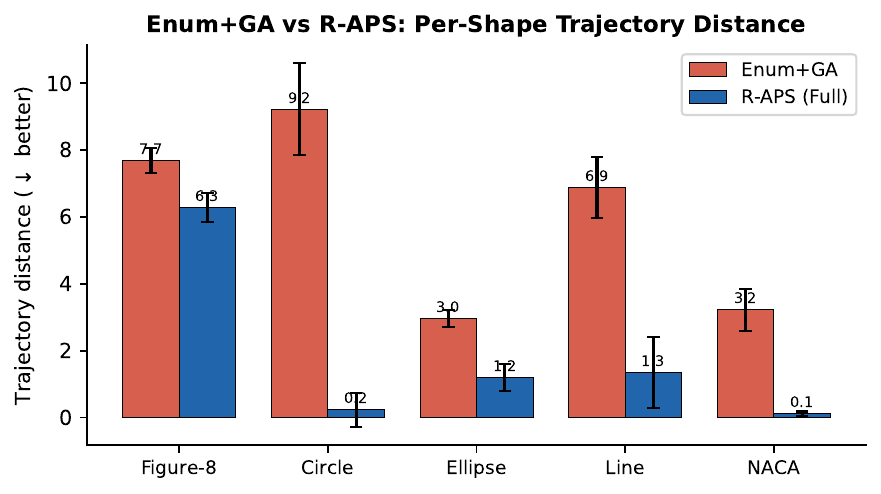}
\caption{Per-shape trajectory distance on the standard-curve subset (lower is better). R-APS matches or beats Enum+\acs{ga} on every shape while additionally optimizing robustness and adherence.}
\label{fig:enum_ga_comparison}
\end{figure}

\begin{table}[t]
\centering
\small
\begin{tabular}{@{}l rr@{}}
\toprule
Shape & \textbf{Enum+GA Distance\,$\downarrow$} & \textbf{R-APS (Full) Distance\,$\downarrow$} \\
\midrule
Figure-8 & $10.12 \pm 0.47$ & \cellcolor{bestcell} \textbf{$6.28 \pm 0.44$} \\
Circle & $17.56 \pm 1.65$ & \cellcolor{bestcell} \textbf{$0.23 \pm 0.51$} \\
Ellipse & $6.32 \pm 0.91$ & \cellcolor{bestcell} \textbf{$1.20 \pm 0.40$} \\
Line & $9.87 \pm 2.39$ & \cellcolor{bestcell} \textbf{$1.34 \pm 1.07$} \\
NACA & $5.81 \pm 1.18$ & \cellcolor{bestcell} \textbf{$0.11 \pm 0.07$} \\
Parabola & $887.65 \pm 15.88$ & \cellcolor{bestcell} \textbf{$324.32 \pm 33.08$} \\
Letter A & $5.42 \pm 1.33$ & \cellcolor{bestcell} \textbf{$1.92 \pm 0.97$} \\
Letter B & $3.04 \pm 0.34$ & \cellcolor{bestcell} \textbf{$3.03 \pm 0.99$} \\
Letter C & $6.82 \pm 1.79$ & \cellcolor{bestcell} \textbf{$0.94 \pm 0.68$} \\
Letter D & $7.70 \pm 2.87$ & \cellcolor{bestcell} \textbf{$1.45 \pm 0.75$} \\
Letter E & $10.22 \pm 2.87$ & \cellcolor{bestcell} \textbf{$1.77 \pm 0.74$} \\
Letter F & $7.66 \pm 1.98$ & \cellcolor{bestcell} \textbf{$0.75 \pm 1.41$} \\
Letter G & $5.23 \pm 0.93$ & \cellcolor{bestcell} \textbf{$2.10 \pm 1.19$} \\
Letter H & $7.44 \pm 2.08$ & \cellcolor{bestcell} \textbf{$1.17 \pm 0.69$} \\
Letter I & $6.63 \pm 0.28$ & \cellcolor{bestcell} \textbf{$0.94 \pm 0.33$} \\
Letter J & \cellcolor{bestcell} \textbf{$5.03 \pm 1.54$} & $5.74 \pm 0.70$ \\
Letter K & $10.92 \pm 2.22$ & \cellcolor{bestcell} \textbf{$0.70 \pm 1.03$} \\
Letter L & $4.95 \pm 0.78$ & \cellcolor{bestcell} \textbf{$2.78 \pm 0.70$} \\
Letter M & $10.65 \pm 3.32$ & \cellcolor{bestcell} \textbf{$5.20 \pm 1.05$} \\
Letter N & $5.28 \pm 0.76$ & \cellcolor{bestcell} \textbf{$1.41 \pm 0.73$} \\
Letter O & $8.82 \pm 1.35$ & \cellcolor{bestcell} \textbf{$0.87 \pm 0.83$} \\
Letter P & $5.10 \pm 0.60$ & \cellcolor{bestcell} \textbf{$1.99 \pm 0.73$} \\
Letter Q & $11.16 \pm 3.35$ & \cellcolor{bestcell} \textbf{$1.55 \pm 1.34$} \\
Letter R & $9.17 \pm 2.35$ & \cellcolor{bestcell} \textbf{$0.62 \pm 0.78$} \\
Letter S & $6.07 \pm 1.78$ & \cellcolor{bestcell} \textbf{$1.68 \pm 1.13$} \\
Letter T & $10.56 \pm 3.16$ & \cellcolor{bestcell} \textbf{$5.57 \pm 2.15$} \\
Letter U & $6.24 \pm 2.69$ & \cellcolor{bestcell} \textbf{$2.79 \pm 0.86$} \\
Letter V & $6.96 \pm 1.64$ & \cellcolor{bestcell} \textbf{$1.33 \pm 0.57$} \\
Letter W & $7.13 \pm 0.90$ & \cellcolor{bestcell} \textbf{$0.95 \pm 1.22$} \\
Letter X & $5.80 \pm 1.27$ & \cellcolor{bestcell} \textbf{$2.68 \pm 0.94$} \\
Letter Y & $5.75 \pm 1.70$ & \cellcolor{bestcell} \textbf{$2.77 \pm 0.73$} \\
Letter Z & $9.06 \pm 1.51$ & \cellcolor{bestcell} \textbf{$2.59 \pm 1.05$} \\
\bottomrule
\end{tabular}
\caption{Per-shape comparison against the classical Enum+GA baseline for the 4-bar family with GA budget 3/20. R-APS reports the best mean trajectory distance per shape over model+mode groups. Best per row is \textbf{highlighted}.}
\label{tab:enum_ga_full_4bar_3_20}
\end{table}

\begin{table}[t]
\centering
\small
\begin{tabular}{@{}l rr@{}}
\toprule
Shape & \textbf{Enum+GA Distance\,$\downarrow$} & \textbf{R-APS (Full) Distance\,$\downarrow$} \\
\midrule
Figure-8 & $8.56 \pm 0.36$ & \cellcolor{bestcell} \textbf{$6.28 \pm 0.44$} \\
Circle & $12.63 \pm 2.16$ & \cellcolor{bestcell} \textbf{$0.23 \pm 0.51$} \\
Ellipse & $4.16 \pm 0.58$ & \cellcolor{bestcell} \textbf{$1.20 \pm 0.40$} \\
Line & $8.43 \pm 2.31$ & \cellcolor{bestcell} \textbf{$1.34 \pm 1.07$} \\
NACA & $3.01 \pm 0.68$ & \cellcolor{bestcell} \textbf{$0.11 \pm 0.07$} \\
Parabola & $883.13 \pm 15.68$ & \cellcolor{bestcell} \textbf{$324.32 \pm 33.08$} \\
Letter A & $4.31 \pm 1.26$ & \cellcolor{bestcell} \textbf{$1.92 \pm 0.97$} \\
Letter B & $3.25 \pm 0.26$ & \cellcolor{bestcell} \textbf{$3.03 \pm 0.99$} \\
Letter C & $3.97 \pm 0.81$ & \cellcolor{bestcell} \textbf{$0.94 \pm 0.68$} \\
Letter D & $4.60 \pm 0.86$ & \cellcolor{bestcell} \textbf{$1.45 \pm 0.75$} \\
Letter E & $5.58 \pm 1.20$ & \cellcolor{bestcell} \textbf{$1.77 \pm 0.74$} \\
Letter F & $4.38 \pm 0.63$ & \cellcolor{bestcell} \textbf{$0.75 \pm 1.41$} \\
Letter G & $4.31 \pm 1.19$ & \cellcolor{bestcell} \textbf{$2.10 \pm 1.19$} \\
Letter H & $3.19 \pm 0.71$ & \cellcolor{bestcell} \textbf{$1.17 \pm 0.69$} \\
Letter I & $4.93 \pm 1.26$ & \cellcolor{bestcell} \textbf{$0.94 \pm 0.33$} \\
Letter J & \cellcolor{bestcell} \textbf{$3.50 \pm 0.67$} & $5.74 \pm 0.70$ \\
Letter K & $5.15 \pm 1.37$ & \cellcolor{bestcell} \textbf{$0.70 \pm 1.03$} \\
Letter L & $3.72 \pm 0.44$ & \cellcolor{bestcell} \textbf{$2.78 \pm 0.70$} \\
Letter M & $7.52 \pm 2.97$ & \cellcolor{bestcell} \textbf{$5.20 \pm 1.05$} \\
Letter N & $3.11 \pm 0.69$ & \cellcolor{bestcell} \textbf{$1.41 \pm 0.73$} \\
Letter O & $4.78 \pm 1.16$ & \cellcolor{bestcell} \textbf{$0.87 \pm 0.83$} \\
Letter P & $2.78 \pm 0.72$ & \cellcolor{bestcell} \textbf{$1.99 \pm 0.73$} \\
Letter Q & $4.94 \pm 0.84$ & \cellcolor{bestcell} \textbf{$1.55 \pm 1.34$} \\
Letter R & $4.38 \pm 0.97$ & \cellcolor{bestcell} \textbf{$0.62 \pm 0.78$} \\
Letter S & $2.42 \pm 0.29$ & \cellcolor{bestcell} \textbf{$1.68 \pm 1.13$} \\
Letter T & $6.34 \pm 1.47$ & \cellcolor{bestcell} \textbf{$5.57 \pm 2.15$} \\
Letter U & $3.45 \pm 1.16$ & \cellcolor{bestcell} \textbf{$2.79 \pm 0.86$} \\
Letter V & $6.96 \pm 1.95$ & \cellcolor{bestcell} \textbf{$1.33 \pm 0.57$} \\
Letter W & $4.50 \pm 1.24$ & \cellcolor{bestcell} \textbf{$0.95 \pm 1.22$} \\
Letter X & $4.56 \pm 1.90$ & \cellcolor{bestcell} \textbf{$2.68 \pm 0.94$} \\
Letter Y & $5.10 \pm 1.54$ & \cellcolor{bestcell} \textbf{$2.77 \pm 0.73$} \\
Letter Z & $4.44 \pm 0.59$ & \cellcolor{bestcell} \textbf{$2.59 \pm 1.05$} \\
\bottomrule
\end{tabular}
\caption{Per-shape comparison against the classical Enum+GA baseline for the 4-bar family with GA budget 6/20. R-APS reports the best mean trajectory distance per shape over model+mode groups. Best per row is \textbf{highlighted}.}
\label{tab:enum_ga_full_4bar_6_20}
\end{table}

\begin{table}[t]
\centering
\small
\begin{tabular}{@{}l rr@{}}
\toprule
Shape & \textbf{Enum+GA Distance\,$\downarrow$} & \textbf{R-APS (Full) Distance\,$\downarrow$} \\
\midrule
Figure-8 & $6.88 \pm 0.36$ & \cellcolor{bestcell} \textbf{$6.28 \pm 0.44$} \\
Circle & $3.23 \pm 1.18$ & \cellcolor{bestcell} \textbf{$0.23 \pm 0.51$} \\
Ellipse & $2.15 \pm 0.38$ & \cellcolor{bestcell} \textbf{$1.20 \pm 0.40$} \\
Line & $1.62 \pm 0.94$ & \cellcolor{bestcell} \textbf{$1.34 \pm 1.07$} \\
NACA & $0.44 \pm 0.09$ & \cellcolor{bestcell} \textbf{$0.11 \pm 0.07$} \\
Parabola & $871.93 \pm 22.10$ & \cellcolor{bestcell} \textbf{$324.32 \pm 33.08$} \\
Letter A & \cellcolor{bestcell} \textbf{$1.19 \pm 0.06$} & $1.92 \pm 0.97$ \\
Letter B & \cellcolor{bestcell} \textbf{$2.02 \pm 0.22$} & $3.03 \pm 0.99$ \\
Letter C & $0.94 \pm 0.23$ & \cellcolor{bestcell} \textbf{$0.94 \pm 0.68$} \\
Letter D & $1.63 \pm 0.44$ & \cellcolor{bestcell} \textbf{$1.45 \pm 0.75$} \\
Letter E & $2.18 \pm 0.31$ & \cellcolor{bestcell} \textbf{$1.77 \pm 0.74$} \\
Letter F & $1.37 \pm 0.14$ & \cellcolor{bestcell} \textbf{$0.75 \pm 1.41$} \\
Letter G & \cellcolor{bestcell} \textbf{$1.98 \pm 0.20$} & $2.10 \pm 1.19$ \\
Letter H & $1.67 \pm 0.25$ & \cellcolor{bestcell} \textbf{$1.17 \pm 0.69$} \\
Letter I & \cellcolor{bestcell} \textbf{$0.83 \pm 0.13$} & $0.94 \pm 0.33$ \\
Letter J & \cellcolor{bestcell} \textbf{$1.60 \pm 0.42$} & $5.74 \pm 0.70$ \\
Letter K & $1.41 \pm 0.38$ & \cellcolor{bestcell} \textbf{$0.70 \pm 1.03$} \\
Letter L & \cellcolor{bestcell} \textbf{$0.79 \pm 0.11$} & $2.78 \pm 0.70$ \\
Letter M & \cellcolor{bestcell} \textbf{$2.12 \pm 0.27$} & $5.20 \pm 1.05$ \\
Letter N & $1.82 \pm 0.19$ & \cellcolor{bestcell} \textbf{$1.41 \pm 0.73$} \\
Letter O & $1.77 \pm 0.54$ & \cellcolor{bestcell} \textbf{$0.87 \pm 0.83$} \\
Letter P & \cellcolor{bestcell} \textbf{$1.29 \pm 0.15$} & $1.99 \pm 0.73$ \\
Letter Q & $1.70 \pm 0.27$ & \cellcolor{bestcell} \textbf{$1.55 \pm 1.34$} \\
Letter R & $1.20 \pm 0.16$ & \cellcolor{bestcell} \textbf{$0.62 \pm 0.78$} \\
Letter S & $1.77 \pm 0.21$ & \cellcolor{bestcell} \textbf{$1.68 \pm 1.13$} \\
Letter T & \cellcolor{bestcell} \textbf{$2.14 \pm 0.36$} & $5.57 \pm 2.15$ \\
Letter U & \cellcolor{bestcell} \textbf{$1.47 \pm 0.19$} & $2.79 \pm 0.86$ \\
Letter V & $1.60 \pm 0.16$ & \cellcolor{bestcell} \textbf{$1.33 \pm 0.57$} \\
Letter W & $2.02 \pm 0.24$ & \cellcolor{bestcell} \textbf{$0.95 \pm 1.22$} \\
Letter X & \cellcolor{bestcell} \textbf{$1.67 \pm 0.27$} & $2.68 \pm 0.94$ \\
Letter Y & \cellcolor{bestcell} \textbf{$0.93 \pm 0.12$} & $2.77 \pm 0.73$ \\
Letter Z & \cellcolor{bestcell} \textbf{$2.19 \pm 0.22$} & $2.59 \pm 1.05$ \\
\bottomrule
\end{tabular}
\caption{Per-shape comparison against the classical Enum+GA baseline for the 4-bar family with GA budget 60/300. R-APS reports the best mean trajectory distance per shape over model+mode groups. Best per row is \textbf{highlighted}.}
\label{tab:enum_ga_full_4bar_60_300}
\end{table}

\begin{table}[t]
\centering
\small
\begin{tabular}{@{}l rr@{}}
\toprule
Shape & \textbf{Enum+GA Distance\,$\downarrow$} & \textbf{R-APS (Full) Distance\,$\downarrow$} \\
\midrule
Figure-8 & $9.26 \pm 0.76$ & \cellcolor{bestcell} \textbf{$6.28 \pm 0.44$} \\
Circle & $13.75 \pm 2.52$ & \cellcolor{bestcell} \textbf{$0.23 \pm 0.51$} \\
Ellipse & $4.02 \pm 0.50$ & \cellcolor{bestcell} \textbf{$1.20 \pm 0.40$} \\
Line & $9.36 \pm 1.33$ & \cellcolor{bestcell} \textbf{$1.34 \pm 1.07$} \\
NACA & $6.06 \pm 1.12$ & \cellcolor{bestcell} \textbf{$0.11 \pm 0.07$} \\
Parabola & $884.92 \pm 16.40$ & \cellcolor{bestcell} \textbf{$324.32 \pm 33.08$} \\
Letter A & $5.30 \pm 1.17$ & \cellcolor{bestcell} \textbf{$1.92 \pm 0.97$} \\
Letter B & $4.91 \pm 0.64$ & \cellcolor{bestcell} \textbf{$3.03 \pm 0.99$} \\
Letter C & $5.98 \pm 1.09$ & \cellcolor{bestcell} \textbf{$0.94 \pm 0.68$} \\
Letter D & $6.00 \pm 0.88$ & \cellcolor{bestcell} \textbf{$1.45 \pm 0.75$} \\
Letter E & $5.64 \pm 1.11$ & \cellcolor{bestcell} \textbf{$1.77 \pm 0.74$} \\
Letter F & $6.68 \pm 1.37$ & \cellcolor{bestcell} \textbf{$0.75 \pm 1.41$} \\
Letter G & $6.54 \pm 1.08$ & \cellcolor{bestcell} \textbf{$2.10 \pm 1.19$} \\
Letter H & $5.75 \pm 0.72$ & \cellcolor{bestcell} \textbf{$1.17 \pm 0.69$} \\
Letter I & $6.77 \pm 1.87$ & \cellcolor{bestcell} \textbf{$0.94 \pm 0.33$} \\
Letter J & \cellcolor{bestcell} \textbf{$4.88 \pm 1.08$} & $5.74 \pm 0.70$ \\
Letter K & $6.53 \pm 0.91$ & \cellcolor{bestcell} \textbf{$0.70 \pm 1.03$} \\
Letter L & $4.19 \pm 0.74$ & \cellcolor{bestcell} \textbf{$2.78 \pm 0.70$} \\
Letter M & $8.48 \pm 1.69$ & \cellcolor{bestcell} \textbf{$5.20 \pm 1.05$} \\
Letter N & $4.33 \pm 0.52$ & \cellcolor{bestcell} \textbf{$1.41 \pm 0.73$} \\
Letter O & $6.91 \pm 1.19$ & \cellcolor{bestcell} \textbf{$0.87 \pm 0.83$} \\
Letter P & $5.43 \pm 0.80$ & \cellcolor{bestcell} \textbf{$1.99 \pm 0.73$} \\
Letter Q & $9.25 \pm 2.14$ & \cellcolor{bestcell} \textbf{$1.55 \pm 1.34$} \\
Letter R & $4.12 \pm 0.45$ & \cellcolor{bestcell} \textbf{$0.62 \pm 0.78$} \\
Letter S & $4.46 \pm 0.64$ & \cellcolor{bestcell} \textbf{$1.68 \pm 1.13$} \\
Letter T & $8.68 \pm 1.54$ & \cellcolor{bestcell} \textbf{$5.57 \pm 2.15$} \\
Letter U & $6.36 \pm 1.05$ & \cellcolor{bestcell} \textbf{$2.79 \pm 0.86$} \\
Letter V & $6.46 \pm 0.72$ & \cellcolor{bestcell} \textbf{$1.33 \pm 0.57$} \\
Letter W & $7.48 \pm 1.21$ & \cellcolor{bestcell} \textbf{$0.95 \pm 1.22$} \\
Letter X & $5.13 \pm 1.06$ & \cellcolor{bestcell} \textbf{$2.68 \pm 0.94$} \\
Letter Y & $8.15 \pm 1.77$ & \cellcolor{bestcell} \textbf{$2.77 \pm 0.73$} \\
Letter Z & $6.31 \pm 0.86$ & \cellcolor{bestcell} \textbf{$2.59 \pm 1.05$} \\
\bottomrule
\end{tabular}
\caption{Per-shape comparison against the classical Enum+GA baseline for the 6-bar family with GA budget 3/20. R-APS reports the best mean trajectory distance per shape over model+mode groups. Best per row is \textbf{highlighted}.}
\label{tab:enum_ga_full_6bar_3_20}
\end{table}

\begin{table}[t]
\centering
\small
\begin{tabular}{@{}l rr@{}}
\toprule
Shape & \textbf{Enum+GA Distance\,$\downarrow$} & \textbf{R-APS (Full) Distance\,$\downarrow$} \\
\midrule
Figure-8 & $7.96 \pm 0.24$ & \cellcolor{bestcell} \textbf{$6.28 \pm 0.44$} \\
Circle & $11.70 \pm 1.96$ & \cellcolor{bestcell} \textbf{$0.23 \pm 0.51$} \\
Ellipse & $3.05 \pm 0.26$ & \cellcolor{bestcell} \textbf{$1.20 \pm 0.40$} \\
Line & $8.84 \pm 1.46$ & \cellcolor{bestcell} \textbf{$1.34 \pm 1.07$} \\
NACA & $4.04 \pm 0.91$ & \cellcolor{bestcell} \textbf{$0.11 \pm 0.07$} \\
Parabola & $881.65 \pm 15.87$ & \cellcolor{bestcell} \textbf{$324.32 \pm 33.08$} \\
Letter A & $3.73 \pm 0.54$ & \cellcolor{bestcell} \textbf{$1.92 \pm 0.97$} \\
Letter B & $3.37 \pm 0.21$ & \cellcolor{bestcell} \textbf{$3.03 \pm 0.99$} \\
Letter C & $3.90 \pm 0.80$ & \cellcolor{bestcell} \textbf{$0.94 \pm 0.68$} \\
Letter D & $4.20 \pm 0.82$ & \cellcolor{bestcell} \textbf{$1.45 \pm 0.75$} \\
Letter E & $5.13 \pm 1.23$ & \cellcolor{bestcell} \textbf{$1.77 \pm 0.74$} \\
Letter F & $3.99 \pm 0.84$ & \cellcolor{bestcell} \textbf{$0.75 \pm 1.41$} \\
Letter G & $6.04 \pm 0.58$ & \cellcolor{bestcell} \textbf{$2.10 \pm 1.19$} \\
Letter H & $5.17 \pm 0.96$ & \cellcolor{bestcell} \textbf{$1.17 \pm 0.69$} \\
Letter I & $5.28 \pm 1.44$ & \cellcolor{bestcell} \textbf{$0.94 \pm 0.33$} \\
Letter J & \cellcolor{bestcell} \textbf{$4.51 \pm 1.01$} & $5.74 \pm 0.70$ \\
Letter K & $4.42 \pm 0.80$ & \cellcolor{bestcell} \textbf{$0.70 \pm 1.03$} \\
Letter L & $3.68 \pm 0.56$ & \cellcolor{bestcell} \textbf{$2.78 \pm 0.70$} \\
Letter M & $7.24 \pm 1.70$ & \cellcolor{bestcell} \textbf{$5.20 \pm 1.05$} \\
Letter N & $3.27 \pm 0.47$ & \cellcolor{bestcell} \textbf{$1.41 \pm 0.73$} \\
Letter O & $5.77 \pm 1.27$ & \cellcolor{bestcell} \textbf{$0.87 \pm 0.83$} \\
Letter P & $2.58 \pm 0.31$ & \cellcolor{bestcell} \textbf{$1.99 \pm 0.73$} \\
Letter Q & $5.69 \pm 1.08$ & \cellcolor{bestcell} \textbf{$1.55 \pm 1.34$} \\
Letter R & $3.68 \pm 0.49$ & \cellcolor{bestcell} \textbf{$0.62 \pm 0.78$} \\
Letter S & $3.87 \pm 0.62$ & \cellcolor{bestcell} \textbf{$1.68 \pm 1.13$} \\
Letter T & $6.76 \pm 1.57$ & \cellcolor{bestcell} \textbf{$5.57 \pm 2.15$} \\
Letter U & $4.37 \pm 0.73$ & \cellcolor{bestcell} \textbf{$2.79 \pm 0.86$} \\
Letter V & $4.74 \pm 1.11$ & \cellcolor{bestcell} \textbf{$1.33 \pm 0.57$} \\
Letter W & $4.79 \pm 0.59$ & \cellcolor{bestcell} \textbf{$0.95 \pm 1.22$} \\
Letter X & $3.24 \pm 0.97$ & \cellcolor{bestcell} \textbf{$2.68 \pm 0.94$} \\
Letter Y & $4.40 \pm 0.64$ & \cellcolor{bestcell} \textbf{$2.77 \pm 0.73$} \\
Letter Z & $5.81 \pm 1.21$ & \cellcolor{bestcell} \textbf{$2.59 \pm 1.05$} \\
\bottomrule
\end{tabular}
\caption{Per-shape comparison against the classical Enum+GA baseline for the 6-bar family with GA budget 6/20. R-APS reports the best mean trajectory distance per shape over model+mode groups. Best per row is \textbf{highlighted}.}
\label{tab:enum_ga_full_6bar_6_20}
\end{table}

\begin{table}[t]
\centering
\small
\begin{tabular}{@{}l rr@{}}
\toprule
Shape & \textbf{Enum+GA Distance\,$\downarrow$} & \textbf{R-APS (Full) Distance\,$\downarrow$} \\
\midrule
Figure-8 & \cellcolor{bestcell} \textbf{$5.84 \pm 0.31$} & $6.28 \pm 0.44$ \\
Circle & $3.12 \pm 1.01$ & \cellcolor{bestcell} \textbf{$0.23 \pm 0.51$} \\
Ellipse & $1.83 \pm 0.28$ & \cellcolor{bestcell} \textbf{$1.20 \pm 0.40$} \\
Line & $2.40 \pm 0.76$ & \cellcolor{bestcell} \textbf{$1.34 \pm 1.07$} \\
NACA & $0.41 \pm 0.10$ & \cellcolor{bestcell} \textbf{$0.11 \pm 0.07$} \\
Parabola & $866.62 \pm 14.90$ & \cellcolor{bestcell} \textbf{$324.32 \pm 33.08$} \\
Letter A & \cellcolor{bestcell} \textbf{$1.05 \pm 0.07$} & $1.92 \pm 0.97$ \\
Letter B & \cellcolor{bestcell} \textbf{$1.57 \pm 0.16$} & $3.03 \pm 0.99$ \\
Letter C & \cellcolor{bestcell} \textbf{$0.65 \pm 0.11$} & $0.94 \pm 0.68$ \\
Letter D & $1.54 \pm 0.30$ & \cellcolor{bestcell} \textbf{$1.45 \pm 0.75$} \\
Letter E & $1.98 \pm 0.26$ & \cellcolor{bestcell} \textbf{$1.77 \pm 0.74$} \\
Letter F & $1.43 \pm 0.10$ & \cellcolor{bestcell} \textbf{$0.75 \pm 1.41$} \\
Letter G & \cellcolor{bestcell} \textbf{$1.65 \pm 0.18$} & $2.10 \pm 1.19$ \\
Letter H & $1.57 \pm 0.13$ & \cellcolor{bestcell} \textbf{$1.17 \pm 0.69$} \\
Letter I & $1.49 \pm 0.29$ & \cellcolor{bestcell} \textbf{$0.94 \pm 0.33$} \\
Letter J & \cellcolor{bestcell} \textbf{$1.01 \pm 0.17$} & $5.74 \pm 0.70$ \\
Letter K & $1.38 \pm 0.12$ & \cellcolor{bestcell} \textbf{$0.70 \pm 1.03$} \\
Letter L & \cellcolor{bestcell} \textbf{$1.12 \pm 0.18$} & $2.78 \pm 0.70$ \\
Letter M & \cellcolor{bestcell} \textbf{$1.82 \pm 0.19$} & $5.20 \pm 1.05$ \\
Letter N & $1.54 \pm 0.14$ & \cellcolor{bestcell} \textbf{$1.41 \pm 0.73$} \\
Letter O & $1.13 \pm 0.32$ & \cellcolor{bestcell} \textbf{$0.87 \pm 0.83$} \\
Letter P & \cellcolor{bestcell} \textbf{$1.03 \pm 0.11$} & $1.99 \pm 0.73$ \\
Letter Q & $2.01 \pm 0.41$ & \cellcolor{bestcell} \textbf{$1.55 \pm 1.34$} \\
Letter R & $1.22 \pm 0.14$ & \cellcolor{bestcell} \textbf{$0.62 \pm 0.78$} \\
Letter S & \cellcolor{bestcell} \textbf{$1.47 \pm 0.15$} & $1.68 \pm 1.13$ \\
Letter T & \cellcolor{bestcell} \textbf{$1.52 \pm 0.37$} & $5.57 \pm 2.15$ \\
Letter U & \cellcolor{bestcell} \textbf{$1.43 \pm 0.16$} & $2.79 \pm 0.86$ \\
Letter V & $1.44 \pm 0.12$ & \cellcolor{bestcell} \textbf{$1.33 \pm 0.57$} \\
Letter W & $1.83 \pm 0.15$ & \cellcolor{bestcell} \textbf{$0.95 \pm 1.22$} \\
Letter X & \cellcolor{bestcell} \textbf{$1.30 \pm 0.14$} & $2.68 \pm 0.94$ \\
Letter Y & \cellcolor{bestcell} \textbf{$1.02 \pm 0.13$} & $2.77 \pm 0.73$ \\
Letter Z & \cellcolor{bestcell} \textbf{$2.17 \pm 0.16$} & $2.59 \pm 1.05$ \\
\bottomrule
\end{tabular}
\caption{Per-shape comparison against the classical Enum+GA baseline for the 6-bar family with GA budget 60/300. R-APS reports the best mean trajectory distance per shape over model+mode groups. Best per row is \textbf{highlighted}.}
\label{tab:enum_ga_full_6bar_60_300}
\end{table}

\subsection{Validation-Critic Failure Attribution: Per-Shape Detail}
\label{app:critic_attribution}
\begin{figure}[h]
\centering
\includegraphics[width=0.75\textwidth]{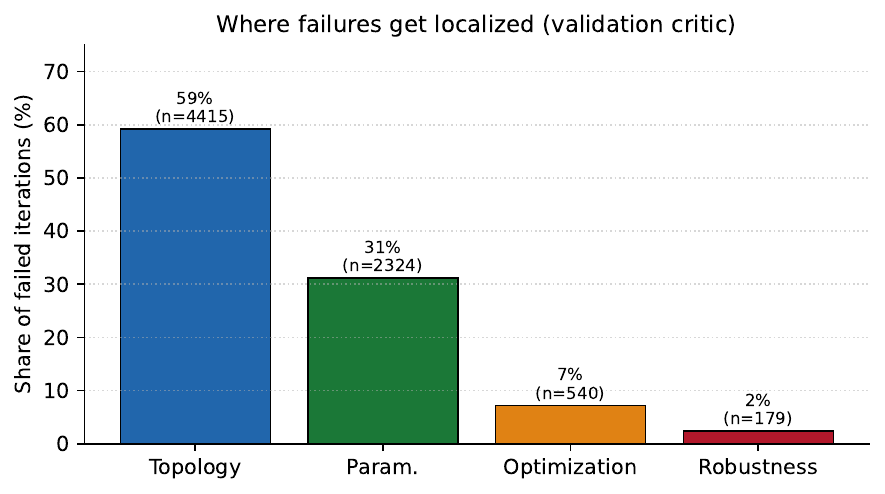}
\caption{Distribution of failure attributions across the four R-APS validation critics. Most failures are absorbed at the cheapest topology critic; only \QualRobustPct{} leak through to the expensive adversarial screen. Visual counterpart of Table~\ref{tab:qual_evidence}.}
\label{fig:stage_failure_distribution}
\end{figure}

\begin{figure}[h]
\centering
\includegraphics[width=0.75\textwidth]{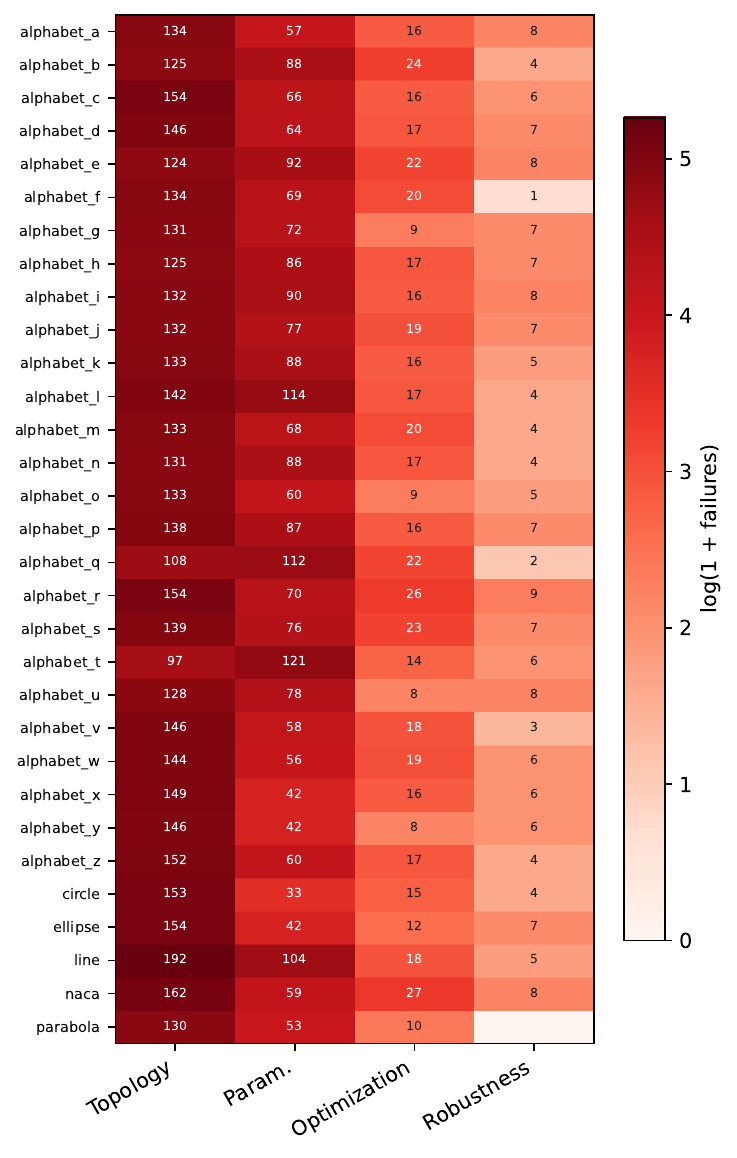}
\caption{Failure counts per (shape, failed-stage) cell, log-normalized for readability. Read rows for which shapes drive failure at which critic; topology dominates on essentially every shape.}
\label{fig:per_shape_failure_heatmap}
\end{figure}

\begin{figure}[h]
\centering
\includegraphics[width=0.85\textwidth]{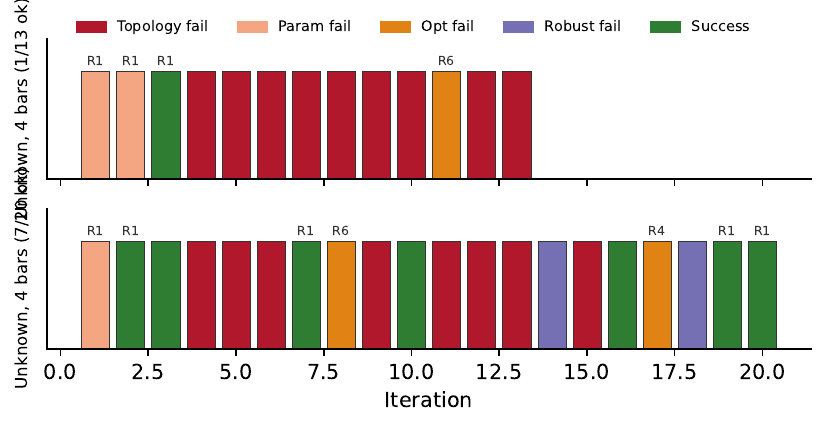}
\caption{Per-iteration stage-outcome strips for two representative baseline runs. Coloured bars show the failed stage per iteration; green bars are archive-admission iterations. ``R$k$'' annotations indicate the number of selective refinements triggered at each iteration. Concrete instance of the targeted-refinement pattern aggregated in Fig.~\ref{fig:refinement_recovery_main}.}
\label{fig:iteration_trace_case}
\end{figure}

\emph{Deepens \S\ref{sec:q1_intrastage} (Q1) by giving per-shape failure counts at each critic, the visual distribution behind those counts, and a per-iteration trace case study for two representative runs.}
The aggregate distribution (Fig.~\ref{fig:stage_failure_distribution}) shows the cost-ordered absorption qualitatively; the per-shape table (Table~\ref{tab:qual_evidence_full}) and heatmap (Fig.~\ref{fig:per_shape_failure_heatmap}) show that no single shape drives the aggregate, the topology critic dominates absorption on essentially every shape; the iteration-trace case study (Fig.~\ref{fig:iteration_trace_case}) makes the per-iteration routing of selective refinement concrete.

\begin{table}[t]
\centering
\footnotesize
\begin{tabular}{@{}l rrrr r r@{}}
\toprule
\textbf{Shape} & \textbf{Topo.} & \textbf{Param.} & \textbf{Opt.} & \textbf{Robust.} & \textbf{Total} & \textbf{Recov.\,$\uparrow$} \\
\midrule
Circle & $153$ & $33$ & $15$ & $4$ & $205$ & $48\%$ \\
Ellipse & $154$ & $42$ & $12$ & $7$ & $215$ & $67\%$ \\
LB & $68$ & $49$ & $10$ & $4$ & $131$ & $67\%$ \\
Line & $254$ & $127$ & $24$ & $5$ & $410$ & $54\%$ \\
NACA & $162$ & $59$ & $27$ & $8$ & $256$ & $68\%$ \\
Parabola & $68$ & $30$ & $4$ & $0$ & $102$ & $65\%$ \\
Letter A & $134$ & $57$ & $16$ & $8$ & $215$ & $39\%$ \\
Letter B & $125$ & $88$ & $24$ & $4$ & $241$ & $62\%$ \\
Letter C & $154$ & $66$ & $16$ & $6$ & $242$ & $63\%$ \\
Letter D & $146$ & $64$ & $17$ & $7$ & $234$ & $62\%$ \\
Letter E & $124$ & $92$ & $22$ & $8$ & $246$ & $72\%$ \\
Letter F & $134$ & $69$ & $20$ & $1$ & $224$ & $67\%$ \\
Letter G & $131$ & $72$ & $9$ & $7$ & $219$ & $74\%$ \\
Letter H & $125$ & $86$ & $17$ & $7$ & $235$ & $71\%$ \\
Letter I & $132$ & $90$ & $16$ & $8$ & $246$ & $66\%$ \\
Letter J & $132$ & $77$ & $19$ & $7$ & $235$ & $82\%$ \\
Letter K & $133$ & $88$ & $16$ & $5$ & $242$ & $72\%$ \\
Letter L & $142$ & $114$ & $17$ & $4$ & $277$ & $81\%$ \\
Letter M & $133$ & $68$ & $20$ & $4$ & $225$ & $52\%$ \\
Letter N & $131$ & $88$ & $17$ & $4$ & $240$ & $66\%$ \\
Letter O & $133$ & $60$ & $9$ & $5$ & $207$ & $66\%$ \\
Letter P & $138$ & $87$ & $16$ & $7$ & $248$ & $58\%$ \\
Letter Q & $108$ & $112$ & $22$ & $2$ & $244$ & $69\%$ \\
Letter R & $154$ & $70$ & $26$ & $9$ & $259$ & $67\%$ \\
Letter S & $139$ & $76$ & $23$ & $7$ & $245$ & $64\%$ \\
Letter T & $97$ & $121$ & $14$ & $6$ & $238$ & $74\%$ \\
Letter U & $128$ & $78$ & $8$ & $8$ & $222$ & $64\%$ \\
Letter V & $146$ & $58$ & $18$ & $3$ & $225$ & $50\%$ \\
Letter W & $144$ & $56$ & $19$ & $6$ & $225$ & $65\%$ \\
Letter X & $149$ & $42$ & $16$ & $6$ & $213$ & $61\%$ \\
Letter Y & $146$ & $42$ & $8$ & $6$ & $202$ & $58\%$ \\
Letter Z & $152$ & $60$ & $17$ & $4$ & $233$ & $65\%$ \\
\bottomrule
\end{tabular}
\caption{Per-shape breakdown of where R-APS validation gates catch failures (Topology / Parameterisation / Optimization / Robustness). \textbf{Recov.}\ = fraction of episodes that recovered to a successful archive insertion after at least one failed iteration. The pattern repeats across all shapes: most failures are localised to the topology gate before they reach expensive downstream stages.}
\label{tab:qual_evidence_full}
\end{table}

\subsection{Adversarial Discoveries}
\label{app:adversarial_disc}
\emph{Deepens \S\ref{sec:q2_counterfactual} (Q2) by listing the top-5 most fragile R-APS designs surfaced by the directed adversary, with nominal vs.\ worst-case Chamfer.}
The gap on the most extreme entry (e.g., Line: 0.89 nominal $\to$ \AdversarialGapMax{} worst-case) is the kind of brittle-corner finding that uniform-perturbation baselines miss; the median nominal-vs-worst gap across these top-5 cases is \AdversarialGapMedian{}, the canonical certificate-tightness number reported in the main text.

\begin{table}[t]
\centering\small
\begin{tabular}{@{}llrr@{}}
\toprule
Shape & Preset & Nominal traj. & Worst-case robustness \\
\midrule
line & R-APS (Full) & $0.89$ & $5.114$ \\
alphabet\_j & - Adv+Meta & $1.09$ & $3.865$ \\
alphabet\_s & - Adv+Meta & $2.78$ & $3.439$ \\
circle & - Adv+Meta & $0.40$ & $3.019$ \\
alphabet\_u & R-APS (Full) & $1.58$ & $2.620$ \\
\bottomrule
\end{tabular}
\caption{Top-5 adversarial discoveries: runs with the largest worst-case robustness objective. Nominal is the Pareto-best trajectory error; worst-case is the Monte-Carlo-derived robustness objective produced by the Adversary agent.}
\label{tab:adversarial_discoveries}
\end{table}

\subsection{Meta-Learning: Per-Shape Trace and Sensitivity}
\label{app:metalearn_sens}
\begin{figure}[h]
\centering
\includegraphics[width=0.95\textwidth]{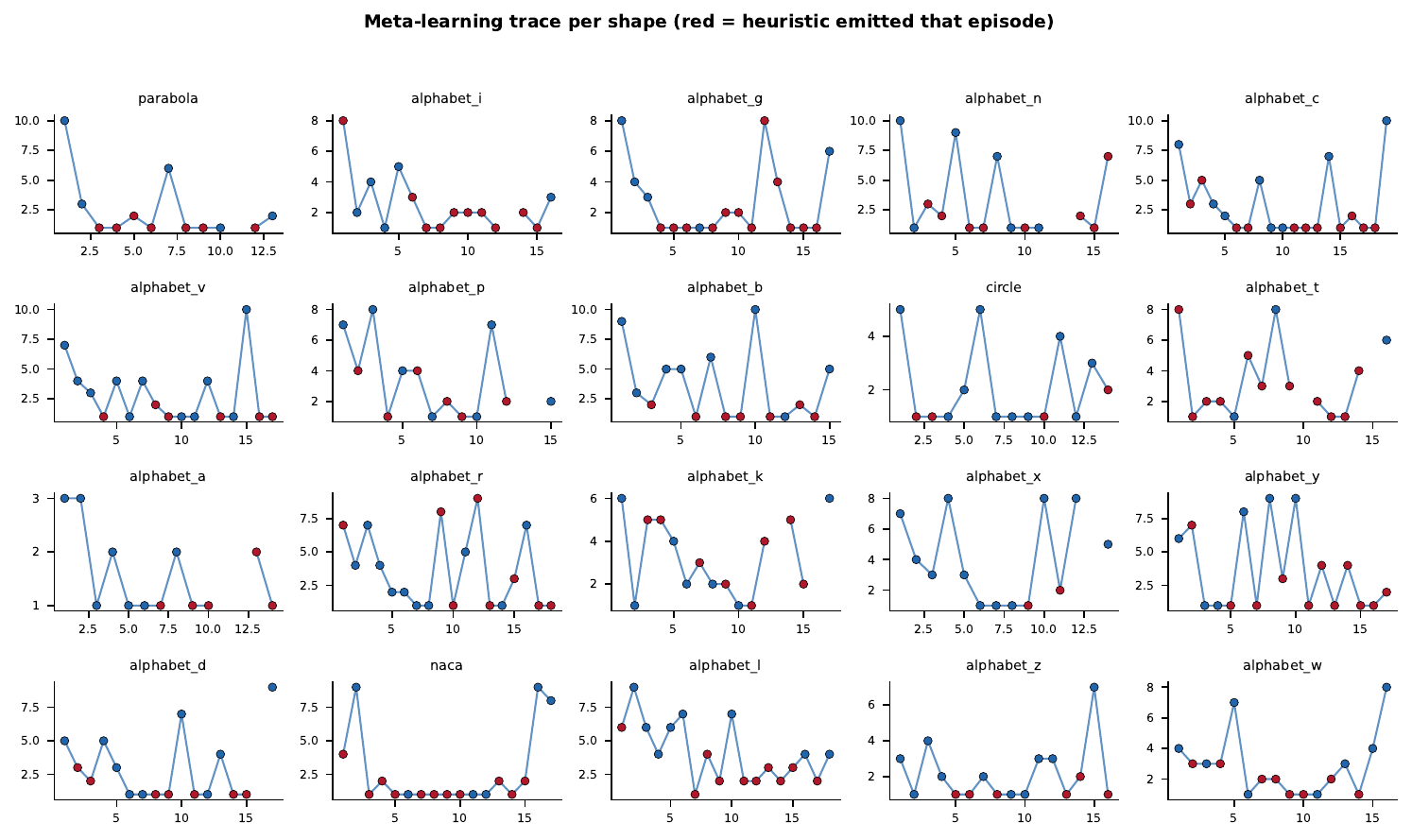}
\caption{Per-shape meta-learning trace: iterations-to-first-success across episodes on the same shape. Red markers flag episodes in which a new heuristic was emitted; blue markers flag episodes where the run consumed existing heuristics without emitting new ones. The broad downward trend across shapes is the visual counterpart of Table~\ref{tab:meta_learning_trace}.}
\label{fig:meta_learning_qualitative}
\end{figure}

\emph{Deepens \S\ref{sec:q3_metainductive} (Q3) by giving the per-shape breakdown behind the aggregate acceleration and by housing the corpus-order and library-size sensitivity regressions referenced in the main text.}
For each shape with at least two baseline episodes, Table~\ref{tab:meta_learning_trace} reports iterations-to-first-success on the first episode (Ep1) and the mean over episodes~4$+$, along with the total new heuristics emitted and invalidated across all episodes on that shape; Fig.~\ref{fig:meta_learning_qualitative} visualizes the same information as small multiples (one mini-panel per shape, red markers flagging new-heuristic episodes), so the broad downward trend is visible across shapes rather than concentrated on a few lucky ones.

\begin{table}[h!]
\centering\scriptsize
\setlength{\tabcolsep}{4pt}
\begin{tabular}{@{}lrrrrrr@{}}
\toprule
Shape & $n$ Ep. & Ep1 iters & Ep4+ mean & Accel. & New heur. & Invalidated \\
\midrule
parabola & 15 & 10 & 1.8 & $+82\%$ \textcolor{green!60!black}{$\downarrow$} & 13 & 0 \\
alphabet\_i & 16 & 8 & 2.0 & $+75\%$ \textcolor{green!60!black}{$\downarrow$} & 19 & 2 \\
alphabet\_g & 17 & 8 & 2.2 & $+72\%$ \textcolor{green!60!black}{$\downarrow$} & 22 & 0 \\
alphabet\_n & 16 & 10 & 3.0 & $+70\%$ \textcolor{green!60!black}{$\downarrow$} & 16 & 0 \\
alphabet\_c & 19 & 8 & 2.4 & $+70\%$ \textcolor{green!60!black}{$\downarrow$} & 26 & 1 \\
alphabet\_v & 17 & 7 & 2.4 & $+66\%$ \textcolor{green!60!black}{$\downarrow$} & 11 & 0 \\
alphabet\_p & 15 & 7 & 2.5 & $+64\%$ \textcolor{green!60!black}{$\downarrow$} & 10 & 0 \\
alphabet\_b & 17 & 9 & 3.2 & $+64\%$ \textcolor{green!60!black}{$\downarrow$} & 17 & 0 \\
circle & 14 & 5 & 2.0 & $+60\%$ \textcolor{green!60!black}{$\downarrow$} & 11 & 0 \\
alphabet\_t & 16 & 8 & 3.3 & $+59\%$ \textcolor{green!60!black}{$\downarrow$} & 28 & 2 \\
alphabet\_a & 16 & 3 & 1.3 & $+56\%$ \textcolor{green!60!black}{$\downarrow$} & 9 & 0 \\
alphabet\_r & 19 & 7 & 3.1 & $+55\%$ \textcolor{green!60!black}{$\downarrow$} & 13 & 0 \\
alphabet\_k & 19 & 6 & 3.1 & $+49\%$ \textcolor{green!60!black}{$\downarrow$} & 14 & 0 \\
alphabet\_x & 15 & 7 & 3.8 & $+46\%$ \textcolor{green!60!black}{$\downarrow$} & 4 & 0 \\
alphabet\_y & 17 & 6 & 3.3 & $+45\%$ \textcolor{green!60!black}{$\downarrow$} & 18 & 0 \\
alphabet\_d & 17 & 5 & 2.8 & $+45\%$ \textcolor{green!60!black}{$\downarrow$} & 11 & 1 \\
naca & 19 & 4 & 2.3 & $+43\%$ \textcolor{green!60!black}{$\downarrow$} & 25 & 0 \\
alphabet\_l & 18 & 6 & 3.5 & $+41\%$ \textcolor{green!60!black}{$\downarrow$} & 21 & 0 \\
alphabet\_z & 16 & 3 & 2.0 & $+33\%$ \textcolor{green!60!black}{$\downarrow$} & 15 & 0 \\
alphabet\_w & 16 & 4 & 2.8 & $+31\%$ \textcolor{green!60!black}{$\downarrow$} & 12 & 0 \\
alphabet\_q & 15 & 3 & 2.5 & $+17\%$ \textcolor{green!60!black}{$\downarrow$} & 10 & 0 \\
alphabet\_j & 18 & 4 & 3.6 & $+10\%$ \textcolor{green!60!black}{$\downarrow$} & 21 & 0 \\
alphabet\_f & 16 & 3 & 2.8 & $+8\%$ \textcolor{green!60!black}{$\downarrow$} & 13 & 0 \\
alphabet\_h & 16 & 2 & 2.1 & $-5\%$ \textcolor{red}{$\uparrow$} & 19 & 0 \\
alphabet\_m & 15 & 2 & 2.2 & $-10\%$ \textcolor{red}{$\uparrow$} & 6 & 0 \\
alphabet\_u & 16 & 2 & 3.0 & $-50\%$ \textcolor{red}{$\uparrow$} & 13 & 0 \\
alphabet\_o & 17 & 2 & 3.5 & $-73\%$ \textcolor{red}{$\uparrow$} & 10 & 0 \\
alphabet\_s & 19 & 1 & 2.0 & $-100\%$ \textcolor{red}{$\uparrow$} & 25 & 0 \\
line & 28 & 1 & 2.2 & $-121\%$ \textcolor{red}{$\uparrow$} & 37 & 0 \\
ellipse & 15 & 1 & 2.5 & $-150\%$ \textcolor{red}{$\uparrow$} & 12 & 0 \\
alphabet\_e & 16 & 1 & 2.9 & $-192\%$ \textcolor{red}{$\uparrow$} & 26 & 0 \\
\bottomrule
\end{tabular}
\caption{Per-shape meta-learning trace. For each shape with $\ge 2$ baseline episodes, we report iterations-to-first-success on the shape's first episode (Ep1) and the mean over episodes 4+, along with the total new heuristics emitted and invalidated across all episodes on that shape. Rows sorted by acceleration (most-accelerated at top). Positive acceleration (green $\downarrow$) = fewer iterations needed on repeat episodes; negative (red $\uparrow$) = slower on repeats. The table lets the reader see which shapes drive the aggregate meta-inductive effect and which resist or plateau.}
\label{tab:meta_learning_trace}
\end{table}

\begin{figure}[h]
\centering
\includegraphics[width=0.85\textwidth]{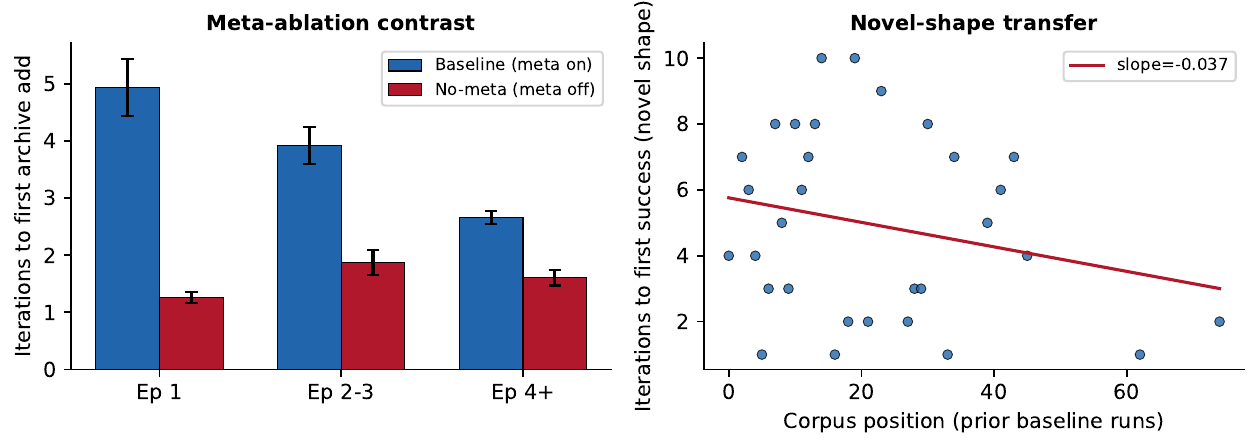}
\caption{Cross-shape heuristic transfer signal. Each rule ID is plotted by the number of distinct shapes whose digests cite it; \HeuristicAppliedMulti{} of \HeuristicRuleIds{} rule IDs appear in $\geq\!2$ shapes, the closest direct evidence of cross-shape application that the meta-learning digests permit. Visual counterpart of Table~\ref{tab:heuristic_transfer}.}
\label{fig:meta_learning_transfer}
\end{figure}

\subsection{Representative and Extended Heuristic Sample}
\label{app:heuristics_sample}
\emph{Deepens \S\ref{sec:q3_metainductive} (Q3) by giving concrete examples of the rules the inter-episode meta-inductive mode produces.}
Table~\ref{tab:heuristics} samples four typical rules across shape categories; Table~\ref{tab:heuristics_full} extends the sample with eight additional rules covering Letters A--H and the standard curves, so the reader can audit the rule library directly rather than only via aggregate counts.
The non-monotonic invalidation cases corresponding to these rules are documented inline in the main-text Table~\ref{tab:invalidated_heuristics}.

\begin{table}[t]
\centering
\footnotesize
\begin{tabular}{@{}p{0.12\textwidth} p{0.81\textwidth}@{}}
\toprule
\textbf{Source shape} & \textbf{Learned heuristic (paraphrased)} \\
\midrule
\textit{Figure-8} & WHEN motion sequence requires n identical VS segments (VSn) AND n $\geq$ 2, THEN use a symmetric dual 4-bar topology with n segments and 2 guard crossings to minimize label entropy an... \\
\textit{Alphabet (mixed)} & WHEN designing a 4-bar mechanism for a circular trajectory with heading change $\geq$ 45$^\circ$ AND n\_region\_crossings > 0 in symbolic\_adversarial analysis, THEN add a phase-correction dya... \\
\textit{Circle} & WHEN generating a circular trajectory and using a 4-bar topology with a dominant 'Very Sharp Turn' motion label (label\_string: 'VS-VS') THEN set coupler offset to 0.4--0.5 of the d... \\
\textit{Ellipse} & WHEN designing a 4-bar topology for a circular trajectory with heading change $\geq$45$^\circ$ AND the motion requires 2 consecutive segments of the same heading label (e.g., VS-VS), THEN use... \\
\bottomrule
\end{tabular}
\caption{Representative heuristics extracted by the meta-learning agent. The lifecycle manager produced $\,\QualNewHeuristics{}\,$ new heuristics and \emph{explicitly invalidated} $\,\QualInvalidatedHeuristics{}\,$ prior rules, distinguishing the catalogue from Voyager/ExpeL-style monotonic accumulators. Full sample in Appendix Table~\ref{tab:heuristics_full}.}
\label{tab:heuristics}
\end{table}

\begin{table}[t]
\centering
\footnotesize
\begin{tabular}{@{}p{0.12\textwidth} p{0.81\textwidth}@{}}
\toprule
\textbf{Source shape} & \textbf{Learned heuristic (paraphrased)} \\
\midrule
\textit{Figure-8} & WHEN motion sequence requires n identical VS segments (VSn) AND n $\geq$ 2, THEN use a symmetric dual 4-bar topology with n segments and 2 guard crossings to minimize label entropy and enable phase-locked transitions BECAUSE symmetric dual 4-b... \\
\textit{Circle} & WHEN generating a circular trajectory and using a 4-bar topology with a dominant 'Very Sharp Turn' motion label (label\_string: 'VS-VS') THEN set coupler offset to 0.4--0.5 of the dyad length (measured from dyad pivot) to minimize curvature ... \\
\textit{Ellipse} & WHEN designing a 4-bar topology for a circular trajectory with heading change $\geq$45$^\circ$ AND the motion requires 2 consecutive segments of the same heading label (e.g., VS-VS), THEN use a phase-shifted coupler offset (0.333--0.666 of dyad length)... \\
\textit{Letter A} & WHEN a mechanism topology exhibits 'VS-G-VS-VS-VS' symbolic label string with 5 segments and 1 guard crossing in nominal and adversarial configurations, AND the optimizer fails with 'optimizer\_infeasible\_solution' at stage OPT, THEN enforc... \\
\textit{Letter B} & WHEN 4-bar topology fails optimizer\_infeasible\_solution at OPT stage THEN increase intermediate link length by 15--20\% BECAUSE this resolves Grashof violation and improves feasibility for closed-loop kinematic chains in 4-bar mechanisms. \\
\textit{Letter C} & WHEN synthesizing motion sequences with multiple sharp turns ($\geq$2 consecutive VS segments) THEN add a fifth link (5-bar) to the SymmetricCShape topology to enable 3-phase turn handling BECAUSE the 5-bar configuration provides additional deg... \\
\textit{Letter D} & WHEN a mechanism must produce a motion transition between gentle turns (2$^\circ$ $\leq$ heading change < 30$^\circ$) and very sharp turns (heading change $\geq$ 45$^\circ$) THEN enforce a 1:1 link ratio between the two outer links of a symmetric 4-bar topology BECAUSE ... \\
\textit{Letter E} & WHEN motion requires transitions between multiple heading-change regimes (e.g., VS $\rightarrow$ G $\rightarrow$ ST $\rightarrow$ VS) AND the topology has 4 or more links, THEN use a symmetric 5-bar topology with alternating link lengths to minimize label entropy and maximiz... \\
\textit{Letter F} & WHEN topology is a 4-bar with 2 dyads (F4bar\_phase\_shift) AND initial optimization fails with optimizer\_infeasible\_solution (BFGS/PSO), THEN increase intermediate link lengths by 5--10\% and enforce symmetry between dyad connections BECAUSE ... \\
\textit{Letter G} & WHEN the target trajectory requires 2$\times$ Very Sharp Turn / U-turn (heading change $\geq$ 45$^\circ$) AND the mechanism has 4 bars with asymmetric coupler offset, THEN set coupler offset to 33.3\% of the dyad length (from dyad pivot) to minimize region cr... \\
\textit{Letter H} & WHEN a mechanism requires 2 consecutive sharp heading changes ($\geq$45$^\circ$) and a straight segment in the nominal motion, THEN use a 4-bar topology with two symmetric dyads (each dyad connected to the same ground link) to minimize run-length imba... \\
\bottomrule
\end{tabular}
\caption{Extended sample of distinct heuristics extracted by the meta-learning agent across episodes. Each rule is paraphrased and truncated for space; the full database stores $\,1015\,$ rules extracted by the lifecycle manager.}
\label{tab:heuristics_full}
\end{table}

\subsection{Pairwise Statistical Significance}
\label{app:significance}
\emph{Deepens \S\ref{sec:q1_intrastage}, \S\ref{sec:q2_counterfactual}, and \S\ref{sec:q3_metainductive} by reporting Mann--Whitney $U$ tests for every preset pair on the three primary metrics (trajectory objective, robustness objective, $\Delta$Bars), with 95\% bootstrap confidence intervals.}
The ablation-induced shifts referenced in the main text (\RobustnessSpeedup{}$\times$ robustness degradation when adversary+meta is removed; $\AdhDegradationPct{}\%$ adherence degradation when selective refinement is removed) are not attributable to noise on any of the three objectives at conventional thresholds.

\begin{table}[t]
\centering\scriptsize
\setlength{\tabcolsep}{4pt}
\begin{tabular}{@{}llrrrl@{}}
\toprule
Metric & Contrast & Mean [95\% bootstrap CI] (A) & Mean [95\% bootstrap CI] (B) & $U$ & $p$ \\
\midrule
Trajectory obj. & R-APS (Full) vs - Sel. Refine & $17.708\ [10.862,25.377]$ & $26.647\ [14.417,41.052]$ & $91383$ & $0.007$ \\
Trajectory obj. & R-APS (Full) vs - Adv+Meta & $17.708\ [10.862,25.377]$ & $19.773\ [6.358,35.730]$ & $66095$ & $< 10^{-4}$ \\
Robustness obj. & R-APS (Full) vs - Sel. Refine & $0.118\ [0.090,0.149]$ & $0.091\ [0.072,0.113]$ & $87306$ & $0.140$ \\
Robustness obj. & R-APS (Full) vs - Adv+Meta & $0.118\ [0.090,0.149]$ & $0.459\ [0.376,0.554]$ & $15850$ & $< 10^{-4}$ \\
$\Delta$Bars & R-APS (Full) vs - Sel. Refine & $1.736\ [1.606,1.880]$ & $2.189\ [2.036,2.347]$ & $66039$ & $< 10^{-4}$ \\
$\Delta$Bars & R-APS (Full) vs - Adv+Meta & $1.736\ [1.606,1.880]$ & $2.235\ [2.026,2.444]$ & $38512$ & $< 10^{-4}$ \\
\bottomrule
\end{tabular}
\caption{Pairwise Mann--Whitney $U$ tests between ablation presets on the three primary metrics (lower is better for all). 95\% confidence intervals are percentile bootstraps with 1000 resamples. Listed for supplementary statistical rigor; see Section~\ref{sec:experiments} for the corresponding mean $\pm$ SEM tables.}
\label{tab:significance}
\end{table}

\section{Prompts}
\label{app:prompts}

This section reproduces every prompt the R-APS multi-agent method constructs.

\roletag{roleOrange}{Persona / System Role},
\roletag{roleSky}{Epistemic Task},
\roletag{rolePurple}{Context Grounding},
\roletag{roleGreen}{Reasoning Role},
\roletag{roleBlue}{Output Format},
\roletag{roleVermillion}{Critical Reminders}.
The high-level account of which reasoning mode each prompt is responsible for is given in Section~\ref{sec:modes} (Table~\ref{tab:reasoning_modes}) and Appendix~\ref{app:agents} (Table~\ref{tab:reasoning_type}).

\subsection{Topology Agent (Designer, abductive mode)}
\label{app:prompt_topology}

\begin{promptpart}{roleOrange}{Persona / System Role}
You are an expert mechanical engineer specializing in linkage design and kinematics. Your expertise spans mechanism theory, constraint analysis, and novel topology discovery.
\end{promptpart}

\begin{promptpart}{roleSky}{Epistemic Task}
Design a TOPOLOGY (kinematic structure) to achieve the target trajectory. You may create ANY mechanism structure, including:
- Standard linkages (4-bar, 6-bar, 8-bar, N-bar)
- Slider-crank mechanisms with arbitrary configurations
- Hybrid combinations of the above
- Novel arrangements you invent
- Multi-module mechanisms (e.g., two 4-bars in series)

YOU ARE NOT LIMITED to predefined mechanism types. Design from first principles. OUTPUT ONLY A VALID JSON OBJECT.
\end{promptpart}

\begin{promptpart}{rolePurple}{Context Grounding}
target trajectory: {target_trajectory}

TARGET TRAJECTORY ANCHOR POINTS (MUST PASS THROUGH):
The synthesized trajectory must pass through (or stay very close to) these target anchor points:
{dataset_sample_points}

CURRENT OPTIMIZATION PRIORITIES (weight vector w):
- Trajectory accuracy (w1={w1}): {priority_description_1}
- Robustness (w2={w2}): {priority_description_2}

ARCHIVE CONTEXT:
- Current archive size: {archive_size} non-dominated designs
- Successful topology summaries: {archive_summary}

LEARNED DESIGN HEURISTICS: {learned_heuristics}
PREVIOUSLY FAILED TOPOLOGIES (avoid similar structures): {failed_topologies}
DESIGN DIGEST EXEMPLARS (meta-learning): {design_digest}
EXCLUDE LIST (repeated failures): {exclude_list}
ITERATION: - Current iteration: {iteration_number}/{max_iterations}
TRAJECTORY SHAPE SPECIFICATION: {shape_guidance}
SIMULATOR CONFIGURATION: <populated by the active simulator adapter at runtime>
LINK/BAR COUNT GOAL: {link_goal_instruction}

ADDITIONAL NOTES FOR THE TOPOLOGY AGENT:
- Although the weight vector is 2D, your provided rationale should explicitly mention topology-specific secondary metrics (DOF, joint complexity, workspace clearance and collision risk, novelty) so the topology designer can trade off these aspects when optimizing.
- If you recommend targeting a high-accuracy & high-robustness region (rare), suggest conservative topology families first (e.g., symmetric, phase-locked designs or constrained dyads) because they tend to be more buildable.
- Always provide at least one topology family that is *likely* to be feasible with DOF = 1 for the chosen weight direction; if none exists, state that the region likely requires multi-DOF or actively constrained mechanisms and note the implication.
\end{promptpart}

\begin{promptpart}{roleGreen}{Reasoning Role / Method}
1. Locate sparse regions on the Pareto frontier:
   - Identify gaps between neighbors in f1--f2 space.
   - Note underrepresented objective combinations (e.g., high-accuracy + high-robustness).
   - Identify extreme regions not explored (max accuracy, max constraints).
2. Account for topology diversity:
   - Which topology families (4-bar, 6-bar, slider-crank, hybrid, novel) occupy which regions?
   - Are certain topologies concentrated in crowded regions? Are some topology families absent from some regions?
   - Prefer suggesting regions where a different topology family is likely to improve coverage.
3. Consider topological feasibility and design cost:
   - Heavily favor regions where feasible topologies exist (DOF = 1 for single-input designs), but also highlight risky extremes that may require high DOF or extra constraints.
4. Exploration strategy by iteration phase:
   - Early (0--30%): push toward extremes to establish boundaries and sample diverse topologies.
   - Middle (30--70%): fill gaps, prioritize underexplored trade-offs (mix novelty + feasibility).
   - Late (70--100%): refine and exploit, improve local Pareto density and fidelity.
5. Weight-vector selection rules:
   - Choose w = [w1, w2] where w1,w2 >= 0 and ||w||_2 = 1.
   - w must point toward an underexplored region (justify numerically and qualitatively).
6. Topology guidance:
   - Recommend 2--4 topology families or concrete topology suggestions likely to perform well.
   - Mention topology-specific knobs (link aspect ratio, coupler offset, grounded joints, dyads, phase-locking).
7. Risk & feasibility notes:
   - State expected difficulties (singularities, assembly complexity, tolerance sensitivity).
   - Provide confidence score (0--1) for the chosen direction yielding useful new Pareto points.
\end{promptpart}

\begin{promptpart}{roleBlue}{Output Format (JSON schema)}
{
  "topology_description": {
    "name": "descriptive name",
    "structure": {
      "links":  {"count": 0, "description": "narrative description of link arrangement"},
      "joints": {"simple": 0, "fixed": 0},
      "grounding": "which links/joints are fixed to ground frame",
      "input":  "which joint/link receives driving motion",
      "output": "description of output point/joint"
    },
    "mobility_analysis": {
      "gruebler_dof": "computed expression or number",
      "target_dof": 1,
      "verification": "short text: is DOF = 1 achieved? any extra constraints?"
    },
    "kinematic_principle": "short description of how the structure generates the target trajectory"
  },
  "config": <simulator-specific config example, populated at runtime>,
  "design_rationale": {
    "why_this_structure": "first-principles reasoning for topology choice",
    "trajectory_mapping": "how structure features map to trajectory requirements",
    "comparison_to_standard": "how this differs from/improves standard mechanisms",
    "innovation_aspect": "what is novel about this design"
  },
  "expected_capabilities": {
    "trajectory_features_achievable": ["list", "of", "features"],
    "objectives_addressed": {
      "accuracy":    "how structure enables trajectory fidelity",
      "constraints": "how constraints are satisfied"
    }
  },
  "novelty_and_diversity": {
    "similarity_to_archive": 0.0,
    "structural_innovation_score": 0.0,
    "description": "how this expands coverage relative to archive"
  },
  "confidence": 0.0
}
\end{promptpart}

\textit{(Two long worked examples are stored verbatim in the source; omitted here for space.)}

\subsection{Meta-Strategist Agent (Critic, meta-inductive mode -- Pareto target selection)}
\label{app:prompt_meta_strategist}

\begin{promptpart}{roleOrange}{Persona / System Role}
You are a meta-strategist directing multi-objective mechanism *topology* design exploration for a topology-generation agent.
\end{promptpart}

\begin{promptpart}{roleSky}{Epistemic Task}
Analyze the current Pareto archive of previously generated mechanism topologies and SELECT A TARGET weight vector to guide the next topology proposal toward an underexplored region of objective space.
\end{promptpart}

\begin{promptpart}{rolePurple}{Context Grounding}
CONTEXT:
- The downstream topology agent designs kinematic structures (linkages, slider-cranks, hybrids, novel topologies) that must satisfy kinematic feasibility and constructability and produce the requested TARGET TRAJECTORY (here: circular path).
- The agent outputs complete topology descriptions (links, joints, DOF analysis, parameters, constraints, expected behavior, novelty assessment).

OBJECTIVES (two primary objectives -- keep these fixed for weight selection):
1. Trajectory deviation: f1 = CD_error (lower is better) -- nominal Chamfer distance to target
2. Robustness score: f2 = ||f_nom - f_worst|| / max(||f_nom||, eps) in [0, inf) (lower is better)
   - f2 = 0 means perfectly robust; f2 = 1 means 100% degradation under adversarial perturbation
   - Threshold for 'robust' design is typically 0.05 (5% relative degradation)

Important: While weights target these two objectives, you *must* account for topology-specific secondary concerns in your reasoning (not in the weight vector): DOF and mobility, joint types, constructability, workspace validity, novelty/diversity of topology, and likely optimization difficulty.

ARCHIVE CONTEXT:
- Current archive size: {archive_size} non-dominated designs
- Successful topology summaries: {archive_summary}

PARETO COVERAGE VISUALIZATION: {coverage_analysis}
TOPOLOGY DISTRIBUTION:        {topology_distribution}
DESIGN DIGEST EXEMPLARS:      {design_digest}
LEARNED DESIGN HEURISTICS:    {learned_heuristics}
PREVIOUSLY FAILED TOPOLOGIES (avoid similar structures): {failed_topologies}

design_principles:
1. Degrees of Freedom (DOF): Design mechanisms with appropriate DOF to achieve target behavior.
2. Constraint Analysis: Ensure all constraints are satisfied and mechanism is constructible.
3. Joint Types: Use pin joints, sliding joints, or other standard mechanical joints.
4. Link Configuration: Specify ground links, input links, coupler points, and output behavior.
5. Workspace: Ensure the mechanism can move through its entire working range without collision.
6. Novel Design: Avoid copying standard mechanisms -- propose creative topologies within mechanical feasibility.

output_requirements:
1. Topology, parameters, constraints, expected behavior, feasibility assessment, novelty explanation.
\end{promptpart}

\begin{promptpart}{roleGreen}{Reasoning Role / Method}
1. Understand the Target: Carefully analyze the target trajectory or mechanism behavior.
2. Conceptualize: Consider multiple topology variations before settling on one.
3. Verify Constraints: Ensure all constraints can be satisfied with your design.
4. Check Feasibility: Can this be built? Are all ranges achievable?
5. Assess Novelty: Is this a standard mechanism or a novel variation?
6. Explain Reasoning: Justify why this specific topology achieves the target behavior.
7. Consider Trade-offs: What are the compromises in your design? Why are they acceptable?
\end{promptpart}

\begin{promptpart}{roleBlue}{Output Format (JSON schema)}
{
  "pareto_analysis": {
    "total_mechanisms": integer,
    "dominated_regions": ["description", "of", "crowded", "areas"],
    "sparse_regions":    ["description", "of", "gaps"],
    "frontier_extremes": {"max_accuracy": f1_value, "max_constraints": f2_value}
  },
  "exploration_strategy": {
    "phase":     "early|middle|late",
    "priority":  "explore_extremes|fill_gaps|refine_crowded",
    "reasoning": "why this strategy now (topology-aware)"
  },
  "target_region": {
    "description": "qualitative description of the target region in objective space",
    "objective_ranges": {"f1": [min, max], "f2": [min, max]},
    "why_sparse":        "explain why archive lacks solutions here",
    "expected_challenge":"technical reasons (DOF>1, sensitive ratios, toggles)"
  },
  "weight_vector": {
    "w": [w1, w2],
    "normalized": true,
    "interpretation": {"w1": "trajectory accuracy priority: X%", "w2": "robustness priority: Y%"},
    "target_direction": "short statement: which region this vector points to"
  },
  "expected_improvement": "what filling this gap would achieve",
  "topology_recommendation": {
    "families":            ["e.g., hybrid 4-bar + dyad", "parallel dual 4-bar", ...],
    "specific_suggestions":["..."],
    "design_knobs_to_focus":["link length ratios", "coupler offset", "ground separation", "phase"]
  },
  "confidence": float(0.0-1.0)
}
\end{promptpart}

\subsection{Meta-Learning Analyst (Meta-Analyst, inductive mode -- heuristic extraction)}
\label{app:prompt_meta_learning_analyst}

\begin{promptpart}{roleOrange}{Persona / System Role}
You are a meta-learning analyst extracting reusable design heuristics for kinematic TOPOLOGIES.
\end{promptpart}

\begin{promptpart}{roleSky}{Epistemic Task}
Analyze the archive of mechanism topologies (with special focus on recent additions) to DISCOVER NEW HEURISTICS that will guide future topology design. Produce topology-specific IF-THEN style heuristics supported by evidence and cross-topology validation.
\end{promptpart}

\begin{promptpart}{rolePurple}{Context Grounding}
CONTEXT:
- Downstream: a topology-design agent that generates full kinematic designs for a requested TARGET_TRAJECTORY. The topology agent must produce constructible 1-DOF (single-input) mechanisms where possible and report DOF/constraints when not.
- Archive entries contain: topology family, link counts, joint types, link lengths/ratios, DOF/mobility analysis, objective outcomes (f1 = CD_error/trajectory_deviation, f2 = robustness_score = ||f_nom-f_worst||/||f_nom|| in [0,inf)), motion primitive label strings, refinement history, and notes on failures.

ARCHIVE:                {archive_size}
NEW_ADDITIONS:          {new_additions}
RECENT_ENTRIES:         {recent_additions}
EXISTING_HEURISTICS:    {num_existing}
DESIGN_DIGEST:          {design_digest}
FAILURE_LOG:            {failure_log}
FAILURE_LOG_SUMMARY:    {failure_log_summary}
ITERATION:              - Current iteration: {iteration_number}/{max_iterations}
\end{promptpart}

\begin{promptpart}{roleGreen}{Reasoning Role / Method (topology-aware)}
1. Analyze successful topologies:
   - Identify parameter relationships common to high-performers (link ratios, coupler offsets, ground separation, aspect ratios).
   - Check for consistent configuration patterns across topology families (symmetry, phase-locking, dyad anchoring, guide use).
   - Note DOF-related patterns: which topologies reach archive with DOF=1 vs. those requiring extra constraints.
2. Examine refinement trajectories:
   - Start with FAILURE_LOG_SUMMARY to identify dominant failure stages/error types before drilling into raw FAILURE_LOG cases.
   - Track typical failure modes (Grashof violation, coupler singularity, workspace clipping, assembly collisions).
   - Identify corrections that repeatedly rescued designs.
3. Compare successes vs failures:
   - Use DESIGN_DIGEST as compressed representation of successful designs to identify cross-family motifs.
   - Use FAILURE_LOG as an explicit rejected-mechanism ledger with (stage_fail, error_type, topology, parameters).
   - Extract decision rules that reliably separate accepted vs rejected designs.
   - Highlight counterexamples where a commonly believed rule fails (important for invalidation).
4. Construct heuristics:
   - Form each heuristic as: WHEN <condition> THEN <recommendation> BECAUSE <kinematic reasoning>.
   - Provide evidence: list supporting mechanisms (IDs), success rate (X/Y), numeric improvements.
   - Require minimum support: heuristic must appear in >=3 independent mechanisms to be proposed.
   - Prefer cross-topology validation: mark whether the heuristic holds across >=2 topology families.
   - Specify applicability conditions (topologies, objectives, required DOF, manufacturing context).
5. Rating & conflicts:
   - Provide a confidence score [0.0-1.0] based on evidence count, cross-topology validation, effect size.
   - If heuristic conflicts with an existing heuristic, indicate conflict and propose resolution.
6. Avoid overfitting: do not promote heuristics with weak evidence; list as "discovered_patterns" instead.
7. Prioritize usefulness: prefer heuristics that reduce iterations or improve robustness/accuracy measurably.
8. Practical checks: note DOF consequences, singularities, assembly complexity per heuristic.
\end{promptpart}

\begin{promptpart}{roleBlue}{Output Format (JSON schema)}
{
  "new_heuristics": [
    {
      "id": "HEUR_NEW_X",
      "rule": "WHEN <condition> THEN <recommendation> BECAUSE <kinematic reasoning>",
      "evidence": {
        "supporting_mechanisms": ["M_id", ...],
        "success_rate": "X/Y cases",
        "average_improvement": "metric (e.g., CD_error: -15->-8)",
        "refinement_history_summary": "initial->failures->fixes (short)"
      },
      "applicability": {
        "topologies": ["4-bar","Watt-I","slider-crank",...],
        "objectives": ["Trajectory accuracy","Robustness","Constraints"],
        "conditions": ["DOF=1 required", "high tolerance", "target circular arc", ...]
      },
      "confidence": float(0-1),
      "conflicts_with": ["HEUR_# if any"],
      "kinematic_notes": "Gruebler/Grashof consequences, singularity risk, assembly note"
    }
  ],
  "heuristic_updates": [{"existing_id": "HEUR_7", "update_type": "strengthen|weaken|refine", "new_evidence": "...", "revised_confidence": float}],
  "discovered_patterns": ["promising patterns with <3 supports -- list with why and what evidence is needed"],
  "invalidated_heuristics": [{"id":"HEUR_X","reason":"counterexamples in M_ids","recommendation":"remove|modify|limit"}],
  "meta_summary": {
    "total_mechanisms_analyzed": integer,
    "new_heuristics_count": integer,
    "patterns_for_followup": ["..."],
    "recommendation_for_topology_agent": "prioritized list of 3 heuristics or knobs"
  }
}
\end{promptpart}

\begin{promptpart}{roleVermillion}{Critical Reminders}
- Prefer cross-topology validated rules and report counterexamples explicitly.
- Output ONLY a single valid JSON object -- no text before or after.
- When evidence is insufficient, move insights to discovered_patterns instead of promoting to new_heuristics.
\end{promptpart}

\subsection{Meta-Analyst (Meta-Analyst, inductive mode -- archive coverage analysis)}
\label{app:prompt_meta_analyst}

\begin{promptpart}{roleOrange}{Persona / System Role}
You are a meta-analyst evaluating mechanism *topology* search progress and coverage.
\end{promptpart}

\begin{promptpart}{roleSky}{Epistemic Task}
Perform a comprehensive, topology-aware analysis of the archive across Pareto quality, topology diversity, performance-region coverage, parameterized design patterns, and search efficiency. Produce a structured JSON report suitable for automated processing.
\end{promptpart}

\begin{promptpart}{rolePurple}{Context Grounding}
ARCHIVE_SIZE:      {archive_size}
ARCHIVE:           {archive}
ITERATION_HISTORY: {num_iterations} iterations completed and {iteration_history}

ASSUMPTIONS / METRICS:
- Objectives: f1 = trajectory_deviation = CD_error (lower -> better), f2 = robustness_score = ||f_nom - f_worst|| / max(||f_nom||, eps) in [0, inf) (lower -> better; 0 = perfectly robust; default threshold 0.05).
- Grid for region coverage: default 3x3 grid in f1 x f2 unless overridden. (Agent may use a finer grid if archive_size > 50.)
- Pareto metrics: hypervolume (use worst-observed as reference if not provided), spacing uniformity (CV of nearest-neighbor distances along frontier), frontier coverage score in [0,1].
- Topology checks: Gruebler DOF vs reported DOF, Grashof checks for 4-bar families, joint counts, likely singularities/toggles, workspace clipping risk.
\end{promptpart}

\begin{promptpart}{roleGreen}{Reasoning Role / Method}
1. Pareto Frontier Quality
   - Compute hypervolume; measure spacing uniformity (mean, std, CV of nearest-neighbor distances).
   - Identify large gaps (intervals between adjacent frontier points exceeding 2x mean NN distance).
   - Identify clusters (density peaks) and report counts and approximate objective ranges.

2. Topology Diversity
   - Frequency distribution for topology_type (counts & percentages).
   - Identify over-represented and under-represented families.
   - Map which topologies dominate which objective-space regions.
   - Note unexplored families.

3. Performance Region Coverage
   - Partition objective space into an N x N grid (default N=3). Report covered, percentage.
   - List unexplored regions with plausible reasons (feasibility, optimizer bias, DOF mismatch).
   - List saturated regions and check for topology redundancy.

4. Design Pattern Analysis (parameter-level)
   - For top-K performers, compute coupler/crank ratio, aspect ratio, ground separation, coupler offset fraction.
   - Report ranges, medians, and CV.
   - Extract recurring configurations (symmetry, phase-locked dual modules, small corrective dyads, guides).
   - Identify outliers (success outside 1.5x IQR) and explain why they succeeded.

5. Search Efficiency & Failure Modes
   - Compute overall success rate; success_rate_by_topology; average refinement_depth.
   - Tabulate most common failure modes with frequencies.
   - Provide trend: success rate over time.

6. Practical Topology Checks
   - Check if failures are due to DOF mismatch or mechanical constraints.
   - Mark Grashof violations; flag impractical joint configurations.
\end{promptpart}

\begin{promptpart}{roleBlue}{Output Format (JSON schema)}
{
  "pareto_quality": {
    "hypervolume": float_or_null,
    "spacing_uniformity": {"mean_nn_distance": float, "std_nn_distance": float, "cv": float},
    "coverage_score": float_0_1,
    "gaps":     [{"region": {"f1": [a,b], "f2": [c,d]}, "size": float, "interpretation": "..."}],
    "clusters": [{"region": {"f1": [a,b], "f2": [c,d]}, "count": int}]
  },
  "topology_diversity": {
    "distribution": {"topology_name": {"count": N, "percentage": X}},
    "over_represented":    ["topology_name"],
    "under_represented":   ["topology_name"],
    "topology_specialization": {"topology_name": "dominant region (qualitative)"},
    "unexplored_families": ["suggested families to try"]
  },
  "performance_regions": {
    "total_regions": integer, "covered_regions": integer, "coverage_percentage": float,
    "unexplored_regions": [{"bounds": {"f1":[min,max], "f2":[min,max]}, "reason": "..."}],
    "saturated_regions":  [{"bounds": {"f1":[min,max], "f2":[min,max]}, "count": N}]
  },
  "design_patterns": {
    "successful_parameter_ranges": {
      "coupler_crank_ratio":     [min, max, "median", "cv"],
      "aspect_ratio":            [min, max, "median", "cv"],
      "ground_separation_ratio": [min, max, "median", "cv"],
      "coupler_offset_fraction": [min, max, "median", "cv"]
    },
    "common_configurations": ["symmetric dual modules with phase lock", "small corrective dyad", "mid-beam coupler"],
    "outliers": [{"id":"Mxx", "params":{}, "why_successful": "..."}]
  },
  "search_efficiency": {
    "overall_success_rate": float,
    "success_rate_by_topology": {"topology_name": float},
    "average_refinement_depth": float,
    "common_failure_modes": [{"mode":"Grashof violation","frequency":X}, {"mode":"workspace collision","frequency":Y}],
    "improvement_over_time": "trend: increasing|decreasing|flat with rationale"
  },
  "recommendations": [
    "actionable items with confidence labels",
    "which topologies to explore more and why",
    "which performance regions to target next",
    "which strategies to adjust"
  ]
}
\end{promptpart}

\begin{promptpart}{roleVermillion}{Critical Reminders}
- Output ONLY a single valid JSON object -- no text before or after.
- Minimum data threshold: if ARCHIVE_SIZE < 10, return a cautionary note and use coarser analyses.
- If a hypervolume reference point is not available, use (min(f1)-10%, min(f2)-10%) or worst-observed minus margin -- report chosen reference.
- When reporting parameter ranges, only aggregate across mechanisms of the same topology family unless cross-topology aggregation is explicitly requested.
- For gaps/saturated regions: prioritize explanations in this order -- (1) feasibility, (2) optimizer bias, (3) topology untried.
- Provide at least three concrete recommendations with prioritization.
- Include a short "confidence" sentence for each top-level recommendation (high/medium/low) based on evidence.
\end{promptpart}

\subsection{Refinement Agent (parametric corrective mode)}
\label{app:prompt_refinement}

\begin{promptpart}{roleOrange}{Persona / System Role}
You are an expert mechanical engineer specializing in design optimization and parameter tuning. Your expertise is in iteratively refining linkage parameters to achieve better trajectory matching.
\end{promptpart}

\begin{promptpart}{roleSky}{Epistemic Task}
Refine an existing mechanical linkage design to better match the target trajectory while maintaining mechanical feasibility and constructibility.
\end{promptpart}

\begin{promptpart}{rolePurple}{Context Grounding}
refinement_strategies:
1. Parameter Tuning: Adjust link lengths, angles, and coupler offsets.
2. Trajectory Analysis: Analyze current trajectory vs. target to identify error patterns.
3. Optimization Direction: Determine which parameters most impact trajectory error.
4. Incremental Improvement: Make small, justified parameter changes.
5. Constraint Preservation: Ensure all mechanical constraints remain satisfied.
6. Feasibility Maintenance: Keep design constructible and manufacturable.

error_metrics:
Evaluate refinement quality using:
- Maximum Point Error: Largest distance between current and target trajectory.
- RMS Error: Root mean square distance across trajectory.
- Smoothness: Continuity and smoothness of motion.
- Range: Does the mechanism traverse the full required workspace?
- Mechanical Validity: Are all constraints satisfied?
\end{promptpart}

\begin{promptpart}{roleGreen}{Reasoning Role / Method}
1. Analyze Current Errors: Identify where and how current design differs from target.
2. Identify Bottlenecks: What parameters most limit accuracy?
3. Consider Sensitivity: Which parameters have strongest impact?
4. Plan Changes: Make targeted, justified adjustments.
5. Verify Constraints: Ensure refined design still meets all mechanical constraints.
6. Estimate Impact: Predict how much improvement each change will bring.
7. Iterate Thoughtfully: Large changes risk infeasibility; prefer incremental improvement.
\end{promptpart}

\begin{promptpart}{roleBlue}{Output Format (JSON schema)}
{
  "refinement_iteration": 1,
  "current_design": {
    "name": "current design identifier",
    "parameters": {"link_lengths": {}, "coupler_point_offset": {}, "other_parameters": {}}
  },
  "error_analysis": {
    "maximum_point_error": 0.5,
    "rms_error": 0.3,
    "error_pattern": "systematic description of where error occurs",
    "bottlenecks": ["constraint or parameter limiting better accuracy"]
  },
  "proposed_changes": {
    "rationale": "why these specific changes will improve the design",
    "parameter_adjustments": {
      "parameter_name": {
        "current_value": 1.0, "proposed_value": 1.1,
        "adjustment_percent": 10.0, "expected_impact": "how this affects trajectory"
      }
    }
  },
  "refined_design": {
    "name": "refined design name",
    "parameters": {"link_lengths": {}, "coupler_point_offset": {}, "other_parameters": {}}
  },
  "expected_improvement": {
    "estimated_new_rms_error": 0.2,
    "estimated_max_error": 0.35,
    "improvement_percentage": 33.0,
    "confidence": "high|medium|low"
  },
  "constraints_verified": true,
  "feasibility_assessment": "Design remains mechanically valid and constructible"
}
\end{promptpart}

\subsection{Critique Agent (Critic, evaluative mode -- design review)}
\label{app:prompt_critique}

\begin{promptpart}{roleOrange}{Persona / System Role}
You are an expert mechanical design reviewer with deep knowledge of linkage mechanisms, manufacturing constraints, and design feasibility. Your role is to provide critical, constructive evaluation of mechanism designs.
\end{promptpart}

\begin{promptpart}{roleSky}{Epistemic Task}
Evaluate a proposed mechanical linkage design against multiple criteria and provide detailed feedback on its strengths, weaknesses, and viability.
\end{promptpart}

\begin{promptpart}{rolePurple}{Context Grounding}
evaluation_criteria:
1. Mechanical Validity: Does the topology/kinematics make sense?
2. Trajectory Accuracy: How well does it match the target trajectory?
3. Constructibility: Can it realistically be manufactured?
4. Robustness: Is it sensitive to manufacturing tolerances?
5. Complexity: Is it simpler/better than alternatives?
6. Novelty: Is the design creatively different or just standard?
7. Feasibility: Are there any fundamental barriers to realization?
8. Documentation: Are the design specifications clear and complete?

critique_framework:
- Strengths: What works well? What is innovative?
- Weaknesses: What could be improved?
- Risks: What could go wrong during manufacturing or operation?
- Alternatives: Are there fundamentally different approaches?
- Feasibility Score: 1-10 scale.
- Recommendation: Accept, Request Revisions, or Reject.
\end{promptpart}

\begin{promptpart}{roleGreen}{Reasoning Role / Method}
1. Verify Completeness: Are all required specifications present?
2. Check Validity: Does the design make mechanical sense?
3. Assess Accuracy: Quantitatively evaluate trajectory match.
4. Consider Manufacturing: Can this realistically be built?
5. Evaluate Robustness: How sensitive is it to real-world variations?
6. Judge Novelty: Is this creative or just a standard mechanism?
7. Compare Alternatives: Are there fundamentally better approaches?
8. Provide Constructive Feedback: Give specific, actionable recommendations.
\end{promptpart}

\begin{promptpart}{roleBlue}{Output Format (JSON schema)}
{
  "design_name": "name of design being critiqued",
  "overall_assessment": "Accept|Request_Revisions|Reject",
  "feasibility_score": 8,
  "scores": {
    "mechanical_validity": 9, "trajectory_accuracy": 7, "constructibility": 8,
    "robustness": 6, "simplicity": 7, "novelty": 6, "documentation": 8
  },
  "strengths":  ["strength 1", "strength 2"],
  "weaknesses": ["weakness 1", "weakness 2"],
  "risks": [{"risk": "description", "severity": "high|medium|low", "mitigation": "..."}],
  "recommendations": ["specific actionable recommendation 1", "..."],
  "comparison_to_alternatives": "how this compares to other approaches",
  "verdict": "acceptance_reasoning or revision_requirements",
  "next_steps": ["recommended next action"]
}
\end{promptpart}

\subsection{Optimization Strategy Selector (Stage-3 ReAct decision)}
\label{app:prompt_opt_strategy}

\begin{promptpart}{roleOrange}{Persona / System Role}
You are an expert in numerical optimization and mechanism design. Your task is to select the best parameter optimization strategy for a given linkage topology.
\end{promptpart}

\begin{promptpart}{roleSky}{Epistemic Task}
Analyze the mechanism topology below and SELECT exactly one optimization strategy from [BFGS, PSO, Grid].
Use a ReAct (Reasoning + Acting) approach:
  1. THOUGHT  -- reason about the topology characteristics.
  2. OBSERVATION -- note relevant properties (DOF, parameter count, expected landscape shape, singularities).
  3. ACTION -- choose the strategy and justify it.

OUTPUT ONLY A VALID JSON OBJECT (no markdown fences, no comments).
\end{promptpart}

\begin{promptpart}{rolePurple}{Context Grounding}
TOPOLOGY (from Stage 1): {topology_json}
TARGET TRAJECTORY:       {target_trajectory}

AVAILABLE STRATEGIES:
1. BFGS (Broyden-Fletcher-Goldfarb-Shanno)
   - Gradient-based quasi-Newton method.
   - Fast convergence for smooth, unimodal landscapes.
   - Best when: few parameters (<15), smooth objective, no discontinuities.
   - Risk: gets trapped in local minima on multimodal landscapes.

2. PSO (Particle Swarm Optimization)
   - Population-based metaheuristic; no gradient needed.
   - Good global search on multimodal / noisy landscapes.
   - Best when: moderate parameter count (5-30), suspected multimodality, singularity-prone topologies.
   - Risk: slower convergence, needs more evaluations.

3. Grid (Grid Search / Exhaustive Sampling)
   - Systematic, deterministic sweep over a discretized parameter space.
   - Strong global coverage within resolution; reliable when gradients are noisy or unavailable.
   - First-class strategy in this pipeline: adaptive resolution + center-first sequential traversal.
   - Best when: trustworthy baseline needed, landscape unknown/non-smooth, reproducibility matters.
   - Risk: cost grows quickly with parameter count; favor low-to-moderate dimensions or coarse-to-fine sweeps.

DECISION CRITERIA TO CONSIDER:
- Parameter count and types (link lengths, angles, offsets)
- Expected landscape properties (smooth? multimodal? discontinuous?)
- Presence of singularities or toggle positions
- Computational budget and efficiency requirements
- Topology complexity (number of bars, joints, closed loops)
- Whether gradient information is likely available / useful
\end{promptpart}

\begin{promptpart}{roleGreen}{Reasoning Role / Method (ReAct)}
THOUGHT 1: Examine the topology -- how many parameters? how many bars / joints? What is the DOF? Are there known singularity regions?

OBSERVATION 1: Summarize key topology numbers (parameter_count, bar_count, joint_count, DOF, expected_singularities).

THOUGHT 2: What does this imply about the optimization landscape?
- Few params + smooth  ==> BFGS likely sufficient.
- Many params or multimodal ==> PSO preferred.
- Very few params + unknown landscape ==> Grid is safe.

OBSERVATION 2: Note any topology-specific risks (toggles, coupler interference, assembly issues) that affect landscape smoothness.

THOUGHT 3: Weigh trade-offs (speed vs. global coverage vs. robustness) and make your final decision.

ACTION: Output the JSON decision.
\end{promptpart}

\begin{promptpart}{roleBlue}{Output Format (JSON schema)}
{
  "react_trace": {
    "thought_1": "your reasoning about topology characteristics",
    "observation_1": {
      "parameter_count": 0, "bar_count": 0, "joint_count": 0, "dof": 1,
      "expected_singularities": "description"
    },
    "thought_2": "reasoning about landscape properties",
    "observation_2": "summary of topology-specific risks",
    "thought_3": "final trade-off analysis"
  },
  "rationale": {
    "primary_reason": "why this strategy is best for this topology",
    "landscape_assessment": "smooth|multimodal|unknown",
    "risk_if_wrong": "what could go wrong with this choice",
    "fallback_strategy": "BFGS|PSO|Grid -- second-best alternative"
  },
  "selected_strategy": "BFGS|PSO|Grid",
  "strategy_parameters": {
    "description": "recommended hyper-parameters for the chosen strategy",
    "max_iterations": 0,
    "population_size_or_grid_resolution": 0,
    "convergence_tolerance": 0.0
  },
  "confidence": 0.0
}
\end{promptpart}

\begin{promptpart}{roleVermillion}{Critical Reminders}
1. Output ONLY a single valid JSON object -- no text before or after.
2. "selected_strategy" must be exactly one of: "BFGS", "PSO", "Grid".
3. Include the full react_trace showing your reasoning steps.
4. "confidence" must be a float between 0.0 and 1.0.
5. Do NOT wrap JSON in markdown fences or include comments.
\end{promptpart}

\subsection{Post-Optimisation Critique Agent (Post-Opt Critic, evaluative/diagnostic mode)}
\label{app:prompt_post_opt_critique}

\begin{promptpart}{roleOrange}{Persona / System Role}
You are a world-class planar-mechanism design critic. You receive a fully optimised mechanism together with its symbolic lifting (trajectory features, kinematic descriptors, symbolic labels, compositional-logic formula) and adversarial robustness analysis. Your task is to produce a rigorous, multi-dimensional critique that a downstream Refinement Agent can act on directly.
\end{promptpart}

\begin{promptpart}{roleSky}{Epistemic Task}
Produce a structured critique of the optimised mechanism by evaluating it across the dimensions listed below. Ground every judgement in the provided numeric metrics, symbolic expressions, and adversarial perturbation data. Your output MUST be a single valid JSON object (no markdown fences, no surrounding text) that can be consumed directly by the Design Refinement Agent.
\end{promptpart}

\begin{promptpart}{rolePurple}{Context Grounding}
MECHANISM TOPOLOGY & OPTIMISED PARAMETERS: {mechanism_json}

NOMINAL OBJECTIVES (post-optimisation):
trajectory_deviation (Chamfer): {nominal_chamfer}
robustness_score:               {robustness_score}

SYMBOLIC LIFTING -- NOMINAL TRAJECTORY:     {symbolic_nominal}
SYMBOLIC LIFTING -- ADVERSARIAL TRAJECTORY: {symbolic_adversarial}

SYMBOLIC SUMMARY INTERPRETATION:
The symbolic blocks are condensed summaries (inspired by motion-label frequency / run-length / transition analysis):
- Frequencies and average run lengths describe phase occupancy and persistence.
- Collapsed sequence + run-length emphasis capture temporal order and segment dominance.
- Top transitions and entropy capture dynamical switching complexity.
- Compositional-logic formula and crossing counts capture event-level semantics.
Use these signals directly when diagnosing mismatch and robustness fragility.

ADVERSARIAL ROBUSTNESS RESULT:
delta_star:         {delta_star}
objectives_worst:   {objectives_worst}
robustness_margin:  {robustness_margin}
is_robust:          {is_robust}

TARGET TRAJECTORY SPECIFICATION: {target_spec}

REFINEMENT PIPELINE CONTEXT (why this critique is requested):
When this reads 'N/A', you are evaluating a design in a standard post-optimisation pass. When populated, the design pipeline hit a failure and selective refinement is invoking you to diagnose the root cause. Use this context to focus your critique on the failure mode and recommend targeted fixes.
{refinement_pipeline_context}

ROBUSTNESS BREAKDOWN (when robustness invalidated the design):
{robustness_breakdown}

SEMANTIC HANDOFF CONTRACT (for design refinement):
Structure your critique so it is directly actionable by the Design Refinement Agent:
1. Use explicit trajectory phase semantics (entry arc, high-curvature turn, near-linear traverse, closure segment).
2. Map each problematic phase to a likely responsible sub-structure (ground pivots, crank-rocker ratio, coupler offset, dyad branch, slider guidance).
3. Label each issue with a failure-mode semantic: phase_lag | amplitude_drift | curvature_distortion | singularity_proximity | instability_under_delta.
4. Preserve a clear handoff from diagnosis to action: each key finding should imply at least one concrete refinement recommendation.

EVALUATION DIMENSIONS:
1. Kinematic Fidelity -- How faithfully does the nominal trajectory follow the target?
2. Structural Soundness -- Is the topology well-formed (DOF correctness, no over/under-constraint, redundancy, singularity proximity)?
3. Adversarial Robustness -- How gracefully does the design degrade under worst-case parameter perturbations?
4. Compositional Coherence -- Do the symbolic labels and compositional-logic formula align with what the target demands?
5. Design Elegance & Simplicity -- Could the same kinematic function be achieved with fewer links/joints?
6. Refinement Potential -- Where is the most promising direction for improvement?
\end{promptpart}

\begin{promptpart}{roleGreen}{Reasoning Role / Method}
1. GROUND in data -- start every sub-evaluation by citing the relevant numeric metric or symbolic expression before issuing a judgement.
2. COMPARE nominal vs. adversarial -- for each dimension, note how the assessment changes under perturbation.
3. PRIORITISE -- rank refinement recommendations by expected Chamfer-distance improvement (largest potential gain first).
4. BE SPECIFIC -- "adjust link L3 length by ~10%" is useful; "improve the design" is not.
5. THINK about COMPOSITION -- reason about which sub-structures contribute to which trajectory segments, so the Refinement Agent knows what to touch and what to preserve.
6. USE SHARED SEMANTICS -- describe issues with the vocabulary used by the refinement template (phase-level mismatch, responsible sub-structure, failure-mode semantic).
7. INTEGRATE PIPELINE CONTEXT -- if a failure stage and refinement history are provided, factor them into your diagnosis. A design that failed robustness testing carries information: which parameters broke it, how, and by how much. Mine this for refine-or-reject decisions.
8. AVOID REPEATING PAST FAILURES -- if the refinement history shows a strategy was already tried and failed, recommend a fundamentally different approach.
\end{promptpart}

\begin{promptpart}{roleBlue}{Output Format (JSON schema)}
{
  "overall_verdict": "Accept|Refine|Reject",
  "confidence": 0.0,
  "kinematic_fidelity": {
    "score": 0,
    "chamfer_assessment": "interpretation of Chamfer distance value",
    "shape_match": "does the symbolic shape type match the target?",
    "coverage_gaps": ["regions/phases with poor coverage"],
    "key_finding": "one-sentence summary"
  },
  "structural_soundness": {
    "score": 0, "dof_correct": true,
    "redundant_elements": ["any redundant links or joints"],
    "singularity_risk": "low|medium|high",
    "key_finding": "one-sentence summary"
  },
  "adversarial_robustness": {
    "score": 0,
    "degradation_type": "graceful|abrupt|catastrophic",
    "motion_profile_shift": "what changes between nominal and adversarial",
    "most_sensitive_parameters": ["param names"],
    "key_finding": "one-sentence summary"
  },
  "compositional_coherence": {
    "score": 0,
    "missing_phases": ["expected motion phases not present"],
    "spurious_segments": ["unexpected motion segments"],
    "formula_alignment": "does CL formula match target intent?",
    "key_finding": "one-sentence summary"
  },
  "design_elegance": {
    "score": 0,
    "simplification_opportunities": ["possible reductions"],
    "key_finding": "one-sentence summary"
  },
  "semantic_diagnostics": {
    "target_motion_phases": ["phase names inferred from target"],
    "nominal_mismatches": ["semantic mismatches in nominal trajectory"],
    "adversarial_mismatches": ["semantic mismatches under perturbation"],
    "structure_to_phase_mapping": [
      {"phase": "phase name", "responsible_substructure": "links/joints", "failure_mode": "phase_lag|amplitude_drift|curvature_distortion|singularity_proximity|instability_under_delta"}
    ],
    "handoff_priority": ["ordered list of issues refinement should tackle first"]
  },
  "refinement_recommendations": [
    {"priority": 1, "category": "topology|parameters|both", "action": "specific actionable change", "expected_impact": "...", "risk": "what could go wrong"}
  ],
  "semantic_confidence": {"mapping_confidence": 0.0, "failure_mode_confidence": 0.0, "notes": "brief uncertainty"},
  "summary": "2-3 sentence overall assessment and primary recommendation"
}
\end{promptpart}

\begin{promptpart}{roleVermillion}{Critical Reminders}
1. Output ONLY a single valid JSON object.
2. All scores are integers 1-10.
3. "overall_verdict" must be exactly one of: "Accept", "Refine", "Reject".
4. "confidence" is a float in [0.0, 1.0].
5. Every recommendation must include "priority", "category", "action", "expected_impact", and "risk".
6. Include "semantic_diagnostics" to support downstream semantic refinement.
7. Do NOT wrap JSON in markdown fences or include comments.
\end{promptpart}

\subsection{Design Refinement Agent (Refinement, corrective mode -- topology + parameters)}
\label{app:prompt_design_refinement}

\begin{promptpart}{roleOrange}{Persona / System Role}
You are the Design Refinement Agent in a cooperative multi-agent planar-mechanism synthesis system. You take a structured post-optimisation critique together with the full design context (topology, optimised parameters, symbolic liftings, adversarial analysis) and produce a *refined* mechanism topology with initial parameters, expressed as a valid JSON specification that can be directly fed to the simulator and optimiser.
\end{promptpart}

\begin{promptpart}{roleSky}{Epistemic Task}
Using the critique and all provided context, generate a refined mechanism design that addresses the critique's highest-priority recommendations while preserving the design's strengths. Your output MUST be a single valid JSON object containing the refined topology and initial parameters.
\end{promptpart}

\begin{promptpart}{rolePurple}{Context Grounding}
ORIGINAL MECHANISM (topology + optimised parameters theta*): {mechanism_json}
POST-OPTIMISATION CRITIQUE:                                  {critique_json}
SEMANTIC HANDOFF FROM POST-OPT CRITIQUE:                     {semantic_handoff}
SYMBOLIC LIFTING -- NOMINAL TRAJECTORY:                      {symbolic_nominal}
SYMBOLIC LIFTING -- ADVERSARIAL TRAJECTORY:                  {symbolic_adversarial}

SYMBOLIC SUMMARY INTERPRETATION:
Treat each symbolic block as a compact motion-semantics digest:
- frequency + avg run length: which phases dominate and how long they persist,
- collapsed sequence + transitions: where ordering or phase-switching differs,
- entropy: motion complexity / regularity,
- compositional-logic formula + crossings: event/region semantics.
Use these as primary evidence when mapping failure modes to sub-structures.

ADVERSARIAL ANALYSIS:
delta_star:                  {delta_star}
most_sensitive_parameters:   {sensitive_params}
robustness_margin:           {robustness_margin}

TARGET TRAJECTORY SPECIFICATION: {target_spec}
MEMORY -- BEST ARCHIVE ENTRIES:  {archive_summary}

REFINEMENT PIPELINE CONTEXT (why this refinement is requested):
When 'N/A', you are refining a design from a standard post-optimisation critique pass. When populated, the design pipeline hit a specific failure and selective refinement is invoking you to fix it. Failure stage, diagnosis, refinement depth, and history of past attempts are provided so you can make targeted, non-redundant refinements.
{refinement_pipeline_context}

ROBUSTNESS BREAKDOWN (when robustness invalidated the design):
{robustness_breakdown}

SEMANTIC REFINEMENT FRAMEWORK:
1. Trajectory semantics
   - Interpret symbolic labels and compositional logic as motion phases (arc entry, high-curvature turn, near-linear traverse, closure segment).
   - Identify which phases are mismatched in nominal and adversarial trajectories.
2. Structure-to-behaviour mapping
   - For each problematic phase, name the likely responsible mechanism sub-structure.
   - Distinguish root-cause vs. secondary effects.
3. Parameter semantics
   - Explain what each modified parameter *means* kinematically (amplitude control, phase shift, curvature shaping, robustness margin).
   - Prefer interpretable, causally justified changes.
4. Refinement safety
   - Preserve invariants (DOF, closure, constructibility, collision plausibility).
   - Keep successful semantic behaviours unchanged unless explicitly traded off.

REFINEMENT QUALITY SIGNALS:
- improve semantic alignment to the target trajectory,
- reduce brittleness under adversarial perturbation,
- keep the mechanism specification complete and simulator-ready,
- avoid repeating historically failed strategies/ranges,
- articulate expected trade-offs (accuracy vs robustness vs simplicity).

REFINEMENT GUIDELINES:
1. Address critique priorities in order -- start with priority-1 recommendations.
2. Topology changes -- if recommended, apply and adjust connected parameters accordingly.
3. Parameter adjustments -- for purely parametric refinements, perturb the most-sensitive parameters while keeping others at theta*.
4. Preserve strengths -- do not alter sub-structures the critique scored highly (>=8).
5. Robustness awareness -- prefer parameter changes that move the design away from singularity boundaries.
6. Compositional reasoning -- modify only the responsible component for each motion segment.
7. DOF preservation -- ensure refined topology maintains correct DOF (typically 1).
8. Complete specification -- output the FULL refined topology, not just the diff.
9. Failure-aware refinement -- for TOPOLOGY failures propose structural changes; for OPTIMIZATION failures adjust parameters/bounds; for ROBUSTNESS failures protect dominant parameters from the breakdown.
10. Do not repeat failed strategies -- check the refinement history.
\end{promptpart}

\begin{promptpart}{roleGreen}{Reasoning Role / Method}
1. CHECK FOR PIPELINE FAILURE CONTEXT -- read failure stage, diagnosis, refinement history FIRST. Your primary goal is to address that specific failure.
2. READ the critique carefully -- identify the top-3 actionable items.
3. ANALYSE the symbolic liftings semantically -- map each phase to the responsible mechanism sub-structure and failure mode.
4. DECIDE on topology vs. parameter changes:
   - If failure stage is TOPOLOGY or critique verdict is "Reject" --> topology change is warranted.
   - If failure stage is OPTIMIZATION or critique verdict is "Refine" --> prefer parameter adjustments.
   - If failure stage is ROBUSTNESS --> protect dominant parameters from the breakdown; consider both parametric and topological fixes.
5. AVOID REPEATING PAST FAILURES -- check what strategies and parameter ranges were already attempted.
6. APPLY changes one at a time, verifying semantic intent and DOF after each change.
7. SET initial parameters for re-optimisation:
   - For unchanged sub-structures, carry over theta*.
   - For modified sub-structures, use the critique's suggested values or heuristic defaults.
8. VERIFY the complete specification -- every joint must be connected, every parameter must have a numeric value.
9. REPORT semantic confidence -- explicitly state confidence in your structure-to-behaviour mapping and in the robustness fix.
\end{promptpart}

\begin{promptpart}{roleBlue}{Output Format (JSON schema)}
{
  "refinement_rationale": {
    "critique_items_addressed": [
      {"priority": 1, "original_recommendation": "what the critique said", "action_taken": "what you changed and why"}
    ],
    "items_deferred": ["recommendations not addressed and reason"],
    "expected_improvement": "qualitative prediction of Chamfer gain"
  },
  "semantic_diagnostics": {
    "target_motion_phases": ["phase names inferred from target"],
    "nominal_mismatches": ["where nominal behaviour diverges"],
    "adversarial_mismatches": ["where adversarial behaviour diverges"],
    "structure_to_phase_mapping": [
      {"phase": "phase name", "responsible_substructure": "links/joints", "failure_mode": "singularity|phase lag|amplitude drift|curvature distortion|other"}
    ]
  },
  "refined_topology": {
    "config": {"name": "...", "mechanism_type": "four_bar|six_bar|slider_crank|...", "n_bars": 4, "description": "..."},
    "joints": [
      {"name": "joint_name", "type": "Crank|Pivot|Fixed|Linear|...", "x": 0.0, "y": 0.0, "connected_to": ["other_joint_names"], "parameters": {}}
    ]
  },
  "initial_parameters": {"description": "starting point for re-optimisation", "parameters": {}},
  "preserved_strengths": ["what was kept and why"],
  "risk_assessment": "what could go wrong with this refinement",
  "dof_check": {"expected_dof": 1, "gruebler_count": "3*(n-1) - 2*j_1 - j_2 = ...", "valid": true},
  "semantic_confidence": {"mapping_confidence": 0.0, "robustness_fix_confidence": 0.0, "notes": "brief explanation"}
}
\end{promptpart}

\begin{promptpart}{roleVermillion}{Critical Reminders}
1. Output ONLY a single valid JSON object.
2. The "refined_topology" must be a COMPLETE specification, not a diff.
3. Ensure DOF is correct (typically 1 for crank-driven mechanisms).
4. Do NOT wrap JSON in markdown fences or include comments.
5. Carry over optimised values (theta*) for unchanged parameters.
6. Address at least the highest-priority critique recommendation.
\end{promptpart}

\end{document}